\documentclass[11pt]{article}

\usepackage[numericcite]{arxivpreprint}

\usepackage{courier}          % Type 1 monospace for \texttt{}; arXiv-safe
\usepackage{amsfonts}         % \mathbb etc. (subset of amssymb but harmless)
\usepackage{nicefrac}         % \nicefrac{1}{2}
\usepackage{subcaption}       % subfigures
\usepackage{algorithm}        % algorithm float
\usepackage{algorithmic}      % algorithmic environment
\usepackage{multirow}
\usepackage{longtable}
\usepackage{multicol}
\usepackage{array}
\usepackage{url}              % \url{} (hyperref already provides, but the paper
\usepackage[most]{tcolorbox}  % for the takeaway callout box defined below
\usepackage[printonlyused]{acronym}
\usepackage{enumitem}
\setlist[description]{nolistsep}

\newtcolorbox{takeaway}{
  enhanced,
  colback=blue!5,
  colframe=blue!70!black,
  boxrule=0.6pt,
  arc=3pt,
  left=8pt, right=8pt, top=5pt, bottom=5pt,
  width=0.95\linewidth,
  before=\begin{center},
  after=\end{center},
}

\newcolumntype{R}[1]{>{\raggedleft\arraybackslash}m{#1}}

\makeatletter
\let\ap@orig@texttt\texttt
\protected\def\texttt#1{\ap@orig@texttt{\ap@tt@break#1\ap@tt@end}}
\def\ap@tt@break#1{%
  \ifx#1\ap@tt@end \let\next\relax
  \else
    \ifx#1- -\discretionary{}{}{}%
    \else\ifx#1. .\discretionary{}{}{}%
    \else\ifx#1/ /\discretionary{}{}{}%
    \else\ifx#1_ \_\discretionary{}{}{}%
    \else #1\fi\fi\fi\fi
    \let\next\ap@tt@break
  \fi
  \next}
\def\ap@tt@end{\ap@tt@end}
\makeatother

\title{Neuro-Inspired Inverse Learning \\ for Planning and Control}

\author[1,2]{Maryna Kapitonova}
\author[1,2]{Tonio Ball\thanks{Correspondence:
  \href{mailto:tonio.ball@neuromentum.ai}{tonio.ball@neuromentum.ai}}}
\affil[1]{{\ttfamily\normalsize NeuroMentum AI}}
\affil[2]{IMBIT, University of Freiburg, Germany}

\date{May 2026}

\runningtitle{Inverse Learning for Planning and Control}

\begin{document}

\maketitle
\repolink[qronly]{https://github.com/neuromentum/inverse-learning}

\begin{abstract}
We present a neuro-inspired framework for embodied planning and control.  Building on three principles that enable fast and highly effective goal-directed behavior in the mammalian brain --- paired forward/inverse internal models, open-loop multi-step motor commands, and sequential\textcolor{black}{, hierarchical} organization of action --- our \emph{Inverter} framework \textcolor{black}{uses learned components, trained end-to-end through \emph{\acf{IL}} and supplemented where natural by analytic or algorithmic modules; we formalize \acs{IL} and delineate it from supervised, reinforcement, and imitation learning.}  \acs{IL} bridges \acf{RL}-style amortization, which runs in a single forward pass but emits only one action at a time, and \acf{OC}-style sequence planning over whole trajectories, but with iterative test-time computation.  Single Inverters or hierarchical $n{=}2$ Inverter stacks match or improve on offline-\acs{RL} and diffusion-planner baselines on all 3 \texttt{maze2d} and 6 \texttt{antmaze} \acs{D4RL} variants by an average of $+24.2\%$ (range $-1.9\%$ to $+78.2\%$), at one-to-two orders of magnitude less inference compute \textcolor{black}{time}. Distinctively, optimizing through the \acf{FoM} over the entire $T$-step action sequence -- rather than per step -- lets Inverters produce smooth, goal-coherent, trajectory-wide structure and reach control policies closer to the analytic optimum than the policy underlying the training data itself. We also identify a failure mode of \acs{IL}: \acs{FoM} hacking under narrow training-data coverage, which we mitigate by using \emph{random} training data with broader coverage.  As an application example, a Pulse Inverter synthesizes arbitrary single-qubit quantum gates with fidelity matching the standard iterative numerical baseline (\acs{GRAPE}), at more than $1000{\times}$ lower per-gate compute \textcolor{black}{time}. In summary, we conclude that \acs{IL} enables a versatile class of world-interfaces, especially for latency- and resource-critical embodied AI.
\end{abstract}

%\textcolor{black}{Across the Inverter solutions, a fourth principle naturally emerged: to solve their respective tasks, in three of our four Inverters a natural solution was to compose the neural amortized component with a symbolic substrate enabling fast and effective neurosymbolic planning and control.}

% ============================================================================
\section{Introduction}
\label{sec:introduction}
% ============================================================================

Humans are able to generate explicit plans on multiple timescales even before we start acting on them. Imagine, for example, planning a trip to Paris.  Even before we lift a finger, we might already think about  booking a train and finding a hotel.  We also might already know how we will approach each of these steps --- for example, how to navigate the booking site of our choice to find a suitable accommodation.  And when we reach for our laptop, eager to begin, our hand may sweep along a smooth, pre-shaped trajectory --- a \emph{``ballistic''} movement planned as a whole. Human goal-directed behavior is thus fundamentally organized as a hierarchy of planning and control across widely separated timescales, often laid out, in some form, \emph{before} any action is taken.

Such multi-timescale, pre-optimized plans over whole sequences of actions are not the central focus of dominant learning-based paradigms for continuous control such as \acs{RL}, which is designed to emit a single reactive action at a time, or \acs{OC}, which aims to optimize over whole trajectories but typically iteratively at runtime --- a property also shared by related frameworks such as active inference. Here we present a paradigm for learning planning and control whose principles are inspired by the functional organization of the mammalian brain, and whose conceptual center is the fast and effective \emph{pre-generation} of such hierarchical, multi-timescale plans and action sequences for goal-directed, embodied behavior. We start by considering the brain as an inversion machine.

\begin{figure*}[!t]
\centering
\includegraphics[width=0.86\textwidth]{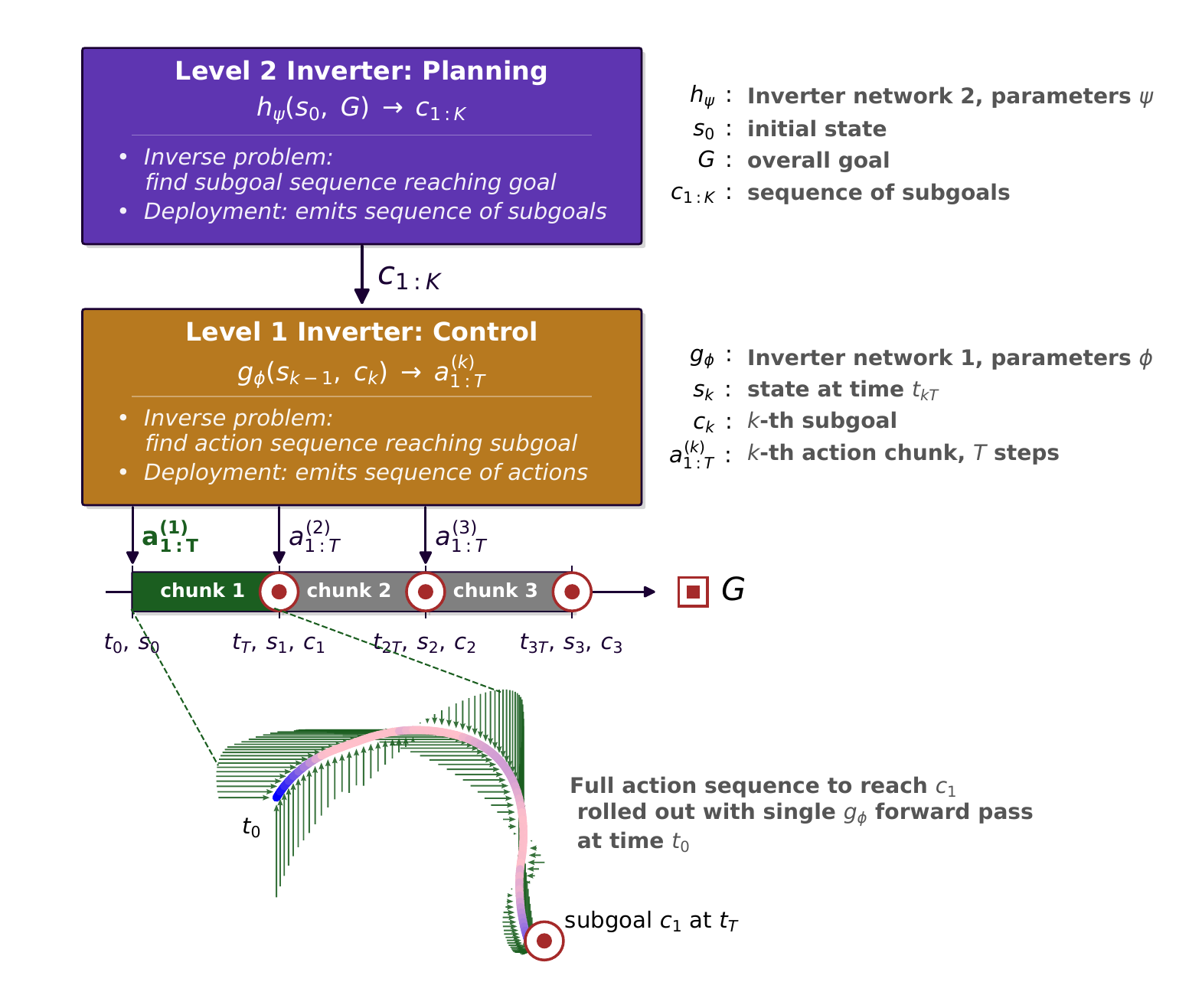}
\caption{\textbf{The Inverter planning and control framework.}  \textbf{(A)}~Schematic with two paired Inverters at different abstraction levels.  The \emph{Level 1 Inverter (Control)} $g_\phi(s_{k-1}, c_k) \!\to\! a^{(k)}_{1:T}$ is trained by \acs{IL} through a \acs{FoM} and produces each $T$-step action chunk in a single feedforward pass per chunk---no inner optimization loop, no autoregressive decoding, no iterative denoising.  The \emph{Level 2 Inverter (Planning)} $h_\psi(s_0, G) \!\to\! c_{1:K}$ emits a sequence of $K$ subgoals, one per chunk. The framework extends naturally to deeper hierarchies ($n\!\geq\!3$, as discussed in the Outlook).  The chunks tile the time axis and ultimately drive the agent to the goal $G$.  \textbf{(B)}~Illustrative single-shot example.  From the start state ($\bullet$), the Inverter emits a full action sequence in one feedforward pass; the green arrows are the planned per-step actions; the curve is the trajectory that emerges from executing those actions through the environment dynamics, with instantaneous speed color-coded.  Mirroring the optimal-feedback-control account of neurobiological motor control~\citep{todorov2002optimal}, our framework thus rejects the notion of a ``desired trajectory'': The properties of the observed motion emerge from the action sequence.}
\label{fig:inverter_sequencer}
\end{figure*}

\paragraph{The brain as an inversion machine.}
The dominant Bayesian-brain view has conceptualized the brain as an inference machine~\citep{knill2004bayesian,friston2010free}. We conceptualize the brain first and foremost as an \emph{inversion} machine: goal-directed behavior poses the inverse problem \emph{``given a desired outcome, which actions realize it?''}; solving this inverse problem minimizes the discrepancy between desired and actual outcomes. The computational core is inversion of a model of how the world responds to action --- a problem admitting a rich landscape of implementations: iterative or amortized (a learned direct inverse mapping in one feedforward pass); probabilistic or deterministic; over fully or partially observable, single- or multi-agent dynamics; with continuous, discrete, or hybrid state and action spaces; and realized through closed-form, algorithmic, or learned components.  We presume the brain flexibly uses whichever solution fits the task, timescale, and resources~\citep{gershman2015computational,lieder2020resource,gigerenzer2009homo}. Here we focus on the amortized, deterministic, single-agent, and fully observable direct mapping variant, following the classical motor control concepts of Jordan \& Rumelhart's distal-teacher framework~\citep{jordan1992forward} and Kawato's internal-model theory~\citep{kawato1999internal,wolpert1998multiple}. The combination of three essential principles organizes our approach:
\begin{itemize}
\setlength{\itemsep}{1pt}\setlength{\parskip}{0pt}\setlength{\topsep}{2pt}
\item \emph{Paired \acfp{FoM} and \acfp{IM}}, with the \acs{FoM}\footnote{Abbreviations used throughout the paper are listed in App.~\ref{app:abbreviations}.} providing the instructive error signal~\citep{jordan1992forward,kawato1999internal,wolpert1998multiple}.
\item \emph{Open-loop multi-step motor commands} ballistically executable in pre-planned chunks too fast for sensory correction~(\textcolor{black}{\citealp{zehr1994ballistic}}, \citealp{desmurget2000forward}).
\item \textcolor{black}{\emph{Sequential, hierarchical organization of action}, segmenting behaviors into sequential sub-plans and nesting them across levels of timescales and abstraction, with higher-order areas issuing subgoals to lower-level loops~\citep{graybiel1998chunking,diedrichsen2015motor,ball1999role,hasson2025uncovering}.}
\end{itemize}

\begin{table}[!htbp]
\centering
\small
\caption{\textcolor{black}{\textbf{The five Inverters in this paper.} Each level inverts a different forward process and emits its multi-step output sequence in one feedforward pass; four are neural Inverters trained by \acs{IL}, one is a simple algorithmic \acs{BFS} Path Inverter.}}
\label{tab:inverters}
{\color{black}
\begin{tabular}{@{}l c p{4.6cm} p{3.6cm} l@{}}
\toprule
\textbf{Inverter} & \textbf{Level} & \textbf{Inverse problem} & \textbf{Emitted sequence} & \textbf{\acs{IM} type} \\
\midrule
Motor      & 1 & Reach subgoal $c$ from state $s_0$ under \texttt{maze2d} point-mass dynamics & 2-dim continuous forces $(a_x, a_y)_{1:T}$, $T{\in}\{16, 128\}$ & Neural \\
Locomotion & 1 & Reach subgoal $c$ from state $s_0$ under \acs{MuJoCo} Ant dynamics            & 8-dim continuous joint torques $a_{1:T}$, $T{=}16$              & Neural \\
Path       & 2 & Route from $s_0$ to goal $G$ through long maze corridors                  & $c_{1:K}$ subgoals, one per chunk                                & Algorithmic \\
Game       & 2 & Maximize pellets in AntMan game from game state $s_0$                     & $c_{1:32}$ waypoint directions                                   & Neural \\
Pulse      & 1 & Synthesize target unitary $U \!\in\! \mathrm{U}(2)$ on a noisy 3-level transmon & 2-channel $(\Omega_x, \Omega_y)_{1:80}$ pulse slices       & Neural \\
\bottomrule
\end{tabular}
}
\end{table}

\paragraph{The Inverter planning and control framework.}
We organize the three principles above into a planning/control framework (Fig.~\ref{fig:inverter_sequencer}) consisting of a hierarchy of Inverters at $n\!=\!1, 2, 3, \ldots$ abstraction levels, all sharing the same building block (an inverse-learning network through a \acs{FoM}). In this paper we focus on the two levels we evaluate empirically (Tab.~\ref{tab:inverters}):
\begin{itemize}
\setlength{\itemsep}{1pt}\setlength{\parskip}{0pt}\setlength{\topsep}{2pt}
\item \emph{Level 1 Inverter (Control)}: emits sequences of actions -- continuous and physical in our experiments (motor torques, control pulses), or discrete signals (e.g., API calls) in other domains.
\item \emph{Level 2 Inverter (Planning)}: emits sequences of subgoals consumed by Level 1.
\end{itemize}
The same recursive shape extends naturally to $n\!\geq\!3$ Inverters, each specifying how the level immediately below generates its output -- for instance by selecting a subgoal-emission strategy, by composing sequences-of-sequences of subgoals, or by other context-dependent forms; we do not evaluate $n\!\geq\!3$ here (see Outlook). {\color{black}Neural architectures used within the framework can vary by domain --- all maze \acp{FoM} and Inverters in this paper are transformers, while the quantum Pulse Inverter is an \acf{MLP}, which suits its compact target without a temporal context.  Where an analytic or algorithmic solution is natural and more useful, we substitute it for the \acf{NN}:
\begin{itemize}
\setlength{\itemsep}{1pt}\setlength{\parskip}{0pt}\setlength{\topsep}{2pt}
\item \emph{Lindblad channel as \acs{FoM}} (Pulse Inverter, Sec.~\ref{sec:quantum}): the noisy-transmon dynamics are governed by a known master equation, so a learned \acs{FoM} would only approximate physics we already have in closed form.
\item \emph{\acf{BFS}-based Path Inverter on the offline-data occupancy grid} (\texttt{maze2d-medium}/\texttt{large}, Sec.~\ref{sec:inverter_plus_sequencer}): waypoint routing through long corridors is a discrete shortest-path subproblem that a simple algorithmic \acs{BFS} already solves on the data's support, so a learned Level~2 Inverter would add complexity without benefit.
\end{itemize}
This strategy corresponds to the task-, timescale-, and resource-dependent flexibility we presume in the brain~\citep{gershman2015computational,lieder2020resource,gigerenzer2009homo}. Interestingly, in practice, this implementation approach gave rise to neurosymbolic patterns across several of our Inverters -- which we revisit as a fourth, emergent organizing principle in the Discussion (Sec.~\ref{sec:discussion}).} Next, we formalize the \acs{IL} paradigm including training rules.

% ============================================================================
\section{Neuro-Inspired Inverse Learning}
\label{sec:inverse_learning}
% ============================================================================

As a \emph{paradigm} here we formalize \textbf{\acl{IL}}--training a network by differentiating a task objective through a learned \acs{FoM} \textcolor{black}{(as a form of self-supervised learning through a trained \acs{FoM})}, delineated from supervised, reinforcement, and imitation learning (Table~\ref{tab:paradigms}).  At the \emph{algorithm} level we then formalize the concrete training rule (Eq.~\ref{eq:il_amortized}) that this paradigm yields when applied to the Inverter's role specification. Acting to achieve a goal admits a direct formulation as action optimization: given state $s_0$ and a task conditioning $c$ (the framework's per-chunk subgoal $c_k$ at Level~1, the overall goal $G$ at Level~2), find
\begin{equation}
\begin{aligned}
a_{1:T}^* \;&=\; \arg\min_{a_{1:T}} \; \mathcal{J}(s_{0:T},\, a_{1:T},\, c) \\
&\text{s.t.}\;\; \textcolor{black}{s_{1:T} = f(s_0,\, a_{1:T})},
\end{aligned}
\label{eq:inverse_learning}
\end{equation}
a discrete-time Bolza problem of classical \acs{OC}~\citep{pontryagin1962mathematical,bolza1913vorlesungen} parameterized by $(s_0, c)$, \textcolor{black}{in chunked form (equivalent to the per-step recursion $s_t = f(s_{t-1}, a_t)$),} with $\mathcal{J}$ any differentiable combination of terminal cost, running reward, and action regularizer.  \textcolor{black}{Each term in $\mathcal{J}$ may be a \textbf{closed-form} function of the predicted state-action trajectory (e.g., an analytic terminal cost, a support-region indicator, an action regularizer), an \textbf{additional learned reward model} -- a separate differentiable critic, or dedicated reward heads of the \acs{FoM} itself -- or any sum of such terms.  Closed-form and learned components \emph{compose freely within a single $\mathcal{J}$}; the only requirement is differentiability in $a_{1:T}$.  This compositional freedom lets a single Inverter target arbitrary differentiable multi-axis specifications.}  Pontryagin's costate equation -- backpropagation through $f$, predating its neural use -- solves Eq.~\eqref{eq:inverse_learning} iteratively per $(s_0, c)$ query, with no learned amortization across queries.  In contrast, the value-recursive Bellman equation $V^*(s) = \max_a [r(s,a) + \gamma V^*(f(s,a))]$~\citep{bellman1957dynamic} does not produce action sequences, but one reactive action at one step at a time.

\paragraph{Amortizing the planner.}
\textcolor{black}{\acs{IL} amortizes classical per-query trajectory optimization into a two-component \emph{learned solver}: (1)~a \textbf{\acf{FoM}} $f_\theta$\textcolor{black}{, the learned approximation of $f$ in Eq.~\ref{eq:inverse_learning}, trained as a chunked sequence model with chunk length $L \leq T$ and stitched across chunks when $L < T$ (per-task $L$ in Apps.~\ref{app:maze2d_details}, \ref{app:antmaze_details})}, and (2)~an \textbf{\acf{IM}} $g_\phi(s_0, c) \to a_{1:T}$ that emits the full $T$-step action sequence in one feedforward pass. In the work presented here, the two are trained sequentially: first $f_\theta$ \emph{supervised} on offline $(s_t, a_t, s_{t+1})$ transitions (or supplied analytically when the physics is known, e.g.,\ the Lindbladian in Sec.~\ref{sec:quantum}); then, with $f_\theta$ frozen, $g_\phi$ is trained by backpropagating a Bolza objective through it (joint $(f_\theta, g_\phi)$ training\textcolor{black}{, as well as joint training across hierarchy levels,} is equally natural and is discussed in Sec.~\ref{sec:stacked_antman}):}
\begin{equation}
\min_\phi \;\mathbb{E}_{(s_0, c)}\!\left[\,\mathcal{J}\!\Big(f_\theta^{(1:T)}\!\big(s_0,\; g_\phi(s_0, c)\big),\;\; g_\phi(s_0, c),\;\; c\Big)\,\right]
\label{eq:il_amortized}
\end{equation}
Jordan \& Rumelhart~\citep{jordan1992forward} introduced this training pattern for single-step distal control; \acs{IL} extends it to multi-step, $H{>}1$ level hierarchical planning and control through learned forward and inverse models for embodied control (Secs.~\ref{sec:single_low_level}--\ref{sec:stacked_antman}). At $T{=}1$ and $H{=}1$ the Bolza objective collapses to a terminal cost and the recipe reduces to a non-hierarchical Jordan--Rumelhart-style distal teacher; the framework's richness lives at $T{>}1$ and $H{>}1$, where running cost and global trajectory structure enter $\mathcal{J}$, sequence-level optimization over the whole chunk becomes possible, hierarchy allows solving more complex tasks, and chunked open-loop emission yields per-episode inference compute time reductions (see Sec.~\ref{sec:discussion} on the empirical consequences). 

We refer to the $T{>}1$ regime of this paradigm as \emph{\acf{ISL}}. All experiments and architectures in this paper operate in the \acs{ISL} regime; we retain the broader ``\acs{IL}'' term for both the $T{=}1$ and $T{>}1$ cases.  The practical advantage of the ISL regime is the gradient structure: where \acs{RL} attributes a scalar reward across all $d_a \!\times\! T$ action dimensions through estimators like \acf{PG} and \acf{TD}, \acs{ISL} backpropagates through $f_\theta$ to deliver an exact gradient $\partial \mathcal{J} / \partial a_{t,i}$ at every action dimension $i$ and every timestep $t$ across the whole $T$-step sequence.

Table~\ref{tab:paradigms} delineates \acs{IL} from supervised, reinforcement, and imitation learning, indexed by the training signal each paradigm relies on.  Inverters combine \acs{RL}'s training-time amortization with the multi-step, sequence-level scope of optimal control, without iterative deployment-time optimization.  Unlike imitation learning, they learn predictable task structure (e.g., physics) through a \acs{FoM} and invert it, rather than cloning behavior.

\begin{table}[!htbp]
\centering
\caption{Inverse learning delineated from the three major established learning paradigms in their typical form: supervised, reinforcement, and imitation of which \acf{BC} is the supervised special case.  The defining axis -- and the key requirement of each paradigm -- is the nature of the training signal.}
\label{tab:paradigms}
\small
\begin{tabular}{@{}lcccc@{}}
\toprule
& \textbf{Supervised} & \textbf{Reinforcement} & \textbf{Imitation} & \textbf{Inverse} \\
\midrule
Training signal & Label $y$ & Reward (scalar) & Expert action $a^*$ & Action gradient $\partial \mathcal{J} / \partial a$ \\
Gradient quality & Exact & Estimated & Exact & Exact \\
Temporal scope & One- or multi-step & One step (recursive) & One- or multi-step & One- or multi-step\\
Output & $\hat y$ (prediction) & $a_t$ (reaction) & $a_t$ or $a_{1:T}$ (cloned) & $a_{1:T}$ (actions) \\
Objective & Differentiable loss & Additive reward & Fixed to expert & Differentiable (Bolza) \\
\bottomrule
\end{tabular}
\end{table}

\section{Related work}
\label{sec:related_work}

\paragraph{\textcolor{black}{\acf{DL} for inverse problems.}}
\textcolor{black}{\acs{IL} belongs to the broader research field of amortizing inverse problems with neural networks~\citep{arridge2019inverse,ongie2020inverse}, in which an \acs{NN} is trained to map measurements (or, more generally, conditioning) to a solution of a forward operator equation\textcolor{black}{; \acs{IL} sits specifically in the \emph{self-supervised / measurement-only} branch of that taxonomy}, with backpropagation through the forward operator providing the training signal \textcolor{black}{(no ground-truth solution labels)}.  This tradition is most developed in imaging and physics: linear inverse problems (\acs{CT}/\acs{MRI} reconstruction, compressed sensing, super-resolution), unrolled iterative solvers (\acs{LISTA}-style)~\citep{gregor2010lista}, and more recently \textcolor{black}{non-amortized} diffusion-prior posterior sampling~\citep{chung2023dps}.  \acs{IL} specializes this pattern to embodied planning and control: the \textcolor{black}{(itself learned)} ``forward operator'' is, e.g., a dynamics model unrolled in time, the ``solution'' is a multi-step action sequence rather than a static signal, and the objective is a Bolza functional (terminal cost + running cost + regularizer).  In this view, \acs{IL} is the planning-and-control instance of the broader \acs{DL}-for-inverse-problems agenda\textcolor{black}{, extending it from analytical to \emph{learned} forward operators and from static signals to time-extended action sequences}.}

\paragraph{Amortized optimal control.}
\textcolor{black}{\acs{IL}} is an instance of \emph{amortized optimal control} that allows training a network to emit a multistep \acs{OC} solution in a single feedforward pass.  Related instances differ in either output shape or in whether deployment-time iteration is avoided: \textcolor{black}{explicit \acf{MPC}~\citep{bemporad2002explicit} offline-precomputes \acs{MPC} into a piecewise-affine state-feedback function for linear-quadratic problems (with modern \acs{NN}-approximate variants for nonlinear settings), but emits a single action per call in receding-horizon fashion;} \textcolor{black}{guided-policy-search / }\textcolor{black}{coupled} trajopt distillation into a policy~(\textcolor{black}{\citealp{levine2013gps}}; \citealp{mordatch2014combining}; \textcolor{black}{\citealp{carius2020mpcnet}}) \textcolor{black}{supervises policy training with} the trajectories of a separate, non-differentiable offline \acs{OC} solver; value-gradient backprop through learned dynamics~\citep{heess2015svg} differentiates the value through model rollouts but emits one action at a time; differentiable \textcolor{black}{optimization and }\acs{MPC} layers~(\textcolor{black}{\citealp{amos2017optnet}}; \citealp{amos2018differentiable_mpc}) embed a constrained \acs{QP} solver as a differentiable layer and so retain test-time iterative optimization.

\paragraph{Differentiable world-model control.}
Backpropagating a control objective through a learned \textcolor{black}{world model (the agent-in-environment composite, not just body kinematics)} has a long lineage~\citep{schmidhuber1990making,jordan1992forward,lecun2022path}\textcolor{black}{ -- Schmidhuber even explicitly proposed using the learned model to plan multi-step action sequences via simulated gradient descent, while flagging its high inference cost}; \textcolor{black}{concrete algorithmic} variants -- \acs{PILCO}~\citep{deisenroth2011pilco}, Dreamer~\citep{hafner2019dream}, Universal Planning Networks~\citep{srinivas2018upn}, \acs{TD-MPC2}~\citep{hansen2024tdmpc2} -- retain test-time iteration, value bootstrapping, or one-step outputs.  \textcolor{black}{\acs{IL}} differs in output shape and query cost as it allows inverse networks to emits the full $T$-step action sequence in one feedforward pass.

\paragraph{Trajectory-level sequence models \textcolor{black}{and iterative planners}.}
A parallel line treats the action sequence as the object: autoregressive sequence modeling~\citep{janner2021offline_tt}, return-conditioned \textcolor{black}{autoregressive policies}~\citep{chen2021decision}, iterative trajectory denoising~\citep{janner2022planning}, online tree search over a learned model~\citep{schrittwieser2020mastering}, test-time-iterative latent planners~\citep{sobal2025latent,bar2024navigation}, and---outside \acs{RL}---learned step-by-step motion planners such as \acs{MPNet}~\citep{qureshi2019mpnet}, which autoregressively emits next configurations from an imitation-trained network; broader hierarchical-planning agendas with learned predictive world models~\citep{lecun2022path} share these commitments.  \textcolor{black}{\acs{IL} collapses both step-by-step autoregression and test-time iterative optimization into a single feedforward emission of the $H$-step action chunk, folding trajectory optimization into training. Action Chunking Transformers~\citep{zhao2023act} and Diffusion Policy~\citep{chi2023diffusion} share our chunked output shape but are supervised on expert demonstrations, restricting data and objective to imitation.}

\paragraph{Hierarchical control and hybrid symbolic-continuous planning.}
Hierarchical \acs{RL} builds on the options framework~\citep{sutton1999options}, typically mixing level-specific mechanisms (e.g., high-level policy gradient over low-level actor-critic); hierarchical latent world models have been used for visual humanoid control but still rely on test-time planning~\citep{hansen2024puppeteer}.  \textcolor{black}{The amortization perspective extends beyond purely continuous problems: hybrid symbolic-continuous control is traditionally handled by iterative task-and-motion planning~\citep{garrett2021tamp}, mixed-integer \acs{MPC} over hybrid systems~\citep{marcucci2021mipmpc}, or signal-temporal-logic \acs{MPC} compiled to \acs{MILP}~\citep{raman2014stlmpc}.  Our AntMan Game Inverter (Sec.~\ref{sec:stacked_antman}) is in this hybrid regime but emits its symbolic plan in a single feedforward pass through a differentiable \acs{FoM}, positioning \acs{IL} as an amortized counterpart in the neurosymbolic \acs{OC} space currently dominated by iterative methods.}

\paragraph{\textcolor{black}{\acl{RL} of generative models via frozen learned critics.}}
\textcolor{black}{A recent line of work in generative modeling performs \acs{RL} of pretrained samplers against a learned reward model: DRaFT~\citep{clark2024draft}, and ReFL~\citep{xu2023refl} for image diffusion; DRAKES~\citep{wang2025drakes} for discrete diffusion via soft-token embeddings; and Adjoint Matching~\citep{domingoenrich2025adjoint}, which casts fine-tuning as stochastic optimal control over the denoising trajectory.  Following the classical Jordan \& Rumelhart distal-teacher pattern~\citep{jordan1992forward}, these methods replace the variance-prone gradient estimators of classical \acs{RL} with an \emph{exact} gradient obtained by backpropagating through a frozen, differentiable critic (Gumbel-relaxed or straight-through for discrete cases).  The objective, however, remains single-step ($T{=}1$) reward maximization on the generator's terminal output (one image / molecule / text sample) as \acs{RL} methods rather than \acs{IL} in the sense used here.}

\paragraph{\textcolor{black}{\acf{IRL}}}
\textcolor{black}{ addresses a different inverse problem: recovering the reward function from expert demonstrations~\citep{ng2000irl}.  \acs{IL} in the sense of this paper instead amortizes the goal$\to$action inversion through a forward model, in the Jordan \& Rumelhart distal-teacher lineage~\citep{jordan1992forward} extended to $T{>}1$.  The two paradigms are complementary and could in principle compose -- e.g., an \acs{IRL}-recovered reward used as the terminal cost in \acs{IL}'s Bolza objective.}

% ============================================================================
\section{Planning and control with Inverters: properties, performance, failure modes}
\label{sec:experiments}
% ============================================================================

We organize the experiments in three steps of increasing structural complexity: a single Level 1 Inverter (\texttt{maze2d-umaze-v1}; Sec.~\ref{sec:single_low_level}), a Level 1 Inverter coupled with a (simple) algorithmic Path Inverter at Level 2 (larger \texttt{maze2d} layouts and all six \texttt{antmaze-v2} variants; Secs.~\ref{sec:inverter_plus_sequencer}--\ref{sec:antmaze_eval}), and finally the full $n\!=\!2$ hierarchical Setup with two paired learned Inverters (AntMan on \texttt{antmaze-large-diverse-v2}; Sec.~\ref{sec:stacked_antman}).

\subsection{Single Motor Inverter (\texttt{maze2d-umaze}): best \acs{D4RL} at less inference compute \textcolor{black}{time}}
\label{sec:single_low_level}

\begin{figure}[!htbp]
\centering
\includegraphics[width=1.0\linewidth]{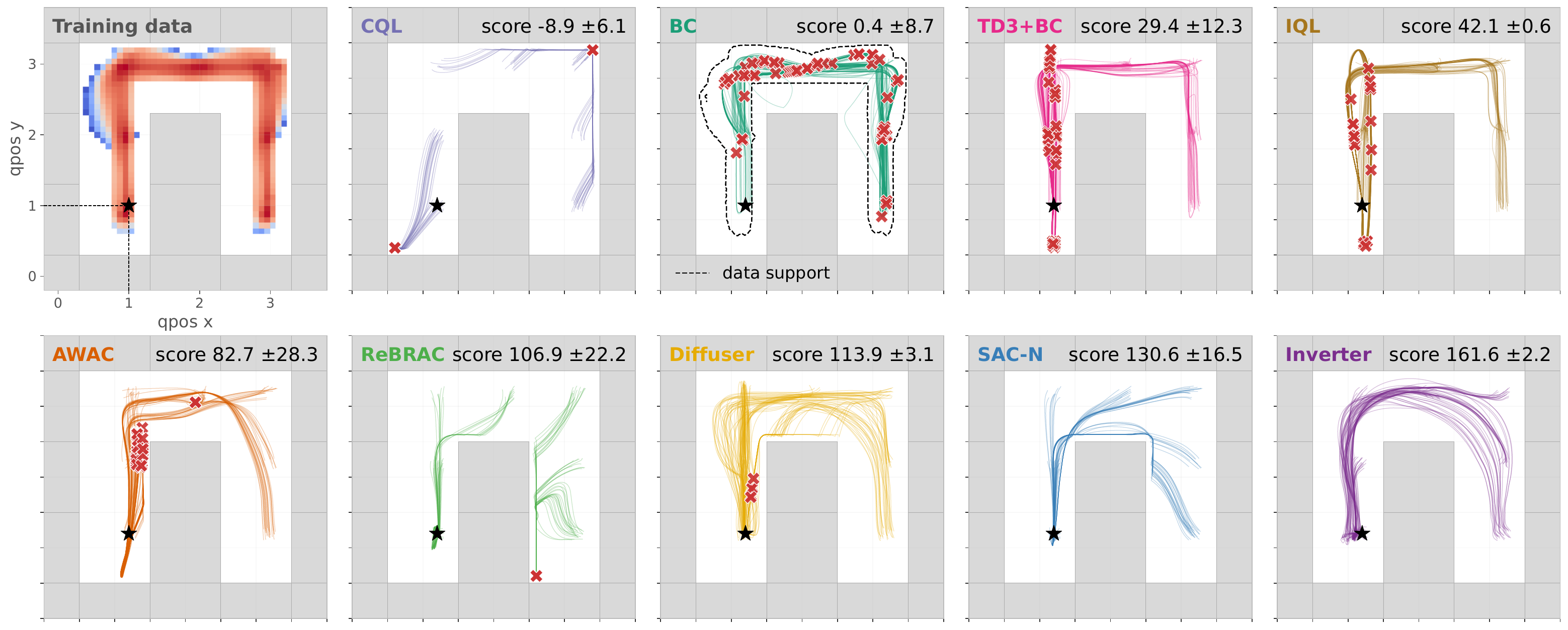}
\caption{\textbf{Trajectory comparison on \texttt{maze2d-umaze-v1}.}  Top-left: training data heatmap (blue to red: low to high density) showing the U-shaped corridor coverage.  Remaining panels: 100 evaluation trajectories per method, ordered by \acs{D4RL} score. ~$\bigstar$~=~goal; red~$\times$~=~timeout (episode did not reach the goal).  The Inverter~(ours) produces smooth, direct paths that are not restricted to the angular data support.  Baselines trained with \acs{CORL}~\citep{tarasov2024corl}.}
\label{fig:trajectories}
\end{figure}

We evaluate \texttt{maze2d-umaze-v1} with a frozen causal Transformer \acs{FoM} paired with a causal Transformer \textcolor{black}{\acs{IM}} that emits a $128$-step action sequence from the current state and goal in one forward pass (architectures, dataset, baselines: App.~\ref{app:maze2d_details}).

% \label{sec:maze2d_perf}

\textbf{Performance and inference \textcolor{black}{compute time}:} Table~\ref{tab:maze2d_headline} summarizes \acs{D4RL} and per-episode compute \textcolor{black}{time} across all three \texttt{maze2d} variants.  \textcolor{black}{Throughout, ``inference compute time'' refers to per-episode wall time on a single device at batch\,1; in a launch-overhead-limited small-model regime this is the deployment-relevant metric, distinct from FLOPs (App.~\ref{app:timing_methodology}).}  On \texttt{umaze-v1} a one-shot $K\!=\!128$ plan delivers $\mathbf{161.6\!\pm\!2.2}$ in just $3$ \acs{NN} forward passes (one to reach the goal, two to stay at the goal until the end of the fixed 300 time step evaluation window) and $\mathbf{11.4}$\,ms total per episode -- $\mathbf{37\times}$ less than Diffuser, $\mathbf{47\times}$ less than \acs{SAC-N}, and estimated nearly three orders of magnitude less than DecisionLLM~\citep{lv2026decisionllm}.  The advantage is \emph{not} a faster single forward pass (the Inverter transformer sits at the same $\sim\!2$\,ms-per-pass floor as the baseline \acp{MLP}) but emitting a full $128$-step action sequence per forward, not just one at a time, reducing per-episode passes by $30$--$100\times$ (App.~\ref{app:timing_methodology}).

\begin{table}[!htbp]
\centering
\small
\caption{\textbf{Maze2d summary: Motor Inverter vs.\ best \acs{D4RL} baseline per maze and fastest PyTorch baseline (\acs{BC}-10\%).}  \textbf{Top}: \acs{D4RL} score (mean$\,\pm\,$std over 100 episodes with 4 seeds).  \textbf{Bottom}: \acs{NN} forward passes and total wall time per episode (\acs{NN} passes plus algorithmic overhead, single A40 GPU, PyTorch, batch\,1, CUDA-synced).  Best-per-column bolded; the Inverter row reports the $K\!=\!128$ one-shot configuration on umaze (Tab.~\ref{tab:maze2d_perf}, App.~\ref{app:perf_tables}) and $K\!=\!16$ on medium/large (Tabs.~\ref{tab:maze2d_perf_medium},~\ref{tab:maze2d_perf_large}).  Full per-method tables (10+ baselines, including JAX+JIT \acs{ReBRAC} and Diffuser): App.~\ref{app:perf_tables}.}
\label{tab:maze2d_headline}
\begin{tabular}{@{}lrrr@{}}
\toprule
& \texttt{umaze-v1} & \texttt{medium-v1} & \texttt{large-v1} \\
\midrule
\multicolumn{4}{l}{\emph{\acs{D4RL} score}} \\
Best \acs{D4RL} baseline (per maze) & \acs{SAC-N}: $130.6\,{\scriptstyle\pm\,16.5}$ & Diffuser: $130.1\,{\scriptstyle\pm\,22.7}$ & \acs{AWAC}: $209.1\,{\scriptstyle\pm\,8.2}$ \\
\textbf{Inverter (ours)} & $\mathbf{161.6\,{\scriptstyle\pm\,2.2}}$ & $\mathbf{166.8\,{\scriptstyle\pm\,1.2}}$ & $\mathbf{220.7\,{\scriptstyle\pm\,0.2}}$ \\
\midrule
\multicolumn{4}{l}{\emph{\# \acs{NN} forward passes per episode / total wall time per episode (ms)}} \\
\acs{BC}-10\% (fastest PyTorch) & $300$ / $374.9$ & $600$ / $749.8$ & $800$ / $999.7$ \\
\textbf{Inverter (ours)} & $\mathbf{3}$ / $\mathbf{11.4}$ & $\mathbf{38}$ / $\mathbf{72.9}$ & $\mathbf{50}$ / $\mathbf{93.7}$ \\
\bottomrule
\end{tabular}
\end{table}

%\subsubsection{Approaching an analytic time-optimal control outside the data support}
%\label{sec:maze2d_gestalt}

\begin{figure}[!htbp]
\centering
\includegraphics[width=1.0\linewidth]{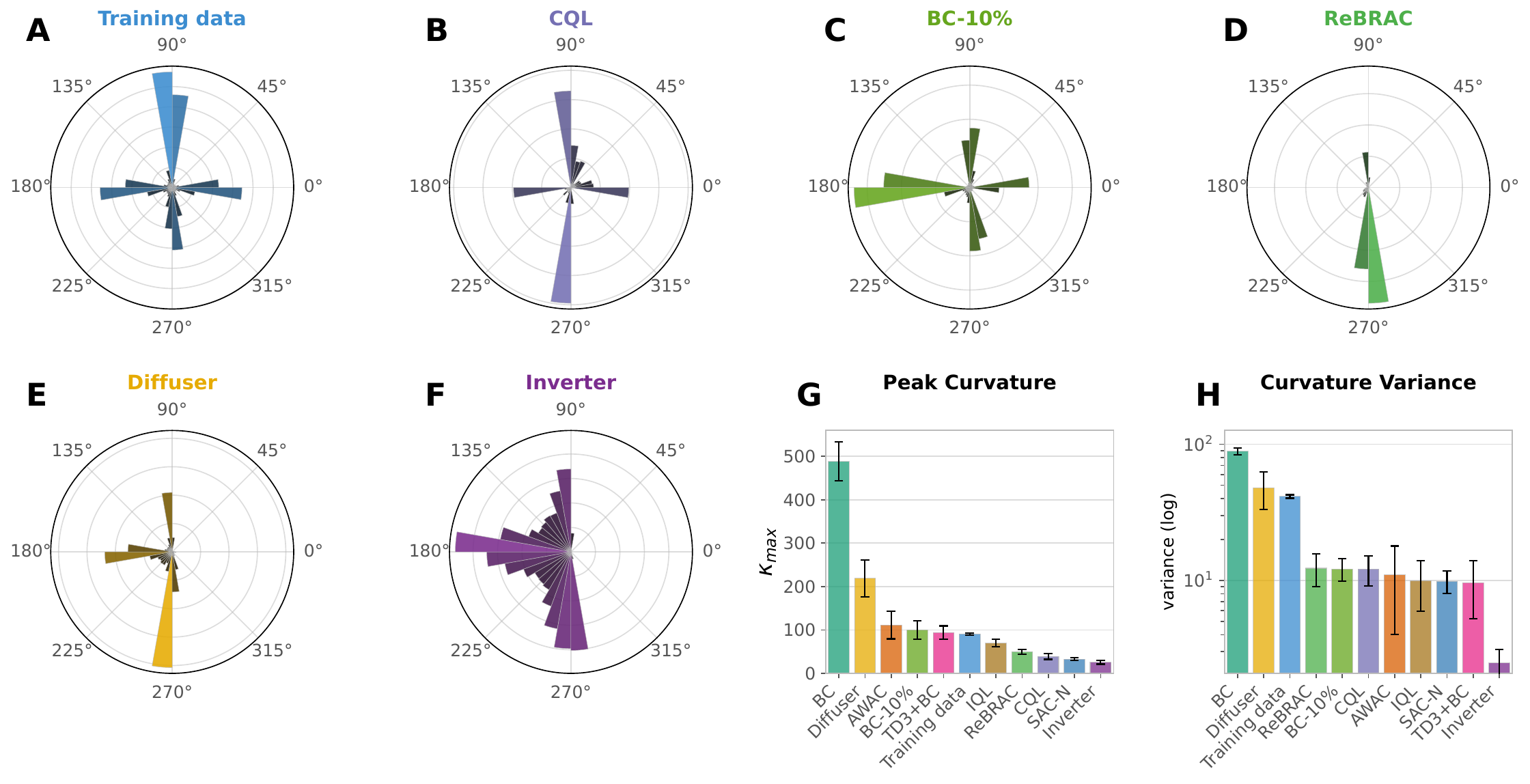}
\caption{\textbf{Directional movement spectrum and \textcolor{black}{sequence-level optimization}.}
Panels~A--F show polar histograms of \textcolor{black}{$T{=}16$-step chunk-displacement} directions for the training data and a representative subset of methods\textcolor{black}{; angles indicate $0^\circ\!=\!+x$, $90^\circ\!=\!+y$ in raw maze data coordinates, based on each method's per-episode rollout until the first step where the agent enters a goal-ball of radius $0.5$, or time out}.  Panel~G and ~H shows median peak curvature $\kappa_{\max}$, and median curvature variance (log scale) over the same trajectory data, respectively.  Whiskers: +/- \acf{SEM}. \textcolor{black}{The Inverter produces the lowest peak curvature and lowest curvature variance (Panels~G,~H). Similar to Diffusor, and unlike the other baseline methods, the inverter produces all of the major movement directions required to solve the task (up, down, and leftwards), but no movements with a dominating rightward component, which are not needed to reach the goal (Panel~A-E). These observations together quantitatively confirm the \emph{sequence-level optimization} that the smooth trajectories of Fig.~\ref{fig:trajectories} suggested qualitatively: optimizing over the \emph{entire} action sequence -- rather than per step -- yield smooth trajectories which allow to reach the goal faster.}}
\label{fig:directional_rose}
\end{figure}

\paragraph{Smooth, coherent trajectories.}  Figure~\ref{fig:trajectories} shows that the motor Inverter produces smooth, direct paths through the maze, while every baseline yields visibly noisier, more angular, or more constrained motion. Figure~\ref{fig:directional_rose} confirms this quantitatively via per-sequence displacement directions and curvature variance: the Inverter achieves the lowest peak curvature and, by a large margin, the lowest curvature variance, demonstrating the by-design ability of Inverters for more holistic sequence-level optimization. The \acs{IM}  achieves this not by mimicking any individual training trajectory, but by learning to invert the forward dynamics globally: gradients flow through the frozen \acs{FoM}  across the entire horizon, enabling the network to discover action sequences whose \emph{integrated} trajectory has favorable geometric properties that local, step-wise optimization cannot guarantee.

\begin{figure}[!htbp]
\centering
\includegraphics[width=1.0\linewidth]{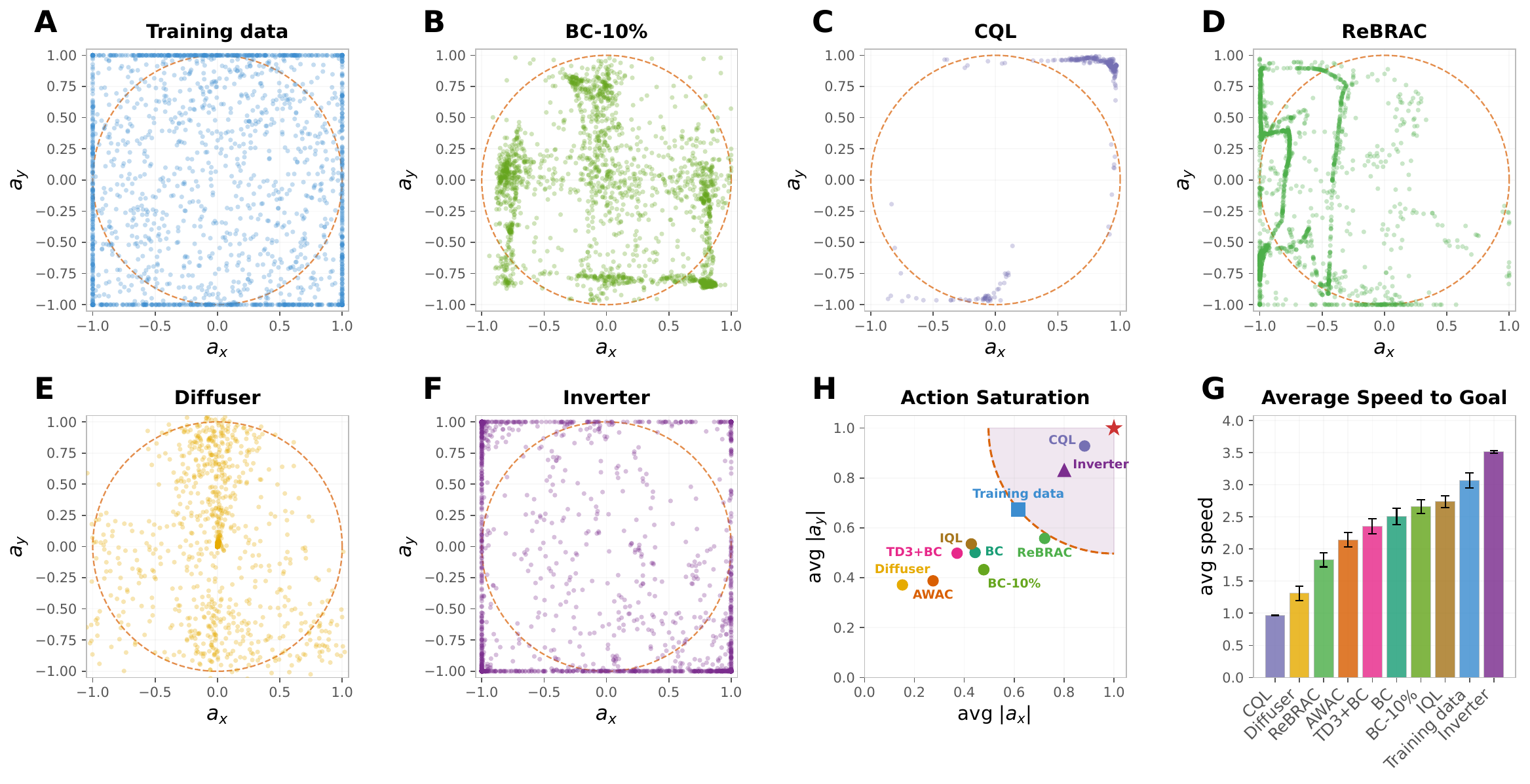}
\caption{\textbf{Action-space structure \textcolor{black}{and control optimality.}} Per-step action scatter plots $(a_x,a_y)$ for the training data and a representative subset of methods (Panels~A--F).  Note that \acs{BC}-10\%~\citep{kumar2022should} lands \emph{below} the training data on action saturation (Panel~H), even though it is trained to imitate it: a unimodal Gaussian policy head contracts the action distribution toward the interior -- the multimodal-action \acs{BC} failure mode~\citep{florence2021implicit,chi2023diffusion}. Panel~H: \textcolor[HTML]{CC3333}{$\bigstar$}~at $(1,1)$ marks the analytically optimal action saturation -- pure bang-bang, derived from Pontryagin's maximum principle for the damped point-mass minimum-time problem (App.~\ref{app:damped_bangbang}). Panel~H: average speed to goal, using the same trajectory data as in Figure~\ref{fig:trajectories}. \textcolor{black}{The Inverter shows both joint signatures of the analytic time-optimal control: high action saturation, beyond the training data -- and high average speed to goal (Panel~G). Note that \acs{CQL} outputs highly saturated actions, but missing action switching times, it fails the task at a negative \acs{D4RL} score, showing that action saturation alone is a necessary but not sufficient condition for approaching control optimality.}}
\label{fig:action_scatter}
\end{figure}

\paragraph{Approaching the analytic optimum outside the data support.}  Under viscous damping with a bounded action box, the minimum-time control problem on \texttt{maze2d-umaze-v1} is approximately bang-bang (App.~\ref{app:damped_bangbang}): the time-optimal policy saturates each actuator to $\pm u_{\max}$ -- an optimum that lies \emph{outside} the training data's action-space support.  Figure~\ref{fig:action_scatter} reveals this structure directly: every baseline populates the interior with unsaturated actions, while the Inverter concentrates mass at the four edges and corners.  \textcolor{black}{Its action saturation substantially exceeds that of the training data -- a necessary structural component of bang-bang time-optimal control. This is not sufficient on its own (a misdirected bang-bang policy can saturate without speed gain, as shown by \acs{CQL}), but the Inverter \emph{also} reaches the highest average speed and consecutively the highest \acs{D4RL} score (Table~\ref{tab:maze2d_headline}); together an empirical signature of approach to the analytic optimum.}  This illustrates that the Inverter uses learned physics through gradients flowing through the frozen \acs{FoM}, rather than to clone dataset actions, leaving the Inverter structurally free to leave the data's support and to find better solutions beyond it.

\subsection{Motor Inverter $+$ simple algorithmic Path Inverter on \texttt{maze2d-medium}/\texttt{large}: confirming best \acs{D4RL} at substantially less compute \textcolor{black}{time}}
\label{sec:inverter_plus_sequencer}

On \texttt{maze2d-medium-v1} and \texttt{maze2d-large-v1}, the dataset's longest corridor paths exceed the Level 1 Inverter's $128$-step horizon, so a single feedforward plan is insufficient.  At the same time, waypoint routing through the maze is a discrete shortest-path sub-problem on which a full-fledged learned Planning Inverter would be overkill. We therefore add a simple \textbf{algorithmic Path Inverter (\acs{BFS} built exclusively on the offline-data density)} at Level 2: it builds a free-space occupancy grid from the training-data density alone, runs a 4-connected \acs{BFS}, and emits one sub-goal per corridor elbow; the Level 1 Inverter then executes chunk by chunk toward the current sub-goal.  No maze geometry, simulator, or \acs{FoM} call enters the planner.  App.~\ref{app:waypoint_planner} gives the full construction; Fig.~\ref{fig:waypoint_sequencer} shows a representative plan overlaid on the data heatmap.

Main numbers for medium and large at $K\!=\!16$ (highest \acs{D4RL} across our $K$-sweep on medium and large; Suppl.~Tab.~\ref{tab:maze2d_ksweep}) appear in Table~\ref{tab:maze2d_headline}; full per-method breakdowns are in App.~\ref{app:perf_tables}.  The stacked Inverter reaches $\mathbf{166.8\!\pm\!1.2}$ \acs{D4RL} on medium and $\mathbf{220.7\!\pm\!0.2}$ on large at $\mathbf{100/100}$ success across $4$ seed \textcolor{black}{\acp{IM}} -- strictly ahead of every \acs{CORL} baseline and Diffuser -- spending $\mathbf{72.9}$ and $\mathbf{93.7}$\,ms of \acs{NN} compute per episode, roughly $\mathbf{10}$--$\mathbf{30\times}$ faster than step-wise offline-\acs{RL} baselines and $\mathbf{39}$--$\mathbf{51\times}$ faster than Diffuser (DecisionLLM reference for this task not available).  The only compute-competitive method is \acs{ReBRAC}$^{\dagger}$ (JAX+JIT, $\sim\!0.2$\,ms/pass), which still lags the Inverter on \acs{D4RL} by $63$ points on medium and $137$ on large.  Sweeping $K\!\in\!\{16, 32, 64, 128, 256\}$ exposes a clean speed--accuracy trade-off (App.~\ref{app:ksweep}): smaller $K$ gives higher \acs{D4RL} at higher per-episode wall-time, and accuracy collapses once $K$ crosses the \textcolor{black}{\acs{IM}'s} $128$-step training horizon -- interestingly reminiscent of Fitts' law and submovement / iterative-correction motor psychophysics~\citep{fitts1954information,crossman1963feedback,meyer1988optimality}.

\subsection{\texttt{antmaze-v2}: Scaling to locomotion and first encounter with \acs{FoM} hacking}
\label{sec:antmaze_eval}

\begin{table}[!t]
\centering
\small
\caption{\textbf{Antmaze summary: Inverter vs.\ strongest \acs{CORL} \acs{D4RL} competitor (\acs{ReBRAC}) and fastest PyTorch baseline (\acs{BC}-10\%).}  \textbf{Top}: \acs{D4RL} score (\%) over 100 episodes on each \texttt{antmaze-v2} variant; both \acs{ReBRAC} and Inverter are mean$\,\pm\,$std over $4$ seeds.  \textbf{Bottom}: ms per environment step (single A40 GPU, PyTorch, batch\,1, cuda-synced).  Best-per-column bolded. Full 10-baseline breakdown including JAX+JIT \acs{ReBRAC} timing: Tab.~\ref{tab:antmaze_perf}, App.~\ref{app:perf_tables}.}
\label{tab:antmaze_headline}
\begin{tabular}{@{}l@{\hspace{6pt}}rrrrrr@{}}
\toprule
& u-umaze & u-divrs & m-play & m-divrs & l-play & l-divrs \\
\midrule
\multicolumn{7}{l}{\emph{\acs{D4RL} score (\%)}} \\
\acs{ReBRAC} & $97.8\,{\scriptstyle\pm\,1.5}$ & $83.5\,{\scriptstyle\pm\,7.0}$ & $\mathbf{89.5\,{\scriptstyle\pm\,3.4}}$ & $83.5\,{\scriptstyle\pm\,8.2}$ & $52.2\,{\scriptstyle\pm\,29.0}$ & $64.0\,{\scriptstyle\pm\,5.4}$ \\
\textbf{Inverter (ours)} & $\mathbf{99.5\,{\scriptstyle\pm\,0.9}}$ & $\mathbf{99.8\,{\scriptstyle\pm\,0.4}}$ & $87.8\,{\scriptstyle\pm\,8.0}$ & $\mathbf{96.5\,{\scriptstyle\pm\,5.0}}$ & $\mathbf{93.0\,{\scriptstyle\pm\,2.6}}$ & $\mathbf{94.0\,{\scriptstyle\pm\,5.0}}$ \\
\midrule
\multicolumn{7}{l}{\emph{ms / env step}} \\
\acs{BC}-10\% (fastest PyTorch) & $1.25$ & $1.25$ & $1.25$ & $1.25$ & $1.25$ & $1.25$ \\
\textbf{Inverter (ours)} & $\mathbf{0.14}$ & $\mathbf{0.14}$ & $\mathbf{0.13}$ & $\mathbf{0.13}$ & $\mathbf{0.13}$ & $\mathbf{0.13}$ \\
\bottomrule
\end{tabular}
\end{table}

We find that the same setup (Level 1 Locomotion Inverter + simple algorithmic Path Inverter at Level 2) successfully transfers to all six \texttt{antmaze-v2} variants -- 29-dim \acf{MuJoCo} ant locomotion with 8-dim torque actions on the same topologies plus play/diverse goal distributions -- and matches within error bars or improves the strongest \acs{CORL} baseline (\acs{ReBRAC}) on every variant, with the gap widening on the large mazes (Tab.~\ref{tab:antmaze_headline} and App.~\ref{app:perf_tables}; for comparison to published model-based offline \acs{RL} baselines see App.~\ref{app:mbrl}).  In compute, the Locomotion Inverter emits a $16$-step chunk per call at $\sim\!2$\,ms, so the total per-env-step cost ($\sim\!0.13$--$0.14$\,ms on every variant) amortizes a small number of chunk-level passes over the $\sim\!600$--$700$ steps to the goal -- $\sim\!9\times$ faster per step than the fastest PyTorch baseline (\acs{BC}-10\%) on every antmaze variant.  

One important new observation from antmaze suggested that a plain task-reward Locomotion Inverter on antmaze may be \emph{forward-model hackable} -- while the \acs{FoM} predicts successful task completion, in simulation the ant may tip, jam, or fall  -- a failure we suspected to be related to the fact that the offline training data contains relatively few failure cases and the \acs{FoM} is therefore uncalibrated off-support.  We compensated with two additive, per-time-step auxiliary losses -- a \acs{BC} action-fidelity anchor and a body-yaw regularizer -- which compose linearly with the task gradient (ablations in App.~\ref{app:antmaze_details}, Suppl.~Tab.~\ref{tab:antmaze_ablation}).  These additions work in practice but are conceptually unsatisfactory: anchoring toward dataset actions partially compromises \acs{IL}'s defining property of being \acs{FoM}-gradient-driven rather than data-anchored.  These observations motivated designing the new AntMan task (Sec.~\ref{sec:stacked_antman}), which lets us (i) control the offline training data ourselves and therefore study \acs{FoM} hacking and its mitigations systematically, and (ii) demonstrate a full $n\!=\!2$ Hierarchical Inverter -- two trained Inverters, one for planning and one for control.

\begin{figure*}[!htbp]
\centering
\includegraphics[width=0.82\textwidth]{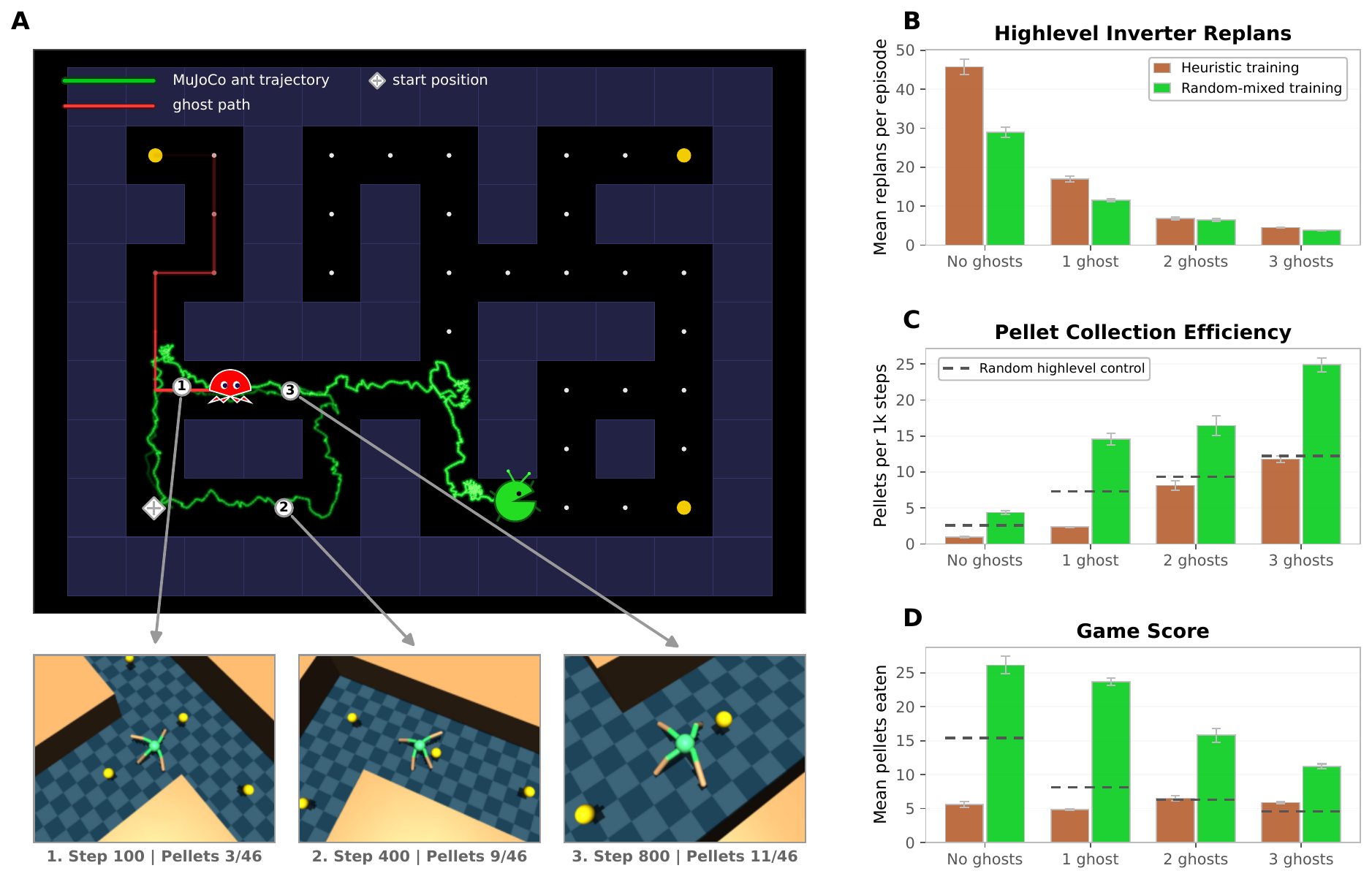}
\caption{\textbf{Simultaneous planning and control with a coupled Game and Locomotion Inverter on the AntMan task.} Panel~A shows a representative rollout in the \acs{MuJoCo} maze with an abstract rendering of the game situation and 3 snapshots showing the underlying motion of the \acs{MuJoCo} Ant through the maze. Successful planning and control was achieved when training on random training data, but not on training data only reflecting a heuristic game controller (Panels~B--D).  Metrics are averaged over 100 evaluation games (25 each for no, 1, 2, and 3 ghosts).  Only random training yields above-chance pellet-collection efficiency and pellet counts at every difficulty, while replanning is comparable or lower.}
\label{fig:antman_eval}
\end{figure*}

\begin{figure*}[!htbp]
\centering
\includegraphics[width=0.72\textwidth]{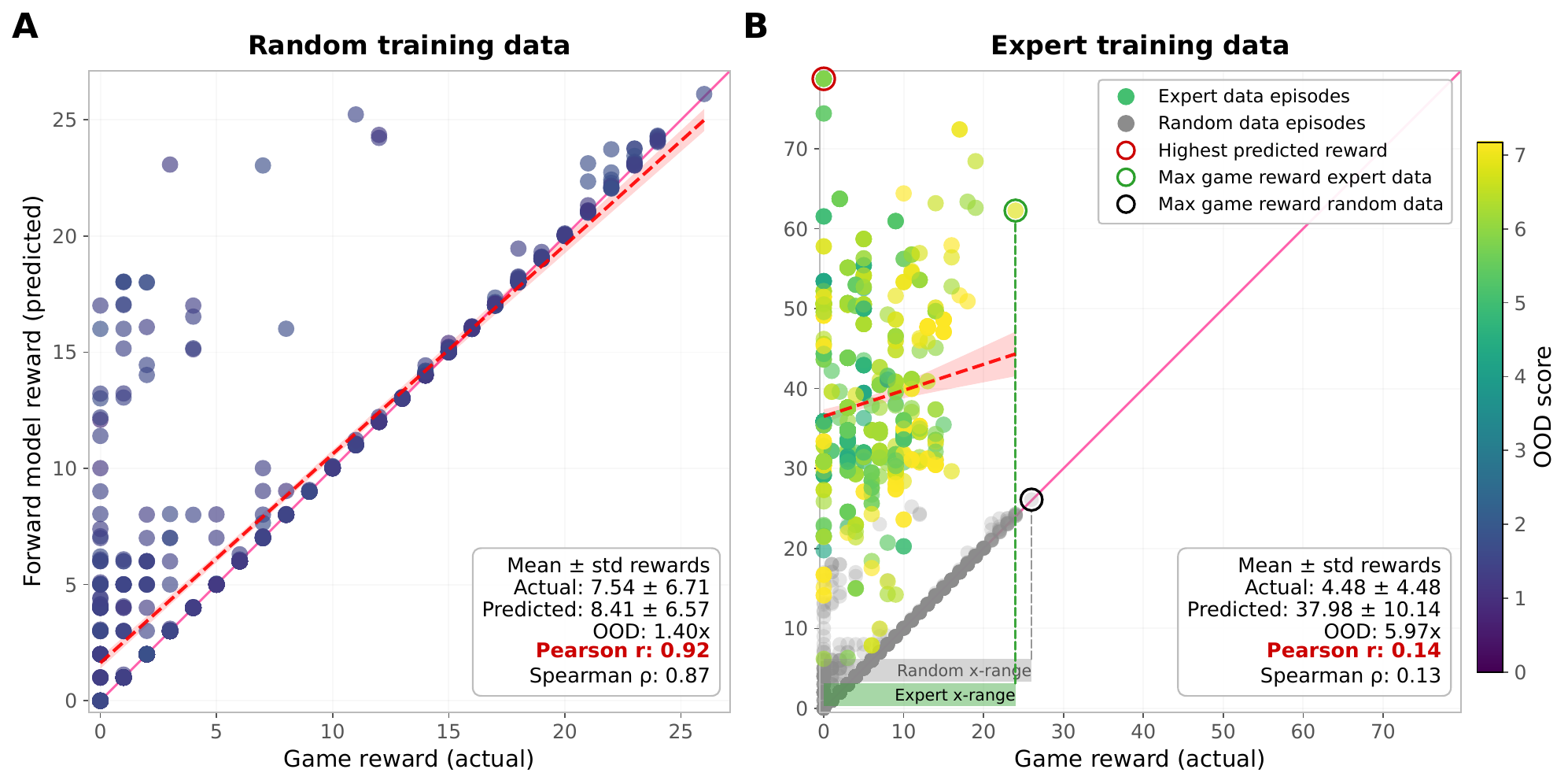}
\caption{\textbf{Random training data yields calibrated \acs{FoM} rewards; narrow expert data induces \acs{FoM} hacking.}  Each panel compares \acs{FoM}-predicted reward to realized game reward over 1000 sampled start states for the final high-level \textcolor{black}{\acs{IM}} checkpoints; point color encodes the \acf{OOD} score $\mathrm{OOD}(x) = d_1(x,\mathcal{T})\,/\,\widetilde d_1(\mathcal{T})$, where $d_1(x,\mathcal{T}) = \min_{y\in\mathcal{T}}\|x-y\|$ is each sampled start state $x$'s nearest-neighbor distance to the training set $\mathcal{T}$, and $\widetilde d_1(\mathcal{T}) = \operatorname{median}_{x'\in\mathcal{T}}\,\min_{y\in\mathcal{T}\setminus\{x'\}}\|x'-y\|$ is the typical nearest-neighbor distance between training points themselves -- so \acs{OOD}~$=$~1 means the sample is no further from the training set than a typical training point is from its own nearest neighbor, and \acs{OOD}~$>$~1 indicates an increasingly off-support sample, the pink diagonal marks perfect calibration, and dashed red lines show linear fits with 95\% confidence bands.  Panel~A (random training data) stays near the training support, has low \acs{OOD} values, and shows strong correlation between \acs{FoM} reward and actual game reward.  Panel~B (expert training data, without additional stabilizers such as a \acs{BC}-style loss) shows the opposite failure mode: the Inverter generates highly \acs{OOD} plans and the \acs{FoM} assigns them large rewards that are no longer predictive of actual outcomes.  The green and gray x-axis projections in Panel~B show the realized game-reward ranges for expert-data and random-data plans, with dashed connectors and circles marking the maximum game-reward sample from each group; these ranges are very similar.  Thus the failure is not that expert-data training removes high-game-reward plans from the sample, but that \acs{FoM} reward becomes decoupled from actual reward. For example, the highest \acs{FoM}-reward sample in Panel~B (red circle) achieves zero game reward.}
\label{fig:antman_fm_hacking}
\end{figure*}

\subsection{Hierarchical Inverse Learning for planning and control}
\label{sec:stacked_antman}

To study a fully learned $n\!=\!2$ Hierarchical Inverter, we designed the new AntMan task, which requires pure offline learning of both low-level locomotion control of the \acs{MuJoCo} Ant and a Pac-Man-style game embodied through the same Ant placed in the \texttt{antmaze-large-v2} setup. 

We solve this challenge with a \emph{Level 1 Locomotion Inverter} which controls the Ant in \texttt{antmaze-large-diverse-v2}, while a paired \emph{Level 2 Game Inverter} controls the Locomotion Inverter by outputting $32$-step waypoint directions from a 58-dim game state (ant pose, ghost states, pellet map, mode indicators). A causal-transformer \acs{FoM} is trained on simulated AntMan games; the Game Inverter is then trained purely through \acs{FoM} reward to collect as many pellets as possible.  \textcolor{black}{Structurally, the Game \acs{FoM} instantiates the \acs{FoM}-with-reward-heads option introduced in Sec.~\ref{sec:inverse_learning}: a dynamics head for ghost positions plus per-step reward heads for pellet-eaten and alive probabilities.  The Game Inverter's Bolza loss backpropagates through the reward heads, $\mathcal{L} = -\mathbb{E}\big[\sum_t \sigma(\ell_{\text{pel},t})\,\sigma(\ell_{\text{alive},t})\big]$ over the $H{=}32$ chunk.  In our $n{=}2$ AntMan stack, the Bolza loss therefore flows through next-state dynamics at the low level and through learned reward heads at the high level -- two instances of the same \acs{FoM}\textcolor{black}/{\acs{IM}} building block.}  To isolate the training-distribution effect, two matched \acs{FoM}/\textcolor{black}{\acs{IM}} pairs differ only in the offline training data generating behavior policy: (i) heuristic (chases pellets, avoids ghosts) and (ii) random (uniform random neighbor-cell moves).  

\paragraph{End-to-end hierarchical Inverse Learning for planning and control.}
The AntMan task demonstrates that a full $n\!=\!2$ Hierarchical Inverter, trained using \acs{FoM}-predicted game reward, successfully plays AntMan games well above chance -- proof of principle for increasing hierarchical depth and for extending the framework from continuous physics to symbolic/discrete planning. \textcolor{black}{Here we train each level separately, but the Inverter stack is end-to-end differentiable: jointly optimizing $(g_\phi^{(1)}, g_\phi^{(2)})$ -- propagating Level-2 task loss through Level-1, and conversely allowing Level-1 execution constraints to reshape Level-2 waypoint emission -- is a natural extension for regimes in which optimal plans depend on what the body can execute, or optimal motor commands depend on the upcoming plan.}

\paragraph{Forward-model hacking and how to mitigate it.}
A second observation reverses the offline-learning intuition that more competent demonstrations make better training data (Fig.~\ref{fig:antman_eval}).  With heuristic training data the \acs{FoM} is severely hacked -- predicted $37.98$ pellets vs.\ $4.48$ realized ($8.48\times$) -- and the Game Inverter drops to chance at 2--3 ghosts.  With random training data the \acs{FoM} is calibrated and the Inverter stays above chance across all difficulty levels.  The failure mechanism (Fig.~\ref{fig:antman_fm_hacking}; extended discussion in App.~\ref{app:antman_fm_hacking}) is that under expert-skewed data, the Inverter generates off-support plans for which the \acs{FoM}-predicted reward \emph{decorrelates} from the actual game outcome.  The implication: inverse-learning data has \emph{different} optimality criteria than \acs{RL}/imitation -- not demonstrator competence but \emph{safely diverse dynamics coverage}.  Our findings thus suggest that Inverters are particularly suited to settings where learning must proceed from random rather than expert data; the converse -- combining Inverters with imitation learning beyond plain \acs{BC} -- is an interesting open future direction.  \textcolor{black}{In this context, as another topic for future work, joint $(f_\theta, g_\phi)$ training (Sec.~\ref{sec:inverse_learning}) would be an adaptive alternative to broad-coverage data: it might keep the \acs{FoM} accurate on the Inverter's evolving action distribution as the two co-evolve.}

\begin{figure*}[!htbp]
\centering
\includegraphics[width=1.0\linewidth]{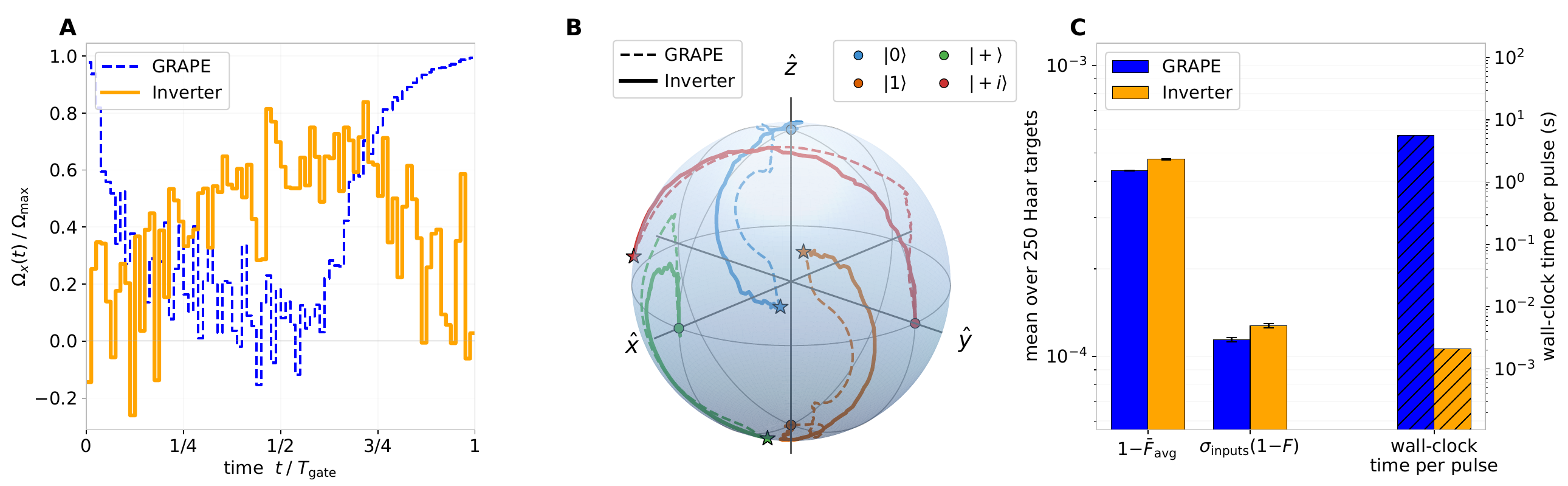}
\caption{\textbf{Single-shot quantum gate synthesis on a 3-level transmon under a known Lindbladian with a Pulse Inverter.}  \textbf{(A)}~Pulse $\Omega_x(t),\Omega_y(t)$ for one Haar-sampled target $U$; Inverter: solid orange, \acs{GRAPE}: dashed blue.  \textbf{(B)}~Bloch trajectories from four axis-aligned input states; stars mark $U|\psi\rangle$.  \textbf{(C)}~Over 250 Haar U(2) targets: $1{-}\bar F_{\mathrm{avg}}$, per-input fidelity uniformity, and median wall-clock per pulse on the same 128-core CPU -- $\sim\!2700{\times}$ speedup over \acs{GRAPE} (App.~\ref{app:quantum}).}
\label{fig:quantum_panel}
\end{figure*}

\subsection{Application example: Quantum gate synthesis}
\label{sec:quantum}

The standard iterative-numerical baseline for quantum gate synthesis, \acf{GRAPE}~\textcolor{black}{\citep{khaneja2005optimal}}, is normally run \emph{offline} to precompute a fixed gate library (\texttt{X}, \texttt{H}, \texttt{T}, \acs{CNOT}, \ldots).  An Inverter that emits pulses in one feedforward would be helpful for quantum computing applications where arbitrary unitaries are needed at $\mu$s timescales -- variational/parameterized algorithms, \acf{QEC}, or for adaptive feedback.

Quantum gate synthesis exemplifies a regime where the \acs{FoM} is available in closed form (the Lindblad channel of a given device). We train an Inverter $g_\phi(U){\to}\Omega$ to synthesize an $80$-slice $(\Omega_x,\Omega_y)$ pulse for arbitrary single-qubit targets $U\!\in\!\mathrm{U}(2)$ on a noisy 3-level transmon, by minimizing $1{-}\bar F_{\mathrm{avg}}$ through the analytic Lindblad channel as in Eq.~\ref{eq:il_amortized} (no \acs{GRAPE}-pulse supervision; full setup, baselines, per-target statistics, and the $\mathrm{U}(2)$/$\mathrm{SU}(2)$ sampling convention in App.~\ref{app:quantum}).  On 250 held-out Haar U(2) targets (Fig.~\ref{fig:quantum_panel}), the Inverter ties \acs{GRAPE} at the dissipation floor \textcolor{black}{-- the irreducible infidelity set by Lindbladian decoherence over the gate duration --} ($1{-}\bar F_{\mathrm{avg}}{=}4.26{\times}10^{-4}$ vs.\ $4.69{\times}10^{-4}$) and matches its per-input fidelity uniformity (within $1.12{\times}$), at $\sim\!2700{\times}$ lower per-gate cost ($2.1$\,ms vs.\ $5.6$\,s).  \textcolor{black}{The \acs{GRAPE} wall-time here reflects the open-system dynamical map (vectorized $9{\times}9$ propagator on the $3{\times}3$ density matrix), 4-input fidelity averaging, and convergence to the dissipation floor; closed-system unitary-synthesis benchmarks on small Hilbert spaces typically run faster~\citep{dall2026random}.}  First two-qubit results on Haar SU(4) reach $\bar F_{\mathrm{avg}}{=}0.957$ (vs.\ a \acs{GRAPE} floor at $0.998$) at the same $\sim\!4{\times}10^4{\times}$ inference speedup; closing the remaining fidelity gap is open and is discussed in App.~\ref{app:quantum_explored}.  \textcolor{black}{Concurrent work~\citep{lipaei2026fidelity} reports a related \acs{NN} pulse compiler on a closed-system \acs{NMR} platform with an added risk-averse re-optimization layer, but with a restricted axis-angle gate family and without a \acs{GRAPE} accuracy / compute-time comparison.}

% ============================================================================
\section{Discussion, limitations, and outlook}
\label{sec:discussion}
% ============================================================================

The mammalian brain achieves fast, highly effective goal-directed behavior leveraging paired forward/inverse internal models, open-loop multi-step motor commands, and the hierarchical organization of action.  Our findings show that the Inverter framework, built on the same three principles, enables fast and effective planning and control through a feedforward, sequence-level \acs{FoM}-and-\textcolor{black}{\acs{IM}} core that emits entire action sequences in single forward passes. We find that Inverters offer consistently high task performance at a fraction of the inference compute \textcolor{black}{time} used by step-wise \acs{RL} or iterative planners (Suppl.~Fig.~\ref{fig:all_mazes_overlay}).  Across the 9 \acs{D4RL} maze variants, Inverters closely matches or improves the strongest reported baseline on every task, by an average of $+24.2\%$ ($+19.9$ \acs{D4RL} points; range $-1.9\%$ to $+78.2\%$); per-task summary in Suppl.~Tab.~\ref{tab:headline_summary}. First, we discuss the contribution of the three initial, \emph{a priori} brain-inspired principles to these results. 

\textcolor{black}{\textbf{(1)~Paired \acs{FoM}/\textcolor{black}{\acs{IM}} with exact gradients.} Because the Inverter is trained by backpropagating an objective \emph{through} a frozen \acs{FoM} rather than imitating dataset actions, the action-space gradient $\partial \mathcal{J}/\partial a_{t,i}$ is exact in every dimension (Tab.~\ref{tab:paradigms}).  This gradient quality is what allows the Inverter to leave the data's action support and find more optimal solutions (Fig.~\ref{fig:action_scatter}), such as synthesize match-to-dissipation-floor pulses on arbitrary Haar U(2) gates without ever seeing a \acs{GRAPE} solution (Sec.~\ref{sec:quantum}).  The matched failure mode -- \acs{FoM} hacking under narrow expert data (Sec.~\ref{sec:stacked_antman}) -- is the same mechanism turning counter-productive, and motivates the data-coverage strategy (App.~\ref{app:antman_fm_hacking}).}

\textcolor{black}{\textbf{(2)~Open-loop multi-step action sequences ($T{>}1$).} A $T$-step Inverter optimizing a Bolza objective propagates gradients across the whole action sequence and can optimize it holistically, and emits the whole action sequence in a single forward pass. Holistic \emph{"gestalt-level"} optimization is reflected in the lowest curvature variance and lowest peak curvature trajectories (Fig.~\ref{fig:directional_rose}, Panels~G,~H), and the smoothest, most goal-coherent rollouts (Fig.~\ref{fig:trajectories}). Second, ballistic execution of action sequences allows an inference compute time reduction from $\mathcal{O}(\text{horizon})$ to $\mathcal{O}(\text{horizon}/T)$ \acs{NN} forward passes per episode, yielding the observed substantial per-episode speedup.}

\textcolor{black}{\textbf{(3)~Sequenced chunks (hierarchical composition).} Chaining chunks lets a higher-level Inverter target a lower-level one.  Empirically this is what makes Secs.~\ref{sec:inverter_plus_sequencer}--\ref{sec:stacked_antman} work: a simple algorithmic Path Inverter routes chunks through long corridors on \texttt{maze2d-medium}/\texttt{large} once a single chunk no longer spans the path; and the fully learned $n{=}2$ AntMan stack -- Game Inverter atop Locomotion Inverter, communicating via $32$-step waypoint plans on a 58-dimensional symbolic game state -- carries the framework from continuous physics into hybrid symbolic/discrete planning.}

\textcolor{black}{In summary, \textbf{all 3 of our \emph{a-priori} principles proved to be indispensable, each with a distinct failure mode when dropped}: dropping~(1) collapses to behavior cloning; dropping~(2) reduces to Jordan \& Rumelhart's single-step distal teacher, paying the full $T$-fold inference overhead and losing access to sequence-level beyond-data structure; dropping~(3) caps the framework at a single Inverter, losing the hierarchical setups required to solve more complex, long-horizon tasks.}

\subparagraph{Neurosymbolic composition.} A \emph{fourth principle} also emerged \emph{post-hoc} from our implementations: \emph{neurosymbolic composition}. Three of our five Inverters pair a symbolic substrate with a neural amortized inverter: (i) the algorithmic Path Inverter discretizes the data-support occupancy into a cardinal grid and routes through it by \acs{BFS} over a finite cell vocabulary (Sec.~\ref{sec:inverter_plus_sequencer}); (ii) the AntMan Game Inverter emits a sequence over a finite cardinal-direction alphabet $\{$U, D, L, R$\}$ via Gumbel-softmax composed with a precomputed cell$\,\times\,$direction transition table, with wall-validity entering as a hard logical mask (Sec.~\ref{sec:stacked_antman}); (iii) the Pulse Inverter is conditioned on a discrete unitary-gate identity and inverts through a closed-form Lindblad master equation rather than a learned \acs{FoM} (Sec.~\ref{sec:quantum}). Therefore, we add \emph{neurosymbolic composition} as a fourth, candidate organizing principle to the framework (Suppl.~Tab.~\ref{tab:inverters_ns}). Notably, our emergent use of symbolic structure is on the \emph{representation-and-constraint} end of the neurosymbolic spectrum (discrete alphabets, \acs{FSM} transitions, \acs{BFS} search) rather than the \emph{inference-and-synthesis} end (differentiable logical deduction, theorem proving, or program synthesis; e.g.\ DeepProbLog, \acs{NS-CL}).  The Inverter framework is in principle compatible with the latter -- a Level-3 meta-Inverter performing differentiable program synthesis over Level-2 subgoal specifications is a natural extension. Neurosymbolic components emerged for engineering reasons, and algorithmic-level similarity does not by itself imply a deeper neuro-analogy~\citep{hassabis2017neuroscience}; yet it seems interesting to note that the neurosymbolic components which turned out to be useful here all have parallels in symbolic processing the mammalian brain: discretization of continuous space by place and grid cells~\citep{okeefe1971place,hafting2005grid}, sequential motor primitives organized in higher-order motor areas and reflected in muscle synergies and option-level chunking~\citep{ball1999role,tanji1994sma,davella2003synergies,graybiel1998chunking,sutton1999options}, and categorical, invariant single-cell representations in medial temporal cortex~\citep{quianquiroga2005concept}.

\subparagraph{Task-specificity vs. generalization.}
In addition to neurosymbolic composition, a complementary axis along which the implementations of different Inverters in our framework vary is task-specific adaptation vs. architectural generality.  Across the 9 \texttt{maze2d}/\texttt{antmaze-v2} benchmarks alone, the strongest offline-\acs{RL} scores require switching among three structurally distinct algorithms (\acs{AWAC}, \acs{SAC-N}, \acs{ReBRAC}), each selected per benchmark. Our framework uses a single universal architecture (universal causal-transformer \acs{FoM} and Inverter PyTorch classes plus a simple Path Inverter) across all 9 maze variants, but with the auxiliary-loss setup as the main per-task adaptation. The quantum case required an \acs{MLP}-based rather than transformer-based Pulse Inverter.  A full list is given in Suppl.~Tab.~\ref{tab:design_inventory}. Such per-task adaptations may be useful and practical, but moving the framework toward more fully \emph{learnable}, \emph{differentiable}, and \emph{neural-substrate-unified} versions -- replacing remaining algorithmic, analytic, or non-neural components with learned differentiable neural alternatives -- is a natural direction for future work, paralleled in \acs{RL} by meta-\acs{RL}, learned \acs{RL} algorithms, and reformulations as conditional sequence modeling.  \acs{IL} is a well-suited substrate for this aim: its Inverter core is differentiable end-to-end and its slots (\acs{FoM}, \acs{IM}, objective $\mathcal{J}$, hierarchy depth) are modular and well-defined.  Suppl.~Tab.~\ref{tab:strategy_classes} outlines four complementary strategy classes -- Data, Model, Objective, Deployment -- along which this direction can be approached. This way, shared organizational principles may be balanced with specialized instances, as in mammalian motor organization:  The same paired forward/inverse-internal-model brain architecture supports radically different motor niches: primates carry a direct corticomotor neuronal pathway for independent finger control~\citep{lemon2008descending,rathelot2009subdivisions}, echolocating bats overlay a ms-scale audio-motor loop on the same paired forward/inverse architecture~\citep{schnitzler2011auditory}, rodents add a dedicated brainstem \acf{CPG} for $\sim$5--12\,Hz whisking~\citep{kleinfeld2011neuronal}, and elephants control a hydrostatic trunk via a massively expanded facial motor nucleus~\citep{shoshani2006elephant} -- without departing from the unifying planning and control framework. 

\subparagraph{Compute-time.}
\textcolor{black}{The $30$--$100\times$ per-episode compute-time reduction that we observed reflects a reduction in the \emph{number} of \acs{NN} forward passes per episode; whether and how this transfers to practical advantages depends on the deployment regime.  At batch~$1$ on a single device -- our measured setup, and the relevant one for many edge applications or embedded targets -- small models are kernel-launch-limited rather than FLOPs-limited~\citep{williams2009roofline,pope2022efficiently} (App.~\ref{app:timing_methodology}), so the reduction in \acs{NN} invocations transfers directly into wall time.  E.g., in batched cloud serving the regime flips to FLOPs-bound, and our $1.5$\,M-parameter $T{=}128$ transformer performs $\sim\!4$--$10\times$ \emph{more} FLOPs per episode than a small step-wise \acs{MLP}.  The practically interesting scenarios therefore include: (i)~\emph{saving energy} per episode on edge devices, dominated by \acs{NN} invocations rather than raw FLOPs at our model size; (ii)~\emph{resource sharing} on a real robot, where most control ticks become \acs{NN}-free action playback, freeing the GPU/CPU for perception, \acf{SLAM}, or online learning; and (iii)~\emph{iterative-numerics replacement}, where the comparison is to non-amortized solvers and the speedups are largest (e.g.\ $\sim\!4{\times}10^4$ for the Pulse Inverter over \acs{GRAPE}, Sec.~\ref{sec:quantum}).  Across these regimes, the framework's deployment-relevant compute-time advantage is carried by the per-episode reduction in \acs{NN} invocations -- a lever that can be compounded by the engineering directions outlined next.}

\subparagraph{Accelerating real-time applications.}
\textcolor{black}{Two natural directions could push the framework toward sub-millisecond per-invocation inference on embedded targets.  \emph{First, inference-engine compilation}: \texttt{torch.compile}~\citep{ansel2024pytorch2}, CUDA graphs, ONNX/TensorRT export, or porting to JAX with \texttt{jit}~\citep{bradbury2018jax} -- the latter being what makes \acs{ReBRAC} the fastest per-pass entry in Tab.~\ref{tab:maze2d_perf} -- can collapse the kernel-launch floor by another $5$--$10\times$, architecture-agnostically.  \emph{Second, long-horizon-friendly architectures}: structured state-space models~\citep{gu2022s4} such as Mamba~\citep{gu2024mamba} or linear-attention transformers~\citep{katharopoulos2020linear} replace attention's $\mathcal{O}(T^2)$ cost with linear-in-$T$ scaling and admit streaming inference at constant per-step state.  At the chunk lengths explored here ($T{=}16$--$128$) the asymptotic advantage of linear-scan architectures is small, but it may become crucial once \acs{IL} is scaled to $T \gg 10^3$ -- the regime relevant for long-horizon symbolic planning, dexterous manipulation, or long-range locomotion.}

\subparagraph{Ballistic execution.}
\textcolor{black}{A second implication of chunked emission, distinct from the per-call latency budget above, is that the emitted action sequence can be \emph{played out} at the actuator's native rate, independently of \acs{NN}-inference latency \emph{or} the sensory-feedback loop rate.  This open-loop ballistic regime mirrors how mammalian motor control handles actions too fast for closed-loop correction --- saccades, ballistic reaching, drumming, the kHz audio--motor loop of bats, and defensive reflexes.  On the engineering side, the same decoupling becomes crucial whenever the closed loop is the bottleneck rather than the actuator: high-speed manipulation or insect-scale flight, where flight dynamics do not admit a full perception--\acs{NN}--actuation round-trip; surgical robotics combining kHz haptic loops with $\sim\!100$\,Hz vision; and reactive collision avoidance requiring sub-tens-of-ms responses.  Closed-loop replanning with Inverters would enter at the sequence boundary rather than at every tick, leveraging the same coarse-feedback / fine-feedforward division biological motor systems use for fast skilled behavior.}
 
\subparagraph{Extensions.} 
Our experiments in the present work are deterministic, fully-observable, and single-agent; when extending to stochastic, partially-observable, and multi-agent settings, Inverters must accommodate the added uncertainty.  A natural first direction therefore is \textbf{probabilistic Inverters}, \textcolor{black}{in which any of the four Inverter components can independently become stochastic -- the forward model $f_\theta$, the inverse model $g_\phi$, the task context $c$, or the input state $s_0$ (the \acs{POMDP} / belief case) -- with two cross-cutting axes parameterizing the objective: where preferences live (Bolza cost vs.\ prior over outcomes) and how uncertainty enters (pragmatic, risk-sensitive, or epistemic-aware). Stochastic optimal control, control-as-inference, Bayes-adaptive / posterior-sampling, and variational Inverters then appear as cells of that grid; active-inference Inverters would occupy the specific outcome-prior $\times$ expected-free-energy corner~\citep{friston2010free}.}  A second direction is \textbf{latent Inverters}, e.g.,\ with bidirectional latent world models~\citep{sobal2025latent,hansen2024puppeteer} as a more abstract planning substrate.  A third direction is \textbf{deeper hierarchies}: For example, a Level 3 Inverter above Level 2, treating the choice of subgoal specification for Level 2 as itself an inverse problem given a task description.  \textcolor{black}{On the neuro-side, among others, two natural next candidate principles for integration into our framework would be \textbf{predictive coding}~\citep{rao1999predictive,bastos2012canonical} and \textbf{active sensing}~\citep{yang2016theoretical,gottlieb2013information}  -- both particularly relevant to the move from the offline regime studied here to online learning and control: predictive coding as an error-correction signal driven by \acs{FoM} prediction errors, and active sensing as an alternative principled remedy for the narrow-data \acs{FoM}-hacking failure mode (Sec.~\ref{sec:stacked_antman}). A third, complementary candidate is \textbf{dual-process organization} along the lines of the fast/slow distinction~\citep{kahneman2011thinking}: having established \acsp{FoM}/\acsp{IM} for a fast `System~1' that emits a plan in a single feedforward pass, the same trained modules could be repurposed as differentiable substrates for a slower, deliberative `System~2' --- e.g., iterative rollouts, inner-loop search, or test-time refinement through the \acs{FoM} --- when accuracy or safety budgets justify the additional cost. The same neural infrastructure would then span a continuum from amortized reactive control to iterative deliberation.} 

\subparagraph{Inverse World Models.} 
Together these extensions form a trajectory scaling the present \acs{FoM}/\acs{IM} core up to paired \textbf{forward and inverse world models}. Such inverse world models would apply the \acs{IL} paradigm not to a deterministic, fully-observable dynamics model but to a world model in the modern sense -- latent, stochastic, partially observable, possibly multi-agent, language-conditioned where useful -- yielding an Inverter that, in a single feedforward pass, emits an action plan that inverts the forward world model to guide goal-directed behavior. In summary, we propose that the Inverter framework, based on the principles of neuro-inspired Inverse Learning, paths a way to a versatile class of world-interfaces, particularly for latency- and resource-critical embodied AI. 

\newpage

% ============================================================================

% ============================================================================
% Appendix
% ============================================================================
% ============================================================================
% References
% ============================================================================
{\small
\bibliographystyle{unsrtnat}
\bibliography{refs}
}

\clearpage
\appendix
\renewcommand{\tablename}{Supplementary Table}

\section{Technical appendices and supplementary material}

\subsection{Abbreviations}
\label{app:abbreviations}
\begin{acronym}
        \setlength\itemsep{-0.8em} 
        \setlength\parsep{0pt}    % Ensures no extra space between paragraphs
        \acro{AWAC}{Advantage-Weighted Actor-Critic}
        \acro{AWG}{Arbitrary Waveform Generator}
        \acro{BC}{Behavior Cloning}
        \acro{BFGS}{Broyden--Fletcher--Goldfarb--Shanno (quasi-Newton optimizer)}
        \acro{BFS}{Breadth-first search}
        \acro{CBOP}{Conservative Bayesian Model-based Value Expansion for Offline Policy Optimization}
        \acro{CNOT}{Controlled-NOT}
        \acro{COMBO}{Conservative Offline Model-Based Policy Optimization}
        \acro{CORL}{Clean Offline Reinforcement Learning}
        \acro{CPG}{Central Pattern Generator}
        \acro{CQL}{Conservative Q-Learning}
        \acro{CT}{Computed Tomography}
        \acro{D4RL}{Datasets for Deep Data-Driven Reinforcement Learning}
        \acro{DL}{Deep Learning}
        \acro{DP}{Dynamic Programming}
        \acro{DT}{Decision Transformer}
        \acro{EDAC}{Ensemble-Diversified Actor-Critic}
        \acro{EMA}{Exponential Moving Average}
        \acro{FoM}{Forward model}
        \acro{FSM}{Finite State Machine}
        \acro{GRAPE}{Gradient Ascent Pulse Engineering}
        \acro{IL}{Inverse Learning}
        \acro{IM}{Inverse model}
        \acro{IQL}{Implicit Q-Learning}
        \acro{IRL}{Inverse Reinforcement Learning}
        \acro{ISL}{Inverse Sequence Learning}
        \acro{KAK}{Cartan (KAK) decomposition}
        \acro{KL}{Kullback--Leibler (divergence)}
        \acro{LEQ}{Lower Expectile Q-learning}
        \acro{LISTA}{Learned Iterative Shrinkage-Thresholding Algorithm}
        \acro{MAML}{Model-Agnostic Meta-Learning}
        \acro{MAPLE}{Model-based Adaptable Policy LEarning}
        \acro{MBRL}{Model-Based Reinforcement Learning}
        \acro{MILP}{Mixed-Integer Linear Programming}
        \acro{MLP}{Multilayer perceptron}
        \acro{MOBILE}{MOdel-Bellman Inconsistency penalized offLinE Policy Optimization}
        \acro{MOPO}{Model-based Offline Policy Optimization}
        \acro{MOReL}{Model-based Offline Reinforcement Learning}
        \acro{MPC}{Model Predictive Control}
        \acro{MPNet}{Motion Planning Networks}
        \acro{MPPI}{Model Predictive Path Integral}
        \acro{MRI}{Magnetic Resonance Imaging}
        \acro{MuJoCo}{Multi-Joint dynamics with Contact}
        \acro{NeSy}{Neuro-Symbolic AI}
        \acro{NMR}{Nuclear Magnetic Resonance}
        \acro{NN}{Neural network}
        \acro{NS-CL}{Neuro-Symbolic Concept Learner}
        \acro{OC}{Optimal Control}
        \acro{ODE}{Ordinary Differential Equation}
        \acro{OOD}{Out-of-Distribution}
        \acro{PD}{Proportional--Derivative}
        \acro{PG}{Policy Gradient}
        \acro{PILCO}{Probabilistic Inference for Learning Control}
        \acro{POMDP}{Partially Observable Markov Decision Process}
        %\acro{QAOA}{Quantum Approximate Optimization Algorithm}
        \acro{QEC}{Quantum Error Correction}
        \acro{QP}{Quadratic Programming}
        \acro{RAMBO}{Robust Adversarial Model-Based Offline RL}
        \acro{ReBRAC}{Re-evaluated Behavior-Regularized Actor Critic}
        \acro{RL}{Reinforcement Learning}
        \acro{SAC}{Soft Actor-Critic}
        \acro{SAC-N}{Soft Actor-Critic with $N$-critic ensemble}
        \acro{SEM}{Standard Error of the Mean}
        \acro{SLAM}{Simultaneous Localization and Mapping}
        \acro{SNR}{Signal-to-Noise Ratio}
        \acro{SR}{Success Rate}
        \acro{TAP}{Trajectory Autoencoding Planner}
        \acro{TD}{Temporal Difference}
        \acro{TD-MPC2}{Temporal Difference Learning for Model Predictive Control 2}
        \acro{TD3+BC}{Twin Delayed DDPG + Behavior Cloning}
        \acro{TT}{Trajectory Transformer}
    \end{acronym}

\subsection{Experimental details for \texttt{maze2d-umaze-v1}}
\label{app:maze2d_details}

The \texttt{maze2d-umaze-v1} benchmark from \acs{D4RL}~\citep{fu2020d4rl} is a continuous-control navigation task in which a point mass must reach a goal position in a U-shaped corridor.  The offline dataset consists of undirected trajectories generated by a \acf{PD} controller navigating between random waypoints; these follow the angular U-shaped geometry of the maze walls, visible in the heatmap in Figure~\ref{fig:trajectories}.

Our final \acs{FoM} is a chunked causal Transformer with chunk length $L\!=\!16$, $d_\text{model}\!=\!128$, 4 attention heads, 4 layers, and 847k parameters, trained for 300 epochs with Gaussian state noise $\sigma\!=\!0.01$.  We then freeze this model and train the final causal Transformer \textcolor{black}{\acs{IM}} for 2000 epochs through it, using segment length $16$, $d_\text{model}\!=\!192$, 6 heads, 4 layers.  The \textcolor{black}{\acs{IM}} maps $(s_0, s_g) \mapsto a_{1:128}$ in one forward pass.  \textbf{Training goal sampling}: each minibatch sample is a $(s_0, s_g)$ pair with $s_0 \!=\! s_t$ drawn uniformly from the offline buffer and $s_g \!=\! s_{t + H}$ taken at a fixed horizon offset $H \!=\! 128$ ahead in the same trajectory; the sampler is episode-aware (pairs that would cross an episode boundary are rejected). This matches the deployment query distribution: at test time the Path Inverter emits sub-goals at \texttt{wp\_spacing}$\,=\,$$6$\,m on \texttt{maze2d-medium}/\texttt{large} and the goal directly on \texttt{umaze}, both well within one \textcolor{black}{\acs{IM}}-horizon of typical dataset displacement.

\paragraph{Maze2d \textcolor{black}{\acs{IM}} training objective.}  Let $\hat a_{1:H}\!=\!g_\phi(s_0, s_g)$ be the \textcolor{black}{\acs{IM}} output, $\hat s_{1:H}\!=\!f_\theta^{(1:H)}(s_0, \hat a_{1:H})$ the segmented \acs{FoM} rollout (the rollout is chained across $\lceil H/T\rceil$ chunks of length $T\!=\!16$ so the gradient flows end-to-end through the frozen \acs{FoM}), and $\hat p_t\!=\!\hat s_t^{(xy)}$ the predicted positions.  We train $g_\phi$ to minimize
\begin{equation}
\mathcal{L}_\text{\textcolor{black}{\acs{IM}}-2d}
\;=\; \lambda_\text{term}\,\big\|\hat s_H - s_g\big\|^2
\;+\;\lambda_\text{dense}\,\Big[-\tfrac{1}{H}\sum_{t=1}^{H}\!\exp\!\big({-}\|\hat p_t - p_g\|\big)\Big]
\;+\;\lambda_\text{bnd}\,\tfrac{1}{H}\sum_{t=1}^{H}\!\big(1 - O(\hat p_t)\big),
\label{eq:maze2d_iwm_loss}
\end{equation}
with $\lambda_\text{term}\!=\!0$, $\lambda_\text{dense}\!=\!\lambda_\text{bnd}\!=\!5$ in the final runs (the terminal term is absorbed into the dense reward via the exponential profile and was set to $0$ on \texttt{umaze}).  $O\!:\![x_\text{min},x_\text{max}]\!\times\![y_\text{min},y_\text{max}]\!\to\![0,1]$ is a precomputed support map of the offline data, queried at the \acs{FoM}-rolled-out positions by bilinear interpolation (\texttt{F.grid\_sample}, differentiable in $\hat p_t$).  Construction: discretize the data's $(x,y)$ extent into a $64{\times}64$ grid, count visits per cell, apply the chosen \emph{occupancy mode} (\texttt{binary}: $\mathbb 1[\text{count}>0]$ in published runs, or \texttt{frequency}: $\log(1+\text{count})$), Gaussian-smooth with $\sigma_\text{cells}\!=\!1.5$ for sub-cell gradients, and rescale to $[0,1]$.

\paragraph{Loss-balance toggle.}  The Table~\ref{tab:hp_tuned} entry ``boundary mode: \{binary, z-score\}'' is unrelated to the occupancy mode above; ``z-score'' refers to a separate, orthogonal balance toggle in which each raw loss term ($\mathcal{L}_\text{term}, \mathcal{L}_\text{dense}, \mathcal{L}_\text{bnd}, \mathcal{L}_\text{fid}$) is divided by its running \acf{EMA} standard deviation (momentum $0.99$, $100$-batch warm-up) before its $\lambda$ weight is applied -- so $\lambda$ operates on unit-variance signals.  The published \texttt{maze2d} runs leave this toggle off (raw losses) and use \texttt{binary} occupancy.

At evaluation time we consider two execution modes for the same final \textcolor{black}{\acs{IM}} checkpoint.  In \emph{one-shot} mode, the controller executes the generated $128$-step action sequence open-loop and only requests a new plan once that horizon has been consumed.  In \emph{replanning} mode, it executes the first $K\!=\!64$ actions, observes the new state, and replans if the goal has not yet been reached.

For comparison we trained eight offline \acs{RL} baselines with the \acs{CORL} library~\citep{tarasov2024corl}: \acs{BC}, \acs{TD3+BC}~\citep{fujimoto2021minimalist}, \acs{CQL}~\citep{kumar2020cql}, \acs{IQL}~\citep{kostrikov2021iql}, \acs{AWAC}~\citep{nair2020awac}, \acs{ReBRAC}~\citep{tarasov2023rebrac}, \acs{SAC-N}~\citep{an2021sacn}, and Diffuser~\citep{janner2022planning}.  We verified that our reproduced scores are consistent with the reference values reported by \acs{CORL} and evaluated all methods on 100 episodes under the \acs{D4RL}-official protocol (env-default goal, random initial state per \texttt{env.reset()}, score computed via \texttt{env.get\_normalized\_score()}).

% -- Table: umaze maze2d --
\begin{table}[t!]
\centering
\small
\caption{\textbf{Per-episode inference compute on \texttt{maze2d-umaze-v1}} (100 episodes, 300 steps/ep). Each cell reports the number of \acs{NN} forward passes per episode, mean wall time per pass (CUDA-synced, GPU, PyTorch, batch\,1), \emph{other} --- per-episode non-\acs{NN} overhead inside the algorithm (Inverter: per-chunk replan dispatch only (no Path Inverter on this maze); Diffuser: the \acs{PD} tracker running at every env step; step-wise \acs{RL} baselines: 0), the \emph{sum} $=$ \#\acs{NN}-passes $\times$ ms/\acs{NN}-pass $+$ other, and $\times$\,slower relative to our fastest configuration on this maze. $^{\dagger}$ JAX+JIT; all other PyTorch. Inverter rows are mean $\pm$ std over $4$ seed \acp{IM}; the \acs{FoM} is not used at deployment, only the \acs{IM}-related compute time is counted. $^{\ddagger}$ DecisionLLM sum is an order-of-magnitude estimate from the paper.}
\label{tab:maze2d_perf}
% --- auto-generated body (caption/label live in main.tex; generator: maze2d/timing/make_table2.py) ---
\begin{tabular}{@{}l@{\hspace{6pt}}r@{\hspace{6pt}}r@{\hspace{6pt}}r@{\hspace{6pt}}r@{\hspace{5pt}}r@{\hspace{5pt}}r@{\hspace{5pt}}r@{}}
\toprule
Method & score $\uparrow$ & std & \#NN-passes & ms/NN-pass & other & \textbf{sum} & $\times$\,slower \\
\midrule
\acs{BC}~\citep{tarasov2024corl} & $0.36$ & $8.69$ & 300 & $1.70$\,ms & --- & $511.3\,\text{ms}$ & $44.7\times$ \\
\acs{CQL}~\citep{kumar2020cql} & $-8.90$ & $6.11$ & 300 & $1.84$\,ms & --- & $551.2\,\text{ms}$ & $48.2\times$ \\
\acs{BC}-10\%~\citep{tarasov2024corl} & $12.18$ & $4.29$ & 300 & $1.25$\,ms & --- & $374.9\,\text{ms}$ & $32.8\times$ \\
\acs{AWAC}~\citep{nair2020awac} & $82.67$ & $28.30$ & 300 & $1.28$\,ms & --- & $382.5\,\text{ms}$ & $33.4\times$ \\
\acs{IQL}~\citep{kostrikov2021iql} & $42.11$ & $0.58$ & 300 & $1.69$\,ms & --- & $508.4\,\text{ms}$ & $44.4\times$ \\
\acs{TD3+BC}~\citep{fujimoto2021minimalist} & $29.41$ & $12.31$ & 300 & $1.67$\,ms & --- & $502.1\,\text{ms}$ & $43.9\times$ \\
\acs{ReBRAC}~\citep{tarasov2023rebrac} & $106.87$ & $22.16$ & 300 & $0.21$\,ms$^{\dagger}$ & --- & $63.0\,\text{ms}$$^{\dagger}$ & $5.5\times$ \\
\acs{SAC-N}~\citep{an2021sacn} & $130.59$ & $16.52$ & 300 & $1.80$\,ms & --- & $539.3\,\text{ms}$ & $47.1\times$ \\
\acs{EDAC}~\citep{an2021sacn} & $95.26$ & $6.39$ & 300 & $1.74$\,ms & --- & $522.1\,\text{ms}$ & $45.6\times$ \\
\acs{DT}~\citep{chen2021decision} & $18.08$ & $25.42$ & 300 & $3.57$\,ms & --- & $1.07\,\text{s}$ & $93.7\times$ \\
Diffuser~\citep{janner2022planning} & $116.32$ & $34.70$ & 64 & $6.64$\,ms & $3.0$\,ms & $427.8\,\text{ms}$ & $37.4\times$ \\
DecisionLLM~\citep{lv2026decisionllm} & $145.20$ & $35.32$ & 300 & $33.30$\,ms & --- & $10.00\,\text{s}$$^{\ddagger}$ & $873.9\times$ \\
\midrule
\textbf{Inverter $K\!=\!16$} & $\mathbf{164.25}$ & $\mathbf{0.54}$ & 19 & $3.36$\,ms & $9.3$\,ms & $\mathbf{73.2\,\text{ms}}$ & $\mathbf{6.4\times}$ \\
\textbf{Inverter $K\!=\!32$} & $\mathbf{164.77}$ & $\mathbf{0.34}$ & 10 & $2.67$\,ms & $4.3$\,ms & $\mathbf{31.0\,\text{ms}}$ & $\mathbf{2.7\times}$ \\
\textbf{Inverter $K\!=\!64$} & $\mathbf{165.19}$ & $\mathbf{0.80}$ & 5 & $3.14$\,ms & $2.9$\,ms & $\mathbf{18.7\,\text{ms}}$ & $\mathbf{1.6\times}$ \\
\textbf{Inverter $K\!=\!128$} & $\mathbf{161.64}$ & $\mathbf{2.17}$ & 3 & $3.19$\,ms & $1.9$\,ms & $\mathbf{11.4\,\text{ms}}$ & $\mathbf{1.0\times}$ \\
\bottomrule
\end{tabular}

\end{table}

\subsection{Timing measurement protocol}
\label{app:timing_methodology}

This appendix gives the exact protocol behind Table~\ref{tab:maze2d_perf} and explains why the Inverter transformer is not meaningfully slower than the baselines' \acp{MLP} on a single forward pass.

\textcolor{black}{\paragraph{Terminology.} Throughout this paper, \emph{``inference compute time''} refers to per-episode wall time on a single device at batch\,1; in a launch-overhead-limited small-model regime this is the deployment-relevant metric, distinct from FLOPs.  Per-pass parity between the Inverter transformer and the baseline \acp{MLP} (see ``Why isn't a transformer slower than an \acs{MLP}?'' below) confirms we are in this regime on the architectures evaluated here.}

\paragraph{Hardware and software.} All timings are measured on a single GPU (cuda:0) with PyTorch in \texttt{eval} mode under \texttt{torch.no\_grad}.  \acs{ReBRAC} is the sole exception: its inference is measured from JAX+JIT code and is therefore marked with $^{\dagger}$ in Table~\ref{tab:maze2d_perf}.  \acs{MuJoCo} stepping uses the default single-thread CPU implementation.

\paragraph{Per-pass wall time.} Each neural-network forward pass is timed using \texttt{time.perf\_counter()} wrapped by \texttt{torch.cuda.synchronize()} on both sides, so the reported wall time reflects the actual completion of the GPU work, not just the kernel-launch queue.  Before the $100$-episode evaluation starts we run $10$ warm-up forward passes on dummy inputs (to flush kernel autotuning and CUDA graph compilation) and then clear the timing buffers.  The same CUDA-sync wrapper is used around \texttt{env.step} (to time the \acs{MuJoCo} physics), around each denoising step for Diffuser, and around each \textcolor{black}{\acs{IM}} and \acs{FoM}  forward for the Inverter.

\paragraph{Per-episode accounting.} We decompose an episode's wall time into four additive components: (i) \acs{NN} forward passes, (ii) \emph{other} algorithmic overhead -- Python control loop, tensor prep, \texttt{.cpu().numpy()} transfers, replan dispatch (the Inverter) or the \acs{PD} tracker (Diffuser), (iii) \texttt{env.step}, and (iv) a residual of a few milliseconds for episode-level bookkeeping.  Table~\ref{tab:maze2d_perf}'s \emph{sum} column reports $(i)+(ii)$, i.e.\ the algorithm's inference compute \textcolor{black}{time} budget.  Environment stepping and framework glue are excluded on purpose: we want a quantity that is invariant to the simulator and portable to real-robot deployment, where \texttt{env.step} is replaced by physical dynamics.

\paragraph{Evaluation protocol.}  All methods are evaluated on exactly $100$ episodes of \texttt{maze2d-umaze-v1} under the \acs{D4RL}-official protocol -- env-default fixed target, random initial states from \texttt{env.reset()}, fixed evaluation seed so the 100 starts are reproduced across runs -- with a hard cap of $300$ environment steps per episode.  Inverter numbers average over $4$ independently-trained seed \textcolor{black}{\acp{IM}} (and their matching \acp{FoM}).

\paragraph{Transformer vs. \acs{MLP} compute time.}
A natural question is how a $1.5$\,M-parameter, $4$-layer / $192$-d-model transformer (the Inverter) ends up at essentially the same $\sim\!1.5$\,ms-per-forward floor as a $\sim\!200$\,k-parameter $3$-layer / $256$-hidden \acs{MLP} (\acs{BC}, \acs{IQL}, \acs{SAC-N}, \acs{TD3+BC}, \ldots).  The answer is that, at batch\,$1$ GPU inference, neither network is compute-bound: both run a small number of dense kernels whose wall time is dominated by CUDA kernel-launch overhead ($\sim\!10$\,$\mu$s per launch) rather than by floating-point work.  A $4$-layer transformer incurs roughly $4$--$8\times$ more launches than a $3$-layer \acs{MLP}, but on a modern GPU those extra launches still fit inside the same $\sim\!1$--$2$\,ms fixed floor.  Concretely, Table~\ref{tab:maze2d_perf} shows the \acs{MLP} baselines clustered between $0.21$ and $3.57$\,ms/pass (\acs{BC} at $1.70$, \acs{IQL} at $1.69$, \acs{SAC-N} at $1.80$, \acs{TD3+BC} at $1.67$), and the Inverter transformer at $1.51$--$1.68$\,ms/pass -- inside that same band.  The per-pass parity is also what makes the Inverter's sum-per-episode advantage so large: because the transformer is \emph{not} paying an extra order of magnitude per forward, the benefit of emitting a full $128$-step action chunk per forward (instead of one action per env step) translates directly into a $30$--$100\times$ reduction in total forwards per episode, and a matching reduction in sum wall time.

\subsection{Detailed performance tables (\texttt{maze2d-medium}/\texttt{large} and \texttt{antmaze})}
\label{app:perf_tables}

This appendix collects the per-method performance tables for \texttt{maze2d-medium-v1} (Table~\ref{tab:maze2d_perf_medium}) and \texttt{maze2d-large-v1} (Table~\ref{tab:maze2d_perf_large}) referenced in Sec.~\ref{sec:inverter_plus_sequencer}, and for the six \texttt{antmaze-v2} variants (Table~\ref{tab:antmaze_perf}) referenced in Sec.~\ref{sec:antmaze_eval}.  Table~\ref{tab:headline_summary} below summarizes the per-task improvement of the Inverter over the strongest reported baseline on each variant, both in absolute \acs{D4RL}-score points and as a percentage of the baseline, with the row-wise mean across all 9 tasks.

\begin{table}[h]
\centering
\small
\caption{\textbf{Per-task \acs{D4RL} improvement of the Inverter over the strongest reported baseline, summarized across all 9 \texttt{maze2d}/\texttt{antmaze-v2} variants.}  Baselines per variant: \acs{SAC-N} / Diffuser / \acs{AWAC} are the per-maze winners on \texttt{maze2d} (Table~\ref{tab:maze2d_headline}); \acs{ReBRAC} is used as the comparator on every \texttt{antmaze-v2} variant (Table~\ref{tab:antmaze_headline}).  $\Delta$ pts = Inverter $-$ baseline; $\Delta$ \% = $\Delta$ / baseline.  $^{\ast}$On \texttt{antmaze-medium-play} the Inverter is nominally $1.7$ points behind \acs{ReBRAC} ($87.8\!\pm\!8.0$ vs.\ $89.5\!\pm\!3.4$), well within the joint errorbar.  Final row: mean across all $9$ tasks (\acs{D4RL} points and \%, equally weighted).}
\label{tab:headline_summary}
\begin{tabular}{@{}lllrrrr@{}}
\toprule
Task & Baseline & & Baseline & Inverter & $\Delta$ pts & $\Delta$ \% \\
\midrule
\texttt{maze2d-umaze-v1}        & \acs{SAC-N}    & & $130.6$ & $\mathbf{161.6}$ & $+31.0$ & $+23.7$ \\
\texttt{maze2d-medium-v1}       & Diffuser & & $130.1$ & $\mathbf{166.8}$ & $+36.7$ & $+28.2$ \\
\texttt{maze2d-large-v1}        & \acs{AWAC}     & & $209.1$ & $\mathbf{220.7}$ & $+11.6$ & $+5.5$ \\
\midrule
\texttt{antmaze-umaze-v2}       & \acs{ReBRAC}   & & $97.8$  & $\mathbf{99.5}$  & $+1.7$  & $+1.7$ \\
\texttt{antmaze-umaze-diverse-v2}& \acs{ReBRAC}  & & $83.5$  & $\mathbf{99.8}$  & $+16.3$ & $+19.5$ \\
\texttt{antmaze-medium-play-v2}$^{\ast}$ & \acs{ReBRAC} & & $\mathbf{89.5}$ & $87.8$ & $-1.7$ & $-1.9$ \\
\texttt{antmaze-medium-diverse-v2}& \acs{ReBRAC} & & $83.5$ & $\mathbf{96.5}$ & $+13.0$ & $+15.6$ \\
\texttt{antmaze-large-play-v2}  & \acs{ReBRAC}   & & $52.2$ & $\mathbf{93.0}$ & $+40.8$ & $+78.2$ \\
\texttt{antmaze-large-diverse-v2}& \acs{ReBRAC}  & & $64.0$ & $\mathbf{94.0}$ & $+30.0$ & $+46.9$ \\
\midrule
\textbf{Mean (n=9)}             &          & &        &                  & $\mathbf{+19.9}$ & $\mathbf{+24.2}$ \\
\bottomrule
\end{tabular}
\end{table}

\paragraph{Protocol for \texttt{maze2d-medium}/\texttt{large}.}  Same \acs{D4RL}-official protocol as Table~\ref{tab:maze2d_perf}: $100$ episodes, env-default fixed target, random initial states from \texttt{env.reset()} with a fixed evaluation seed, $600$-step cap on medium, $800$-step cap on large.  \acs{CORL} numbers are taken from the \citet{tarasov2024corl} ``Last Scores'' benchmark (we did not train our own medium/large \acs{CORL} checkpoints); per-pass times come from our local umaze benchmark (same actor architectures) and are scaled by the target maze's step count for the \emph{sum} column.  Diffuser timings on medium/large are estimated from our measured umaze per-denoise-step timing scaled by horizon, since we do not yet have a pretrained Diffuser checkpoint for those mazes locally (flagged $^{\ast}$); the Diffuser \acs{D4RL} numbers are from \citet{janner2022planning} Table~2.  The \emph{other} column (per-episode non-\acs{NN} overhead) is $\sim\!27$\,ms on medium and $\sim\!36$\,ms on large because the algorithmic Path Inverter runs a 4-connected \acs{BFS} on a $\sim\!100$-cell grid once per replan event and occasionally performs a full re-plan when the stuck-check triggers (App.~\ref{app:waypoint_planner}); even so, this is an order of magnitude below the \acs{MLP} cost of every step-wise baseline.

% -- Table: medium maze2d --------------------------------
\begin{table}[t!]
\centering
\small
\caption{\textbf{Per-episode inference compute on \texttt{maze2d-medium-v1}} (100 episodes, 600 steps/ep). Each cell reports the number of \acs{NN} forward passes per episode, mean wall time per pass (CUDA-synced, GPU, PyTorch, batch\,1), \emph{other} --- per-episode non-\acs{NN} overhead inside the algorithm (Inverter: replan dispatch + the data-derived \acs{BFS} Path Inverter; Diffuser: the \acs{PD} tracker running at every env step; step-wise \acs{RL} baselines: 0), the \emph{sum} $=$ \#\acs{NN}-passes $\times$ ms/\acs{NN}-pass $+$ other, and $\times$\,slower relative to our fastest configuration on this maze. $^{\dagger}$ JAX+JIT; all other PyTorch. \acs{CORL} baselines have no trained \texttt{medium} checkpoint in our local benchmark; per-pass time is measured on \texttt{umaze} (same actor architectures) and the \emph{sum} column scales it by $600$ environment steps. The Diffuser row is measured locally on a checkpoint we trained on \texttt{maze2d-medium-v1} for 2M steps matching the configuration of \citet{janner2022planning}; the reported \acs{D4RL} score is the locally measured one. Inverter row is mean $\pm$ std over $4$ seed \acp{IM}; \emph{other} here is the per-episode cost of the data-derived cardinal-\acs{BFS} Path Inverter (Appendix~\ref{app:waypoint_planner}).}
\label{tab:maze2d_perf_medium}
% --- auto-generated body (caption/label live in main.tex; generator: maze2d/timing/make_table2.py) ---
\begin{tabular}{@{}l@{\hspace{6pt}}r@{\hspace{6pt}}r@{\hspace{6pt}}r@{\hspace{6pt}}r@{\hspace{5pt}}r@{\hspace{5pt}}r@{\hspace{5pt}}r@{}}
\toprule
Method & score $\uparrow$ & std & \#NN-passes & ms/NN-pass & other & \textbf{sum} & $\times$\,slower \\
\midrule
\acs{BC} & $0.79$ & $3.25$ & 600 & $1.70$\,ms & --- & $1.02\,\text{s}$ & $14.0\times$ \\
\acs{BC}-10\% & $14.25$ & $2.33$ & 600 & $1.25$\,ms & --- & $749.8\,\text{ms}$ & $10.3\times$ \\
\acs{DT} & $31.71$ & $26.33$ & 600 & $3.57$\,ms & --- & $2.14\,\text{s}$ & $29.4\times$ \\
\acs{IQL} & $34.85$ & $2.72$ & 600 & $1.69$\,ms & --- & $1.02\,\text{s}$ & $14.0\times$ \\
\acs{AWAC} & $52.88$ & $55.12$ & 600 & $1.28$\,ms & --- & $765.0\,\text{ms}$ & $10.5\times$ \\
\acs{EDAC} & $57.04$ & $3.45$ & 600 & $1.74$\,ms & --- & $1.04\,\text{s}$ & $14.3\times$ \\
\acs{TD3+BC} & $59.45$ & $36.25$ & 600 & $1.67$\,ms & --- & $1.00\,\text{s}$ & $13.8\times$ \\
\acs{CQL} & $86.11$ & $9.68$ & 600 & $1.84$\,ms & --- & $1.10\,\text{s}$ & $15.1\times$ \\
\acs{SAC-N} & $88.61$ & $18.72$ & 600 & $1.80$\,ms & --- & $1.08\,\text{s}$ & $14.8\times$ \\
\acs{ReBRAC} & $105.11$ & $31.67$ & 600 & $0.21$\,ms$^{\dagger}$ & --- & $126.0\,\text{ms}$$^{\dagger}$ & $1.7\times$ \\
Diffuser & $130.07$ & $22.70$ & 256 & $14.54$\,ms & $14.5$\,ms & $3.74\,\text{s}$ & $51.3\times$ \\
\midrule
\textbf{Inverter $K\!=\!16$} & $\mathbf{166.82}$ & $\mathbf{1.20}$ & 38 & $1.62$\,ms & $11.2$\,ms & $\mathbf{72.9\,\text{ms}}$ & $\mathbf{1.0\times}$ \\
\bottomrule
\end{tabular}

\end{table}

% -- Table: large maze2d ---------------------------------
\begin{table}[t!]
\centering
\small
\caption{\textbf{Per-episode inference compute on \texttt{maze2d-large-v1}} (100 episodes, 800 steps/ep). Each cell reports the number of \acs{NN} forward passes per episode, mean wall time per pass (CUDA-synced, GPU, PyTorch, batch\,1), \emph{other} --- per-episode non-\acs{NN} overhead inside the algorithm (Inverter: replan dispatch + the data-derived \acs{BFS} Path Inverter; Diffuser: the \acs{PD} tracker running at every env step; step-wise \acs{RL} baselines: 0), the \emph{sum} $=$ \#\acs{NN}-passes $\times$ ms/\acs{NN}-pass $+$ other, and $\times$\,slower relative to our fastest configuration on this maze. $^{\dagger}$ JAX+JIT; all other PyTorch. \acs{CORL} baselines have no trained \texttt{large} checkpoint in our local benchmark; per-pass time is measured on \texttt{umaze} (same actor architectures) and the \emph{sum} column scales it by $800$ environment steps. The Diffuser row is measured locally on a checkpoint we trained on \texttt{maze2d-large-v1} for 2M steps matching the configuration of \citet{janner2022planning}; the reported \acs{D4RL} score is the locally measured one. Inverter row is mean $\pm$ std over $4$ seed \acp{IM}; \emph{other} here is the per-episode cost of the data-derived cardinal-\acs{BFS} Path Inverter (Appendix~\ref{app:waypoint_planner}).}
\label{tab:maze2d_perf_large}
% --- auto-generated body (caption/label live in main.tex; generator: maze2d/timing/make_table2.py) ---
\begin{tabular}{@{}l@{\hspace{6pt}}r@{\hspace{6pt}}r@{\hspace{6pt}}r@{\hspace{6pt}}r@{\hspace{5pt}}r@{\hspace{5pt}}r@{\hspace{5pt}}r@{}}
\toprule
Method & score $\uparrow$ & std & \#NN-passes & ms/NN-pass & other & \textbf{sum} & $\times$\,slower \\
\midrule
\acs{BC} & $2.26$ & $4.39$ & 800 & $1.70$\,ms & --- & $1.36\,\text{s}$ & $14.6\times$ \\
\acs{BC}-10\% & $11.32$ & $5.10$ & 800 & $1.25$\,ms & --- & $999.7\,\text{ms}$ & $10.7\times$ \\
\acs{CQL} & $23.75$ & $36.70$ & 800 & $1.84$\,ms & --- & $1.47\,\text{s}$ & $15.7\times$ \\
\acs{DT} & $35.66$ & $28.20$ & 800 & $3.57$\,ms & --- & $2.86\,\text{s}$ & $30.5\times$ \\
\acs{IQL} & $61.72$ & $3.50$ & 800 & $1.69$\,ms & --- & $1.36\,\text{s}$ & $14.5\times$ \\
\acs{ReBRAC} & $78.33$ & $61.77$ & 800 & $0.21$\,ms$^{\dagger}$ & --- & $168.0\,\text{ms}$$^{\dagger}$ & $1.8\times$ \\
\acs{EDAC} & $95.60$ & $22.92$ & 800 & $1.74$\,ms & --- & $1.39\,\text{s}$ & $14.9\times$ \\
\acs{TD3+BC} & $97.10$ & $25.41$ & 800 & $1.67$\,ms & --- & $1.34\,\text{s}$ & $14.3\times$ \\
Diffuser & $123.07$ & $62.22$ & 256 & $14.32$\,ms & $17.0$\,ms & $3.68\,\text{s}$ & $39.3\times$ \\
\acs{SAC-N} & $204.76$ & $1.19$ & 800 & $1.80$\,ms & --- & $1.44\,\text{s}$ & $15.4\times$ \\
\acs{AWAC} & $209.13$ & $8.19$ & 800 & $1.28$\,ms & --- & $1.02\,\text{s}$ & $10.9\times$ \\
\midrule
\textbf{Inverter $K\!=\!16$} & $\mathbf{220.66}$ & $\mathbf{0.20}$ & 50 & $1.59$\,ms & $14.3$\,ms & $\mathbf{93.7\,\text{ms}}$ & $\mathbf{1.0\times}$ \\
\bottomrule
\end{tabular}

\end{table}

% -- Table: antmaze D4RL + per-step inference compute ---------------
\begin{table}[t!]
\centering
\small
\caption{\textbf{Antmaze: \acs{D4RL} score and per-step inference compute.} \textbf{Top}: \acs{D4RL} success-rate score (\%) over 100 episodes on each \texttt{antmaze-v2} variant (\acs{CORL} baselines~\citep{tarasov2024corl}: mean$\,\pm\,$std over $4$ training seeds; Inverter: mean$\,\pm\,$std over $4$ \acs{IM} seeds; best-per-column in bold). \textbf{Bottom}: $\times$\,slower per env step against the fastest method on each maze, measured on a single A40 GPU (PyTorch, batch\,1, CUDA-synced). For step-wise actors ms/step $=$ ms/pass (the \acs{NN} runs once per env step) and is env-independent; ms/pass values are reported in Table~\ref{tab:maze2d_perf}. For the Inverter, ms/step $=$ (\#\acs{NN}-pass $\times$ ms/pass + \acs{BFS}-Path Inverter overhead) / mean steps-to-goal, averaged across the $4$ \acs{IM} seeds. $^{\dagger}$ \acs{ReBRAC}: JAX+JIT timing. Best-per-column ($\mathbf{1.0\times}$) bolded.}
\label{tab:antmaze_perf}
% --- auto-generated body (caption/label live in main.tex; generator: antmaze/timing/make_antmaze_table.py) ---
\begin{tabular}{@{}l@{\hspace{6pt}}rrrrrr@{}}
\toprule
& \multicolumn{6}{c}{\textbf{D4RL score (\%) on antmaze-v2}} \\
\cmidrule(lr){2-7}
Method & u-umaze & u-divrs & m-play & m-divrs & l-play & l-divrs \\
\midrule
\acs{BC} & $55.2\,{\scriptstyle\pm\,4.2}$ & $47.2\,{\scriptstyle\pm\,4.1}$ & $0.0$ & $0.8\,{\scriptstyle\pm\,0.8}$ & $0.0$ & $0.0$ \\
\acs{BC}-10\% & $65.8\,{\scriptstyle\pm\,5.3}$ & $44.0\,{\scriptstyle\pm\,1.0}$ & $2.0\,{\scriptstyle\pm\,0.7}$ & $5.8\,{\scriptstyle\pm\,9.4}$ & $0.0$ & $0.8\,{\scriptstyle\pm\,0.8}$ \\
\acs{TD3+BC} & $70.8\,{\scriptstyle\pm\,39.2}$ & $44.8\,{\scriptstyle\pm\,11.6}$ & $0.2\,{\scriptstyle\pm\,0.4}$ & $0.2\,{\scriptstyle\pm\,0.4}$ & $0.0$ & $0.0$ \\
\acs{AWAC} & $57.8\,{\scriptstyle\pm\,10.3}$ & $58.0\,{\scriptstyle\pm\,7.7}$ & $0.0$ & $0.0$ & $0.0$ & $0.0$ \\
\acs{CQL} & $92.8\,{\scriptstyle\pm\,1.9}$ & $37.2\,{\scriptstyle\pm\,3.7}$ & $65.8\,{\scriptstyle\pm\,11.6}$ & $67.2\,{\scriptstyle\pm\,3.6}$ & $20.8\,{\scriptstyle\pm\,7.3}$ & $20.5\,{\scriptstyle\pm\,13.2}$ \\
\acs{IQL} & $77.0\,{\scriptstyle\pm\,5.5}$ & $54.2\,{\scriptstyle\pm\,5.5}$ & $65.8\,{\scriptstyle\pm\,11.7}$ & $73.8\,{\scriptstyle\pm\,5.5}$ & $42.0\,{\scriptstyle\pm\,4.5}$ & $30.2\,{\scriptstyle\pm\,3.6}$ \\
\acs{ReBRAC} & $\mathbf{97.8}\,{\scriptstyle\pm\,1.5}$ & $\mathbf{83.5}\,{\scriptstyle\pm\,7.0}$ & $\mathbf{89.5}\,{\scriptstyle\pm\,3.4}$ & $\mathbf{83.5}\,{\scriptstyle\pm\,8.2}$ & $\mathbf{52.2}\,{\scriptstyle\pm\,29.0}$ & $\mathbf{64.0}\,{\scriptstyle\pm\,5.4}$ \\
\acs{SAC-N} & $0.0$ & $0.0$ & $0.0$ & $0.0$ & $0.0$ & $0.0$ \\
\acs{EDAC} & $0.0$ & $0.0$ & $0.0$ & $0.0$ & $0.0$ & $0.0$ \\
\acs{DT} & $57.0\,{\scriptstyle\pm\,9.8}$ & $51.8\,{\scriptstyle\pm\,0.4}$ & $0.0$ & $0.0$ & $0.0$ & $0.0$ \\
\midrule
\textbf{Inverter (ours)} & $\mathbf{99.5}\,{\scriptstyle\pm\,0.9}$ & $\mathbf{99.8}\,{\scriptstyle\pm\,0.4}$ & $87.8\,{\scriptstyle\pm\,8.0}$ & $\mathbf{96.5}\,{\scriptstyle\pm\,5.0}$ & $\mathbf{93.0}\,{\scriptstyle\pm\,2.6}$ & $\mathbf{94.0}\,{\scriptstyle\pm\,5.0}$ \\
\bottomrule
\end{tabular}%
\\[0.6em]
\begin{tabular}{@{}l@{\hspace{6pt}}rrrrrr@{}}
\toprule
& \multicolumn{6}{c}{\textbf{$\times$\,slower per env step on antmaze-v2}} \\
\cmidrule(lr){2-7}
Method & u-umaze & u-divrs & m-play & m-divrs & l-play & l-divrs \\
\midrule
\acs{BC} & $12.0\times$ & $12.1\times$ & $12.8\times$ & $13.0\times$ & $12.7\times$ & $12.7\times$ \\
\acs{BC}-10\% & $8.8\times$ & $8.9\times$ & $9.4\times$ & $9.5\times$ & $9.3\times$ & $9.3\times$ \\
\acs{TD3+BC} & $11.8\times$ & $11.8\times$ & $12.6\times$ & $12.7\times$ & $12.5\times$ & $12.5\times$ \\
\acs{AWAC} & $9.0\times$ & $9.1\times$ & $9.6\times$ & $9.8\times$ & $9.6\times$ & $9.6\times$ \\
\acs{CQL} & $13.0\times$ & $13.0\times$ & $13.8\times$ & $14.0\times$ & $13.7\times$ & $13.7\times$ \\
\acs{IQL} & $11.9\times$ & $12.0\times$ & $12.7\times$ & $12.9\times$ & $12.6\times$ & $12.6\times$ \\
\acs{ReBRAC} & $1.5\times$$^{\dagger}$ & $1.5\times$$^{\dagger}$ & $1.6\times$$^{\dagger}$ & $1.6\times$$^{\dagger}$ & $1.6\times$$^{\dagger}$ & $1.6\times$$^{\dagger}$ \\
\acs{SAC-N} & $12.7\times$ & $12.8\times$ & $13.5\times$ & $13.7\times$ & $13.4\times$ & $13.4\times$ \\
\acs{EDAC} & $12.3\times$ & $12.3\times$ & $13.1\times$ & $13.3\times$ & $13.0\times$ & $13.0\times$ \\
\acs{DT} & $25.1\times$ & $25.3\times$ & $26.8\times$ & $27.3\times$ & $26.6\times$ & $26.6\times$ \\
\midrule
\textbf{Inverter (ours)} & $\mathbf{1.0\times}$ & $\mathbf{1.0\times}$ & $\mathbf{1.0\times}$ & $\mathbf{1.0\times}$ & $\mathbf{1.0\times}$ & $\mathbf{1.0\times}$ \\
\bottomrule
\end{tabular}%

\end{table}

\begin{figure}[!htbp]
\centering
\includegraphics[width=0.86\linewidth]{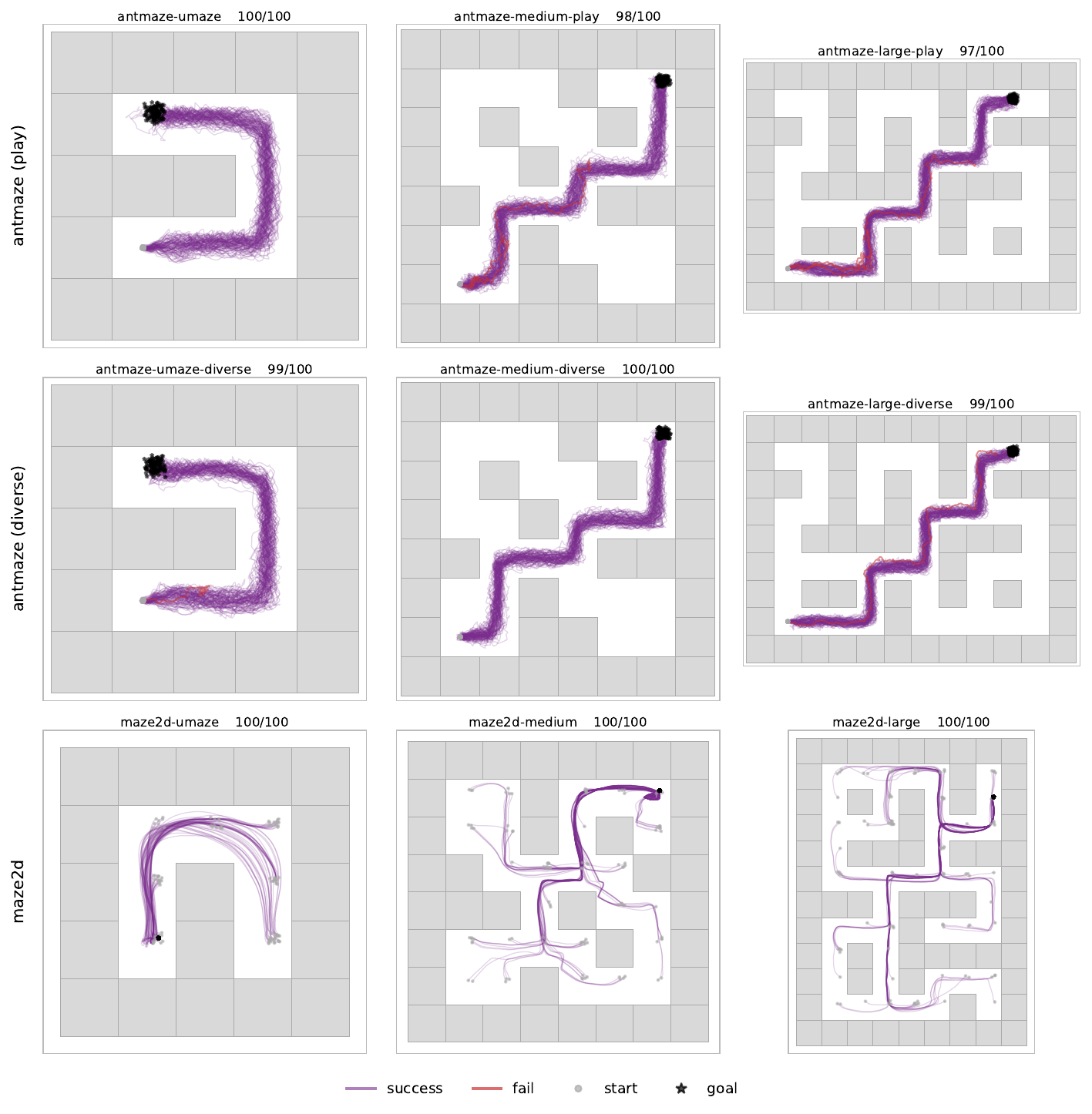}
\caption{\textbf{Per-episode trajectory overlays for the Inverter on every maze variant we evaluate.}  $3\!\times\!3$ grid: rows $=$ (\texttt{antmaze} play / \texttt{antmaze} diverse / \texttt{maze2d}); columns $=$ small / medium / large maze.  Each panel overlays $100$ evaluation trajectories from a single representative \textcolor{black}{\acs{IM}} seed (number-of-successes printed in the panel title).  Antmaze evaluations (top two rows) use a fixed corner goal, giving tight beam-like overlays; \texttt{maze2d} evaluations (bottom row) randomize start/goal pairs per episode, giving fan-of-paths through the corridor graph.}
\label{fig:all_mazes_overlay}
\end{figure}

\subsection{Minimum-time control under viscous damping}
\label{app:damped_bangbang}

To interpret the smoother \textcolor{black}{\acs{IM}} trajectories in Figure~\ref{fig:trajectories}, it is useful to separate the geometry of the \emph{state trajectory} from the structure of the \emph{control signal}.  Consider the idealized one-dimensional point-mass model
\begin{equation}
\dot x = v, \qquad \dot v = -\beta v + u, \qquad |u| \le u_{\max}, \qquad \beta > 0,
\label{eq:damped_pointmass}
\end{equation}
where $\beta$ is a viscous damping coefficient.  This captures bounded actuation and linear damping while abstracting away maze walls, goal radii, and replanning.  We consider the minimum-time transfer from an initial state $(x_0, v_0)$ to the terminal state $(0,0)$.

The minimum-time transfer to the origin for this damped double integrator is a standard textbook result in optimal control~\textcolor{black}{\citep{athans1966optimal,kirk1970optimal}}. Pontryagin's maximum principle dictates that the time-optimal input is strictly bang-bang ($u \in \{+u_{\max}, -u_{\max}\}$) with at most one switch.

Furthermore, the exact switching curve is given in closed form by:
\begin{equation}
x = -\operatorname{sgn}(v)\left[\frac{|v|}{\beta} - \frac{u_{\max}}{\beta^2}\log\!\left(1 + \frac{\beta |v|}{u_{\max}}\right)\right].
\label{eq:damped_switching_curve}
\end{equation}
The optimal policy accelerates maximally toward the goal until it hits this curve, then brakes maximally to arrive at the target with zero velocity. In the limit $\beta \to 0$, Eq.~\eqref{eq:damped_switching_curve} reduces to the familiar undamped switching parabola $x = -\frac{1}{2u_{\max}}|v|v$. Thus, viscous damping changes the shape of the switching condition from a parabola to a logarithmic curve, but it does not alter the fundamental bang-bang nature of the minimum-time input.

This result should be interpreted as a \emph{local straight-segment model}, not as an exact theorem for the full \texttt{maze2d} benchmark.  The benchmark only requires entry into a finite goal region, not arrival with zero terminal velocity, and the full maze introduces walls and path constraints.  In that setting the exact optimal law can differ, and the braking phase may disappear if first arrival to the goal set is all that matters.  

\subsection{Simple algorithmic Path Inverter (Level 2): data-driven \acs{BFS} over offline-data density}
\label{app:waypoint_planner}

\begin{figure}[t]
\centering
\includegraphics[width=0.70\linewidth]{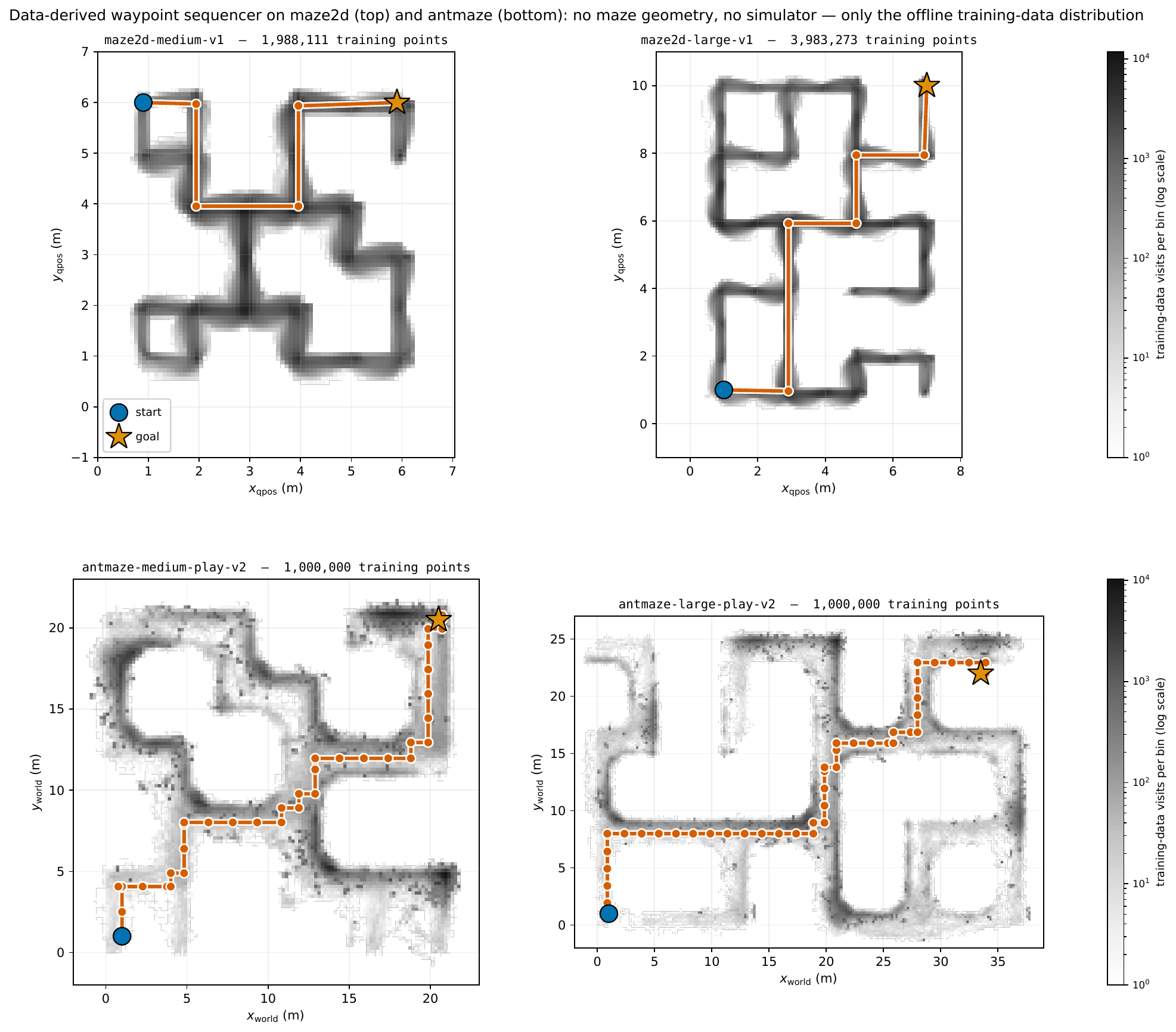}
\caption{\textbf{The simple algorithmic Path Inverter uses only the offline training-data distribution, no maze geometry.}  Background: log-density of the training-data states (shared colorbars on the right; top row in qpos coordinates, bottom row in world coordinates).  No maze walls or simulator information is drawn or provided to the Path Inverter -- the corridors are visible purely because the offline data concentrates there.  A cardinal-\acs{BFS} waypoint chain produced by our data-derived Path Inverter is overlaid in orange (circle markers = intermediate sub-goals); start and goal are shown in blue and orange.  \textbf{Top row}: \texttt{maze2d-medium-v1} (1{,}988{,}111 training points) and \texttt{maze2d-large-v1} (3{,}983{,}273 points), with the maze2d planner settings (\texttt{resolution}\,$=$\,$0.5$\,m, $\tau\!=\!2000$ visits, \texttt{min\_leg}\,$=$\,$1$\,m).  \textbf{Bottom row}: \texttt{antmaze-medium-play-v2} and \texttt{antmaze-large-play-v2} (1{,}000{,}000 points each), with the antmaze planner settings (\texttt{resolution}\,$=$\,$1.0$\,m, $\tau\!=\!100$, \texttt{min\_leg}\,$=$\,$0$); same code path otherwise.  In all four panels the chain turns through exactly those corners where the data concentration itself turns, demonstrating that deployment-time routing can be driven by training-data support alone.}
\label{fig:waypoint_sequencer}
\end{figure}

For the larger \texttt{maze2d} and \texttt{antmaze} layouts, a single Inverter chunk no longer reaches the goal: corridors contain multiple $90^{\circ}$ turns and the longest shortest path through free space exceeds the \textcolor{black}{\acs{IM}'s} training horizon.  We therefore couple the (unchanged) Inverter with a trivial algorithmic Path Inverter at Level~2 that emits intermediate sub-goals along a feasible corridor.  The key constraint: the planner must be buildable from the \emph{offline training data alone} -- no access to maze geometry, simulator, or the underlying occupancy grid.  A single class (\texttt{common.cardinal\_bfs\_planner.CardinalBFSPlanner}) drives both task families through different hyperparameters (Fig.~\ref{fig:waypoint_sequencer}).

\paragraph{Construction.}  Five stages:
(i)~\emph{Data-derived occupancy grid}: discretize qpos into a 2-D grid with \emph{density-aligned origin} (sub-cell phase set to the 1-D marginals' dominant peak, aligning corridor centerlines with cell centers; \acs{SNR} jumps ${\sim}10\times$); call a cell free if it received $\geq \tau$ training points (true corridors get $10^4$--$10^5$ visits, wall-adjacent strays $10^1$--$10^3$, so $\tau\!=\!2000$ cleanly separates them); cell ``centers'' are the density-weighted qpos mean.
(ii)~\emph{4-connected \acs{BFS}} from start to goal cell, tie-breaking equally-short paths by maximum cumulative distance-to-wall via a single backward \acf{DP} pass over the \acs{BFS} layers (tie-break weight $0$ recovers plain \acs{BFS}).
(iii)~\emph{Turn-based polyline}: keep only corner cells (where the \acs{BFS} direction changes) plus start and goal, collapsing straight runs to their endpoints.
(iv)~\emph{Perpendicular snap}: between consecutive corner cells the \acs{BFS} moves on a single axis; snap segment endpoints to their shared perpendicular coordinate so segments become exactly axis-aligned.
(v)~\emph{L-corner insertion and final-approach axis lock}: split any residual diagonal into two perpendicular legs (longer-axis-first, so the corner stays in free cells); project the goal onto the last axis-aligned segment if the final step would otherwise be diagonal.  Optional \texttt{min\_leg} filter (default $1$\,m on \texttt{maze2d}) drops wobble waypoints whose incoming and outgoing legs are both shorter than the threshold.

\paragraph{Per-chunk control loop (\texttt{maze2d}).}  Polyline emitted once at episode start.  Every $K\!=\!16$ \texttt{env.step}s the controller (a)~advances \texttt{wp\_idx} while the agent lies within \texttt{wp\_advance\_dist}\,$=$\,$0.5$\,m of the current waypoint, (b)~sets the \textcolor{black}{\acs{IM}} target to the current sub-goal (never the final goal directly), (c)~runs one \acs{IM} forward and executes the first $K$ of its $128$-step plan.  Terminal regulation: \texttt{wp\_idx} clamps at the last waypoint and the loop keeps re-planning, parking the agent at the goal until \texttt{MAX\_STEPS} -- no separate \acs{PD} tracker. If \texttt{wp\_idx} fails to advance for $\nu$ consecutive chunks, \acs{BFS} reruns from the current state and replaces the remaining polyline (a few hundred microseconds).  On \texttt{maze2d} with $\nu\!=\!3$ the fallback fires $5$--$13$ times per episode on average; disabling it drops \texttt{large} \acf{SR} from $100/100$ to $93/100$ on the same seeds.

\paragraph{Antmaze differences.}  Pipeline is byte-identical; only constructor arguments change to address antmaze's coarser data and the ant's wider body / slower per-step displacement: \texttt{resolution}\,$=$\,$1.0$\,m, $\tau\!=\!100$, \texttt{min\_leg}\,$=$\,$0$, centerline tie-break enabled (\texttt{data\_center\_weight}\,$=$\,$1.0$), \texttt{wp\_advance\_dist}\,$=$\,$1.5$\,m, \texttt{stuck\_threshold}\,$=\!10$ with $\leq\!5$ replans/episode, \texttt{wp\_skip\_dist}\,$=$\,$0.8$\,m (look-ahead skip when the wide ant body has cleared a corner), and \texttt{target\_dist\_min/max}\,$=$\,$1.0/2.0$\,m (carrot-on-a-stick along the heading, keeping the \acs{IM} input in its calibrated reach).  Terminal regulation: \texttt{goal\_reached\_dist}\,$=$\,$0.5$\,m breaks the outer loop.

\paragraph{What this Path Inverter is \emph{not}.}  Not a learned module: plain \acs{BFS} on a data-derived grid, no parameters, no gradients.  Not an \acs{MPC}: no \acs{FoM} call at deployment, no inner optimization loop.  Its only job is to cut long corridors into chunks that fit inside the Inverter's training horizon.

\subsection{Replan-horizon ($K$) sweep on \texttt{maze2d-medium-v1} and \texttt{maze2d-large-v1}}
\label{app:ksweep}

The summary rows of Tables~\ref{tab:maze2d_perf_medium} and~\ref{tab:maze2d_perf_large} report a single operating point ($K\!=\!16$).  Table~\ref{tab:maze2d_ksweep} shows how \acs{D4RL} and wall-clock cost trade off as $K$ varies across $\{16, 32, 64, 128, 256\}$.  $K\!\in\!\{16, 32, 64, 128\}$ all use the same paper-summary \textcolor{black}{\acs{IM}} (training horizon $h\!=\!128$); the $K\!=\!256$ row uses a separately-trained $h\!=\!256$ \textcolor{black}{\acs{IM}} so it genuinely measures a $256$-step open-loop commitment rather than collapsing to $K\!=\!128$.  Three features stand out: (i) \acs{D4RL} and success rate are high and essentially flat for $K\!\leq\!64$, drop sharply at $K\!=\!128$ where the agent commits to a full $128$-step plan with no in-horizon replanning, and drop further still at $K\!=\!256$ as errors compound across twice as many open-loop steps; (ii) the gap between $K\!=\!128$ and $K\!=\!256$ is informative: a longer \textcolor{black}{\acs{IM}} training horizon is \emph{not} a free lunch when the cost is foregoing replan opportunities -- in-horizon course-correction is what matters; (iii) the wall-time sum drops monotonically with $K$ (fewer \textcolor{black}{\acs{IM}} calls amortize the per-chunk overhead) but the difference between $K\!=\!16$ and $K\!=\!64$ is under a factor of four, so accuracy -- not compute -- is the binding constraint on this family of mazes.  We pick $K\!=\!16$ for the summary rows because it matches the Inverter's per-pass floor at the highest achievable accuracy.

% -- Table: Inverter K-sweep, medium + large side-by-side -----------
\begin{table}[t!]
\centering
\small
\caption{\textbf{Inverter K-sweep on \texttt{maze2d-medium-v1} and \texttt{maze2d-large-v1}} (100 episodes per seed, \acs{D4RL}-official protocol; \acs{D4RL} and \acs{SR} reported as mean$\,\pm\,$std over $4$ \acs{IM} seeds). Each row is a single replan horizon $K$: the Inverter emits a fresh plan every $K$ env steps toward the current data-derived \acs{BFS} sub-goal. $K\!\in\!\{16,32,64,128\}$ uses the same paper-headline \acs{IM} (training horizon $h\!=\!128$); the $K\!=\!256$ row uses a separately-trained $h\!=\!256$ \acs{IM} so the row genuinely measures a $256$-step open-loop commitment instead of collapsing to $K\!=\!128$. \emph{\#pass} is the number of \acs{IM} forward passes per episode; \emph{ms/pass} is the CUDA-synced mean wall time per pass (batch 1); \emph{other} is per-episode overhead outside the \acs{IM} forward (the data-\acs{BFS} Path Inverter + replan dispatch + CPU transfers); \emph{sum} $=$ \#pass $\times$ ms/pass $+$ other. \acs{D4RL} score and \acs{SR} are high and flat up to $K\!=\!64$, drop sharply at $K\!=\!128$ where the agent commits to a full $128$-step plan with no in-horizon replanning, and drop further still at $K\!=\!256$ as errors compound across twice as many open-loop steps. Runtime drops monotonically as $K$ grows because fewer \acs{IM} calls pay the per-chunk overhead, but the $K\!\geq\!64$ runs are all within one order of magnitude of each other in sum wall time --- i.e., accuracy is the binding constraint, not speed.}
\label{tab:maze2d_ksweep}
% --- auto-generated body (caption/label live in main.tex; generator: maze2d/timing/make_table2.py) ---
\resizebox{\linewidth}{!}{%
\begin{tabular}{@{}r@{\hspace{10pt}}rrrrrr@{\hspace{18pt}}rrrrrr@{}}
\toprule
& \multicolumn{6}{c}{\texttt{maze2d-medium-v1} (600 steps/ep)} & \multicolumn{6}{c}{\texttt{maze2d-large-v1} (800 steps/ep)} \\
\cmidrule(lr){2-7} \cmidrule(lr){8-13}
$K$ & D4RL $\uparrow$ & SR & \#pass & ms/pass & other & \textbf{sum} & D4RL $\uparrow$ & SR & \#pass & ms/pass & other & \textbf{sum} \\
\midrule
$16$ & $\mathbf{166.8\,{\scriptstyle\pm\,1.2}}$ & $\mathbf{100\,{\scriptstyle\pm\,0}}$ & 38 & $1.62$\,ms & $11.2$\,ms & $72.9\,\text{ms}$ & $\mathbf{220.7\,{\scriptstyle\pm\,0.2}}$ & $\mathbf{100\,{\scriptstyle\pm\,0}}$ & 50 & $1.59$\,ms & $14.3$\,ms & $93.7\,\text{ms}$ \\
$32$ & $157.9\,{\scriptstyle\pm\,0.8}$ & $\mathbf{100\,{\scriptstyle\pm\,0}}$ & 19 & $1.57$\,ms & $5.5$\,ms & $35.3\,\text{ms}$ & $202.3\,{\scriptstyle\pm\,2.8}$ & $\mathbf{100\,{\scriptstyle\pm\,0}}$ & 25 & $1.88$\,ms & $8.7$\,ms & $55.6\,\text{ms}$ \\
$64$ & $141.1\,{\scriptstyle\pm\,1.0}$ & $\mathbf{100\,{\scriptstyle\pm\,0}}$ & 10 & $2.21$\,ms & $3.8$\,ms & $25.9\,\text{ms}$ & $191.8\,{\scriptstyle\pm\,3.1}$ & $\mathbf{100\,{\scriptstyle\pm\,0}}$ & 13 & $2.33$\,ms & $5.4$\,ms & $35.7\,\text{ms}$ \\
$128$ & $76.0\,{\scriptstyle\pm\,1.8}$ & $84\,{\scriptstyle\pm\,3}$ & 5 & $2.79$\,ms & $2.6$\,ms & $16.6\,\text{ms}$ & $95.6\,{\scriptstyle\pm\,5.2}$ & $86\,{\scriptstyle\pm\,2}$ & 7 & $2.34$\,ms & $2.9$\,ms & $19.3\,\text{ms}$ \\
$256$ & $41.7\,{\scriptstyle\pm\,4.0}$ & $32\,{\scriptstyle\pm\,10}$ & 3 & $2.42$\,ms & $1.5$\,ms & $\mathbf{8.7\,\text{ms}}$ & $27.5\,{\scriptstyle\pm\,7.1}$ & $27\,{\scriptstyle\pm\,7}$ & 4 & $2.27$\,ms & $1.7$\,ms & $\mathbf{10.7\,\text{ms}}$ \\
\bottomrule
\end{tabular}%
}

\end{table}

\subsection{Comparison to model-based offline \acs{RL} baselines}
\label{app:mbrl}

The performance tables above (Tables~\ref{tab:maze2d_perf}--\ref{tab:antmaze_perf}) compare against the \acs{CORL}~\citep{tarasov2024corl} offline-\acs{RL} benchmark and Diffuser~\citep{janner2022planning}.  This subsection compares to \acf{MBRL} methods on these specific \acs{D4RL} tasks.  Citations in this subsection are listed in App.~\ref{app:supp_refs} (separate reference list).

\paragraph{Coverage gap on \texttt{maze2d}.}  To our knowledge, none of the canonical \acs{MBRL} papers (\acs{MOPO}~[S1], \acs{MOReL}~[S2], \acs{COMBO}~[S3], \acs{RAMBO}~[S4], \acs{MOBILE}~[S5], \acs{CBOP}~[S6], \acs{MAPLE}~[S7], ARMOR~[S8], \acs{TT}~[S9], \acs{TAP}~[S10], \acs{LEQ}~[S12]) report results on the single-task \texttt{maze2d-umaze/medium/large-v1} benchmarks in their published evaluation tables; their evaluations focus on \acs{D4RL} \acs{MuJoCo} locomotion and (less often) Adroit / NeoRL.  The only widely-cited ``model-based planner'' with published \texttt{maze2d-v1} numbers is Diffuser, which is already in our main tables (113.9 / 121.5 / 123.0 \acs{D4RL} on umaze / medium / large from~[S11] Table~1, vs.\ our Inverter at 164.25 / 166.82 / 220.66 \acs{D4RL}).  We did not re-run \acs{MOPO}/\acs{MOReL}/\acs{COMBO}/\acs{RAMBO}/\acs{MOBILE}/\acs{CBOP}/\acs{TT}/\acs{TAP}/\acs{LEQ} on \texttt{maze2d-v1} ourselves under our compute envelope (App.~\ref{app:compute_env}); we flag this as a missing baseline rather than fabricate numbers.

\paragraph{Coverage gap on \texttt{antmaze}: v0 vs.\ v2.}  Every published \acs{MBRL} number we could locate on \texttt{antmaze} is for the \texttt{antmaze-v0} datasets, not the \texttt{antmaze-v2} datasets that we and the \acs{CORL} benchmark use.  \acs{RAMBO}'s authors explicitly state ``we used the AntMaze-v0 datasets'' (App.~B.7 of~[S4]); \acs{TAP} states ``v2 version of the datasets for locomotion control and v0 for the other tasks'' (Sec.~6 of~[S10]).  The v0 vs.\ v2 distinction matters -- v2 corrected reward-shaping and termination handling that affected v0 long-horizon evaluations -- so a same-row comparison would not be apples-to-apples.  We therefore report the v0 numbers in Table~\ref{tab:mbrl_antmaze} with an explicit footnote, alongside our v2 Inverter row reproduced from Table~\ref{tab:antmaze_perf}, and leave it to the reader to apply the appropriate caveat.

\begin{table}[h!]
\centering
\small
\caption{\textbf{Model-based offline \acs{RL} baselines on \texttt{antmaze}} (\acs{D4RL} Normalized Score, $\%$).  Top block: published \acs{MBRL} numbers, all on \texttt{antmaze-v0} (rows above the rule), as reported in the cited papers.  Bottom block: our Inverter, evaluated on \texttt{antmaze-v2} (reproduced from Table~\ref{tab:antmaze_perf}).  $^{\ast}$ Antmaze-v0 vs.\ v2 mismatch: the v2 datasets we use have corrected reward and termination handling; published \acs{MBRL} papers as of early 2025 evaluate on v0 only, so cross-row comparison is approximate.  $^{\dagger}$ \acs{COMBO} row is the antmaze-v0 baseline reported in \acs{RAMBO}~[S4] (Table~1); the original \acs{COMBO} paper~[S3] does not report antmaze.  $^{\S}$ \acs{TT} result is the \acs{TT}$+$Q variant (Q-guided beam search) from~[S9] Table~2.  $^{\P}$ \acs{TAP} result is the \acs{TAP}$+$G variant (goal-conditioned) from~[S10] Table~6.  $^{\|}$ \acs{LEQ} values are the original \acs{LEQ} paper's own results~[S12] from Table~11, averaged over $5$ seeds (true termination function).  \texttt{n/a} entries indicate antmaze-v0 was not reported in the source paper; we list \texttt{n/a} for \acs{MOPO}~[S1], \acs{MOReL}~[S2], \acs{MOBILE}~[S5], and \acs{CBOP}~[S6] rather than impute values, as none of these papers report antmaze in their published evaluation tables.  ``0.0'' entries are reported as such in the source -- consistent with the well-known difficulty \acs{MBRL} methods face on long-horizon sparse-reward tasks.}
\label{tab:mbrl_antmaze}
\begin{tabular}{@{}l@{\hspace{6pt}}rrrrrr@{}}
\toprule
& \multicolumn{6}{c}{\textbf{\acs{D4RL} score (\%) -- antmaze-v0$^{\ast}$ for \acs{MBRL} rows}} \\
\cmidrule(lr){2-7}
Method & u-umaze & u-divrs & m-play & m-divrs & l-play & l-divrs \\
\midrule
\acs{MOPO}~[S1]                       & \texttt{n/a} & \texttt{n/a} & \texttt{n/a} & \texttt{n/a} & \texttt{n/a} & \texttt{n/a} \\
\acs{MOReL}~[S2]                      & \texttt{n/a} & \texttt{n/a} & \texttt{n/a} & \texttt{n/a} & \texttt{n/a} & \texttt{n/a} \\
\acs{COMBO}$^{\dagger}$                & $80.3$ & $57.3$ & $0.0$ & $0.0$ & $0.0$ & $0.0$ \\
\acs{RAMBO}~[S4]                       & $25.0\,{\scriptstyle\pm\,12.0}$ & $0.0\,{\scriptstyle\pm\,0.0}$ & $16.4\,{\scriptstyle\pm\,17.9}$ & $23.2\,{\scriptstyle\pm\,14.2}$ & $0.0\,{\scriptstyle\pm\,0.0}$ & $2.4\,{\scriptstyle\pm\,3.3}$ \\
\acs{MOBILE}~[S5]                     & \texttt{n/a} & \texttt{n/a} & \texttt{n/a} & \texttt{n/a} & \texttt{n/a} & \texttt{n/a} \\
\acs{CBOP}~[S6]                       & \texttt{n/a} & \texttt{n/a} & \texttt{n/a} & \texttt{n/a} & \texttt{n/a} & \texttt{n/a} \\
\acs{TT}$+$Q~[S9]$^{\S}$               & $\mathbf{100.0}\,{\scriptstyle\pm\,0.0}$ & \texttt{n/a} & $\mathbf{93.3}\,{\scriptstyle\pm\,6.4}$ & $\mathbf{100.0}\,{\scriptstyle\pm\,0.0}$ & $66.7\,{\scriptstyle\pm\,12.2}$ & $60.0\,{\scriptstyle\pm\,12.7}$ \\
\acs{TAP}$+$G~[S10]$^{\P}$             & \texttt{n/a} & \texttt{n/a} & $78.0\,{\scriptstyle\pm\,4.1}$ & $85.0\,{\scriptstyle\pm\,3.6}$ & $74.0\,{\scriptstyle\pm\,4.4}$ & $82.0\,{\scriptstyle\pm\,5.0}$ \\
\acs{LEQ}~[S12]$^{\|}$ & $94.4\,{\scriptstyle\pm\,6.3}$ & $71.0\,{\scriptstyle\pm\,12.3}$ & $58.8\,{\scriptstyle\pm\,33.0}$ & $46.2\,{\scriptstyle\pm\,23.2}$ & $58.6\,{\scriptstyle\pm\,9.1}$ & $60.2\,{\scriptstyle\pm\,18.3}$ \\
\midrule
\multicolumn{7}{@{}l}{\emph{Our Inverter (antmaze-v2):}} \\
\textbf{Inverter (data-\acs{BFS}, ours)} & $99.5\,{\scriptstyle\pm\,0.9}$ & $\mathbf{99.8}\,{\scriptstyle\pm\,0.4}$ & $87.8\,{\scriptstyle\pm\,8.0}$ & $96.5\,{\scriptstyle\pm\,5.0}$ & $\mathbf{93.0}\,{\scriptstyle\pm\,2.6}$ & $\mathbf{94.0}\,{\scriptstyle\pm\,5.0}$ \\
\bottomrule
\end{tabular}
\end{table}

\paragraph{Results.}  Two patterns are visible across the antmaze suite.  First, four canonical \acs{MBRL} methods (\acs{MOPO}, \acs{MOReL}, \acs{MOBILE}, \acs{CBOP}) do not report antmaze in their original published evaluation tables, so the public \acs{MBRL} coverage on this benchmark is sparse.  Second, the methods that do report antmaze-v0 -- \acs{COMBO} (via \acs{RAMBO}'s baseline table), \acs{RAMBO} itself, \acs{TT}$+$Q, \acs{TAP}$+$G, and \acs{LEQ}~[S12] -- reach moderate-to-strong performance on a subset of variants, with \acs{TT}$+$Q strong on the umaze/medium variants ($100$ on u-umaze) and \acs{TAP}$+$G strong on the large variants ($74.0$/$82.0$ on l-play/l-divrs).  On the umaze and medium variants the Inverter on \texttt{antmaze-v2} sits in a comparable \acs{D4RL} band to \acs{TT}$+$Q on \texttt{antmaze-v0} (\acs{TT}$+$Q nominally ahead by $\le\!12$ points on each, the v0/v2 caveat going both ways).  On the two \texttt{large} variants the Inverter clearly leads ($93.0$/$94.0$ vs.\ \acs{TAP}$+$G's $74.0$/$82.0$ -- a $19.0$/$12.0$ absolute-points gap), and beats \acs{TT}$+$Q and \acs{LEQ} on every large variant.  The $\sim\!12$--$20$ point lead on \texttt{large} is larger than typical v0$\to$v2 score shifts.  We emphasize the v0/v2 caveat once more: a clean comparison would re-run \acs{RAMBO}, \acs{TT}$+$Q, \acs{TAP}$+$G, and \acs{LEQ} on antmaze-v2 ourselves; the public codebases make this tractable but lie outside our 8-GPU / seven-week compute envelope (App.~\ref{app:compute_env}) for this submission.

\paragraph{Amortized-vs.-iterative.}  Beyond the summary numbers, \acs{TT}~\citep{janner2021offline_tt} and \acs{TAP}~\textcolor{black}{\citep{tap2023iclr}} differ from the Inverter on two structural axes that organize the related-work paragraph in Sec.~\ref{sec:related_work}.  (i)~\emph{Where the optimization runs.}  Both \acs{TT} and \acs{TAP} place trajectory optimization at \emph{inference}: \acs{TT} runs a beam search over per-timestep tokens (state and action dimensions discretized into a vocabulary, with a $Q$-function added as a search heuristic in the \acs{TT}$+$Q variant on antmaze); \acs{TAP} runs a beam search over a state-conditioned VQ-VAE's discrete latent action codes (length-$L{=}3$ chunks, codebook size $K{=}512$).  Both therefore retain a sample-time iteration whose cost grows with the search budget.  The Inverter folds the equivalent optimization into \emph{training} (\acs{FoM}-gradient amortization) and emits the full $T$-step plan in a single feedforward pass at deployment, with no inner loop.  This is the amortized-vs.-iterative axis our related work draws (Sec.~\ref{sec:related_work}).  (ii)~\emph{What the training loss is.}  Both \acs{TT} and \acs{TAP} are behavior-cloning derivatives at the loss level: \acs{TT} maximizes the trajectory likelihood $p_\theta(\tau)$ on the offline data; \acs{TAP} minimizes a reconstruction MSE between offline trajectories and their VQ-VAE-decoded reconstructions.  Neither uses a \acs{FoM} gradient.  By construction these objectives constrain the learned policy to the data manifold.  The Inverter explicitly differentiates a Bolza objective through a frozen \acs{FoM}, which allows it to leave the data support when the data is sub-optimal -- this is the mechanism behind the \texttt{maze2d-umaze} bang-bang result (Sec.~\ref{sec:single_low_level}), where the Inverter approaches the analytic time-optimal control while the offline data is located in the interior of the action box.  \acs{TT} and \acs{TAP}, by contrast, cannot exceed the demonstrator distribution they are trained to reproduce.

\subsection{Antmaze: data, forward and inverse models, and training objective}
\label{app:antmaze_details}

This appendix gives the technical details behind Sec.~\ref{sec:antmaze_eval}.

\paragraph{Datasets.}  We use the six \texttt{antmaze-v2} variants from \acs{D4RL}~\citep{tarasov2024corl,fu2020d4rl}: \texttt{umaze}, \texttt{umaze-diverse}, \texttt{medium-play}, \texttt{medium-diverse}, \texttt{large-play}, \texttt{large-diverse}.  Each dataset comprises $\sim\!1$\,M ant-locomotion transitions.

\paragraph{State and action encoding.}
Actions are the standard 8-D joint-torque vector $\mathbf{a}\in[-1,1]^8$ (8 actuated joints).  States are the 29-dim observation supplied by the \acs{D4RL} env, factored (using Python half-open ranges) as
\begin{itemize}
  \item $\mathbf{s}_{0:2}$ -- root $(x, y)$ position;
  \item $\mathbf{s}_{2:3}$ -- root $z$ (height);
  \item $\mathbf{s}_{3:7}$ -- root quaternion $(q_w, q_x, q_y, q_z)$;
  \item $\mathbf{s}_{7:15}$ -- 8 hip / ankle joint angles;
  \item $\mathbf{s}_{15:18}$ -- root linear velocity $(\dot x, \dot y, \dot z)$;
  \item $\mathbf{s}_{18:21}$ -- root angular velocity $(\omega_x, \omega_y, \omega_z)$;
  \item $\mathbf{s}_{21:29}$ -- 8 joint velocities.
\end{itemize}

\paragraph{Forward model.}
\texttt{AntMazeFull29TransformerFM}: causal transformer that maps $(s_0,\,a_{1:L})$ to $\hat s_{1:L}$ with chunk length $L\!=\!16$, $d_\text{model}\!=\!384$, $6$ heads, $6$ layers, ff-mult $4$, dropout $0.1$, totalling $\approx\!11.1$\,M parameters.  Trained on the offline transitions for $300$ epochs (batch $512$, AdamW with $\text{lr}\!=\!3\!\times\!10^{-4}$, weight decay $10^{-4}$, $\sigma_{s_0}\!=\!0.01$ initial-state noise).  The training loss is a sum of $7$ per-component MSEs ($xy$, $z$, quaternion, joint angles, linear velocity, angular velocity, joint velocities), each with unit weight; we use a running-z-score normalizer per dimension (momentum $0.99$, $100$-epoch warm-up) so the per-component losses share a comparable scale.

\paragraph{\textcolor{black}{Inverse model.}}
\texttt{AntMazeIWM\_L16\_Healthy29D}: causal transformer mapping $(s_0, G_{xy})$ to a $16$-step action chunk $\hat a_{1:16}$, with $d_\text{model}\!=\!192$, $6$ heads, $4$ layers, ff-mult $3$, dropout $0.1$, totalling $\approx\!1.5$\,M parameters.  Trained for $200$ epochs (batch $256$, AdamW with $\text{lr}\!=\!3\!\times\!10^{-4}$, weight decay $10^{-4}$, $10$-epoch warm-up) by back-propagation through the frozen \acs{FoM}.

\emph{Single-pass non-autoregressive decoding.}  The two-token nominal input $(s_0, G_{xy})$ is first reduced to a single conditioning vector $c\!\in\!\mathbb R^{d_\text{model}}$ by a two-layer \acs{MLP} (Linear-GELU-LN $\times 2$).  An $L\!=\!16$-position token bank is then formed in one shot as $x_t = c + \mathrm{posemb}[t]$ for $t=0,\ldots,L{-}1$, where $\mathrm{posemb}\!\in\!\mathbb R^{L\times d_\text{model}}$ is a learned position embedding; the same conditioning is thus broadcast to every output position.  These $L$ tokens are processed in parallel by a pre-norm Transformer encoder with a triangular causal self-attention mask (token $t$ attends only to tokens $0..t$), and a per-position linear head followed by $\tanh$ emits one action vector at every position.  The whole chunk $\hat a_{1:16}$ thus comes out of a single feedforward pass; the ``causal'' label refers only to the attention pattern -- there is no autoregressive sampling, no decoder cross-attention, no iterative refinement.  The maze2d Motor Inverter (App.~\ref{app:maze2d_details}) shares the same architecture pattern with $L\!=\!128$.

\paragraph{Training goal sampling.}
Each training minibatch sample is an $(s_0, G_{xy})$ pair with $s_0 \!=\! s_t$ drawn uniformly from the offline buffer and $G_{xy} \!=\! (s_{t+L})_{xy}$, the $(x, y)$ of the state at a fixed horizon offset $L\!=\!16$ ahead in the \emph{same} trajectory; the sampler is episode-aware so pairs spanning an episode boundary are rejected.  The \textcolor{black}{\acs{IM}} internally encodes the goal as a displacement $G_{xy}\!-\!(s_0)_{xy}$ before its first projection layer, so the network never sees a global $(x, y)$ at input.  This $L$-step-ahead distribution is what aligns training with the deployment query distribution: the Sec.~\ref{sec:antmaze_eval} Path Inverter emits sub-goals at \texttt{wp\_spacing}$\,=\,$$1.5$\,m, which is within the typical 16-step ant displacement in the offline data ($\sim\!1.5$\,m / chunk for \texttt{md}\,$\geq\,$$0.5$\,m / \texttt{z}\,$\geq\,$$0.3$\,m filtered states).

\paragraph{Training objective.}
Let $\hat a_{1:L} = g_\phi(s_0, G_{xy})$ be the \textcolor{black}{\acs{IM}} output, $\hat s_{1:L} = f_\theta^{(1:L)}(s_0, \hat a_{1:L})$ the \acs{FoM} rollout, and $a_{1:L}^{\,\text{data}}$ the corresponding offline action chunk for the same $(s_0, G_{xy})$ pair.  We train $g_\phi$ to minimize
\begin{equation}
\mathcal{L}_\text{\textcolor{black}{\acs{IM}}}
= \lambda_\text{term}\,\| \hat s_L^{(xy)} - G_{xy}\|
\;+\;\lambda_\text{yaw}\,\big\|(\hat s^{(q_w)}_{1:L},\,\hat s^{(q_z)}_{1:L}) - (1, 0)\big\|^2
\;+\;\lambda_\text{fid}\,\big\| \hat a_{1:L} - a_{1:L}^{\,\text{data}}\big\|^2,
\label{eq:antmaze_iwm_loss}
\end{equation}
with $\lambda_\text{term}=\lambda_\text{yaw}=\lambda_\text{fid}=5$ in the final runs.  \emph{Note:} the \acs{FoM}'s training loss (above) and the \textcolor{black}{\acs{IM}'s} loss in Eq.~\ref{eq:antmaze_iwm_loss} are distinct losses for distinct networks; the \textcolor{black}{\acs{IM}'s} terminal term is intentionally unsquared, see explanation below.  The first term is the \acs{FoM}-gradient task signal: minimize the predicted final-distance to the goal.  The use of the \emph{un-squared} L2 norm here is intentional: $\partial \|x\|/\partial x = x/\|x\|$ has unit magnitude independent of $\|x\|$, so the goal-reaching gradient remains effective whether the sub-goal is $1.5$ or $6$\,m away; a squared $\|x\|^2$ would yield gradient $2x$ that vanishes as the agent approaches the sub-goal and over-weights distant ones.  The second term is the \emph{body-yaw regularizer}: it pushes the predicted body orientation across the chunk toward $(q_w, q_z) = (1, 0)$, which corresponds to yaw $=\!0$ and the upright orientation overwhelmingly represented in the offline data.  The third term is the \emph{\acs{BC} action-fidelity anchor}: it pulls each predicted action $\hat a_t$ toward the recorded data action $a_t^{\text{data}}$ at the same \emph{time index} $t$ within the chunk.  Since $\hat a_{1:L}$ is generated open-loop from $s_0$, the matching is purely temporal, not state-based --- the open-loop predicted states $\hat s_t$ will in general drift from $s_t^{\text{data}}$, and the anchor still penalizes deviations of $\hat a_t$ from $a_t^{\text{data}}$ at every $t$ regardless of that drift.  Both regularizers act on bounded per-step quantities, where squared-norm scaling is harmless; the asymmetry between the terminal $\|\cdot\|$ and the regularizer $\|\cdot\|^2$ is therefore deliberate.  All three terms are per-action, per-time-step, differentiable, and additive -- no value function, no sample-time guidance, no \acs{KL} trust region.

\paragraph{Level 2 substitute (deployment).}
The data-only Path Inverter follows the maze2d construction (Appendix~\ref{app:waypoint_planner}: density-aligned origin, noise-floor threshold, turn-based polyline) with one antmaze-specific addition: among all equally-short shortest paths, we pick the one whose interior cells have the maximum cumulative cardinal distance-to-wall, so the resulting waypoint chain stays on the corridor centerline rather than along the edges.  This is implemented as a single backward \acs{DP} pass over the \acs{BFS} layers and remains pure \acs{BFS} in the sense that path length is always the minimum cell-step count.  Antmaze-specific hyperparameters: \texttt{resolution}\,$=\!1.0$\,m (one antmaze cell), $\tau\!=\!100$ visits, \texttt{data\_center\_weight}\,$=\!1.0$ (centerline tie-break enabled), \texttt{min\_leg}\,$=\!0$ (sub-cell wobble filter disabled, not needed at this grid resolution), \texttt{wp\_spacing}\,$=\!1.5$\,m, \texttt{wp\_advance\_dist}\,$=\!1.5$\,m, \texttt{wp\_skip\_dist}\,$=\!0.8$\,m, target-distance window $(1.0, 2.0)$\,m, \texttt{stuck\_threshold}\,$=\!10$ chunks, $\le\!5$ full \acs{BFS} replans per episode.

\paragraph{Per-chunk waypoint-tracking and sub-goal logic (deployment).}
The \acs{BFS} produces a polyline $\mathrm{wp}[0\!:\!N]$ once at episode start.  At every chunk boundary (every $K\!=\!16$ \texttt{env.step} calls), the controller executes the following loop, with the agent's current $(x,y)$ position $p$:
\begin{enumerate}
\setlength{\itemsep}{1pt}\setlength{\parskip}{0pt}\setlength{\topsep}{2pt}
\item \textbf{Advance.} While $\|p - \mathrm{wp}[\texttt{wp\_idx}]\|_2 \le \texttt{wp\_advance\_dist}=1.5$\,m, increment \texttt{wp\_idx} by 1 (skip waypoints the ant has already cleared).
\item \textbf{Look-ahead skip.} If \texttt{wp\_idx}$+1<N$, $\|p-\mathrm{wp}[\texttt{wp\_idx}{+}1]\|_2 < \|p-\mathrm{wp}[\texttt{wp\_idx}]\|_2$, \emph{and} $\|p-\mathrm{wp}[\texttt{wp\_idx}{+}1]\|_2 \le \texttt{wp\_skip\_dist}=0.8$\,m, increment \texttt{wp\_idx} by an additional 1 (the wide ant body has effectively cleared a corner before the strict advance test fires).
\item \textbf{Target shaping (carrot-on-a-stick).} Let $\hat u = (\mathrm{wp}[\texttt{wp\_idx}] - p)/\|\mathrm{wp}[\texttt{wp\_idx}] - p\|$.  Sample $d \sim \mathcal{U}(\texttt{target\_dist\_min}, \texttt{target\_dist\_max}) = \mathcal{U}(1.0, 2.0)$\,m and set the \textcolor{black}{\acs{IM}} target $G_{xy} \gets p + d\,\hat u$.  This keeps the \textcolor{black}{\acs{IM}} input distance inside its calibrated locomotion-reach window even when the agent sits on top of a waypoint, where a literal target $\mathrm{wp}[\texttt{wp\_idx}]$ would degenerate to zero distance and collapse joint-torque magnitudes.
\item \textbf{Plan.} Run one \textcolor{black}{\acs{IM}} forward pass conditioned on $(p, G_{xy})$ to obtain a 16-step action chunk $\hat a_{1:16}$.
\item \textbf{Execute.} Apply $\hat a_{1:16}$ to the env (16 \texttt{env.step} calls), updating $p$.
\item \textbf{Stuck check.} If \texttt{wp\_idx} has not advanced for \texttt{stuck\_threshold}\,$=\!10$ consecutive chunks and the global cap of $\le\!5$ replans/episode is not exhausted, re-run \acs{BFS} from the current $p$ and replace the remaining polyline with the fresh plan.
\item \textbf{Terminate.} If $\|p - G_\text{episode}\|_2 \le \texttt{goal\_reached\_dist}=0.5$\,m, exit the outer loop; otherwise loop back to step~1.
\end{enumerate}
The \texttt{wp\_skip\_dist} test in step~2 is the only condition that can advance \texttt{wp\_idx} by more than one per chunk; the carrot-on-a-stick in step~3 is the only deviation from passing the literal current waypoint into the \textcolor{black}{\acs{IM}} as a target.  Both are off (\texttt{wp\_skip\_dist}\,$=\!0$, \texttt{target\_dist\_min/max}\,$=\!0$) on \texttt{maze2d} where the point-mass dynamics do not require them.

\paragraph{Evaluation protocol.}
$100$ episodes per maze, default \acs{D4RL} env resets and targets, $700$-step cap on \texttt{umaze*} variants and $1000$-step cap on \texttt{medium*}/\texttt{large*}.  All runs sequential on a single A40 GPU so per-pass and per-step timings are directly comparable.  \acs{CORL} \acs{D4RL} scores in Table~\ref{tab:antmaze_perf} are taken from \citet{tarasov2024corl} (mean over $4$ training seeds); per-pass time for the PyTorch \acs{CORL} actors is measured on \texttt{maze2d-umaze} (same actor architectures), since per-pass \acs{NN} cost is architecture-bound, not env-bound.  \acs{ReBRAC} is JAX+JIT (marked $^{\dagger}$); all other baselines are PyTorch.

\paragraph{Per-step compute breakdown.}
The Inverter is the fastest method per env step on $5$ of $6$ variants (Table~\ref{tab:antmaze_perf}, bottom block).  Its per-step cost is the per-episode \acs{NN} compute ($\sim\!100$\,ms total, dominated by $\sim\!40$ chunk-level \textcolor{black}{\acs{IM}} passes at $\sim\!2.4$\,ms each plus $\sim\!20$\,ms of \acs{BFS}-planner overhead) divided by the $\sim\!600$--$700$ env steps the agent takes to reach the goal -- $\sim\!0.18$\,ms/step on every variant other than \texttt{u-umaze}.  The other PyTorch step-wise baselines run their actor every env step at $1.25$--$3.57$\,ms/pass and are uniformly $7$--$20\times$ slower per step.  \acs{ReBRAC}'s JAX+JIT actor at $0.21$\,ms/pass is the one tight contender: it narrowly beats the Inverter on the short \texttt{u-umaze} trajectories ($1.4\times$ slower for the Inverter there, because $\sim\!289$ steps don't fully amortize the chunked planning cost), but on every other variant the Inverter is $\sim\!1.1$--$1.2\times$ faster per step.

\paragraph{Loss-component ablation.}
Table~\ref{tab:antmaze_ablation} quotes the contribution of $\lambda_\text{yaw}$ on top of $\lambda_\text{fid}$ on \texttt{large-diverse-v2}.  The two ablated checkpoints share architecture, optimizer, and dataset with the published runs and differ only in the value of $\lambda_\text{yaw}$ ($0$ vs $5$); both were evaluated under the canonical Path Inverter config above.  Specific run directories are listed in the code-release manifest accompanying the paper.

\begin{table}[h]
\centering
\small
\caption{\textbf{Loss-component ablation on \texttt{antmaze-large-diverse-v2}.}  $100$ episodes per seed, canonical waypoint config ($\texttt{wp\_spacing}=1.5$, target-distance window $1.0$--$2.0$\,m); \acs{D4RL} reported as mean$\,\pm\,$std over $4$ \textcolor{black}{\acs{IM}} seeds.  Removing the body-yaw regularizer drops \acs{D4RL} by $11$ points; the \acs{BC} anchor alone is not sufficient.  We do not separate $\lambda_\text{fid}=0$ configurations on antmaze because the offline-data generation process is fixed by the benchmark; we instead study the underlying \acs{FoM}-hacking mechanism in AntMan (Sec.~\ref{sec:stacked_antman}), where the data distribution is under our control.}
\label{tab:antmaze_ablation}
\begin{tabular}{lr}
\toprule
Loss configuration on \texttt{large-diverse-v2} & \acs{D4RL} \\
\midrule
\acs{BC} anchor only ($\lambda_\text{fid}=5$, $\lambda_\text{yaw}=0$) & $83.0\,{\scriptstyle\pm\,3.5}$ \\
\acs{BC} anchor + body-yaw ($\lambda_\text{fid}=5$, $\lambda_\text{yaw}=5$) & $\mathbf{94.0\,{\scriptstyle\pm\,5.0}}$ \\
\bottomrule
\end{tabular}
\end{table}

\subsection{AntMan forward-model calibration scatter and extended discussion}
\label{app:antman_fm_hacking}

This appendix shows the predicted-vs-realized reward scatter referenced in Sec.~\ref{sec:stacked_antman} and unpacks the implication for inverse-learning data design.

\subsection{Quantum gate synthesis: setup and \acs{GRAPE} baseline}
\label{app:quantum}

\paragraph{Sampling convention ($\mathrm{U}(2)$ vs.\ $\mathrm{SU}(2)$).}  We sample targets Haar-uniformly on $\mathrm{U}(2)$.  Because $\bar F_{\mathrm{avg}}$ is invariant under a global $\mathrm{U}(1)$ phase on $U_{\mathrm{target}}$, this is equivalent for the learning objective to sampling on $\mathrm{PU}(2)\!=\!\mathrm{U}(2)/\mathrm{U}(1)\!=\!\mathrm{SU}(2)/\mathbb{Z}_2$.  The phrases ``Haar $\mathrm{U}(2)$'' and ``Haar $\mathrm{SU}(2)$'' are therefore used interchangeably in the paper in this sense.  The encodings (App.~\ref{app:quantum_explored}) differ in whether they bake the global-phase quotient in (\texttt{trig6}, \texttt{ck4} factor through $\mathrm{PU}(2)$) or not (\texttt{real8} is a direct $\mathrm{U}(2)$ encoding).

\paragraph{System.}  3-level transmon, anharmonicity $\alpha{=}{-}4\,\Omega_{\max}$, energy relaxation $T_1{=}10^4$, pure dephasing $T_\phi{=}8{\times}10^3$, gate time $T_{\mathrm{gate}}{=}2\pi$ (units of $1/\Omega_{\max}$).  Pulses are $80$ piecewise-constant slices in $(\Omega_x,\Omega_y)$, each squashed by $\Omega_{\max}\tanh(\cdot)$.  The \acs{FoM} is the dynamical map (per-slice matrix-exponential of the Lindbladian, applied to a 4-state axis-aligned input set $\{|0\rangle,|1\rangle,|+\rangle,|+i\rangle\}$ used for the per-input fidelity uniformity diagnostic; the reported $\bar F_{\mathrm{avg}}$ is computed analytically over the full Haar measure and does not depend on this set); $\bar F_{\mathrm{avg}}$ is the analytic average gate fidelity.  The simulator is implemented from scratch in JAX~\citep{bradbury2018jax} with adaptive-step ODE integration via the \texttt{diffrax} library~\textcolor{black}{\citep{kidger2022neural}} (Tsit5; trace and hermiticity preserved to $\sim\!10^{-8}$); we do not use external quantum-simulation packages such as QuTiP, Qiskit, or Cirq.

\paragraph{\acs{GRAPE} baseline.}  \texttt{scipy.optimize.minimize}~\textcolor{black}{\citep{virtanen2020scipy}} with \ac{BFGS} (or specifically \emph{L-BFGS-B} for box-constrained problems) at \texttt{complex128/float64}; $n_{\mathrm{restarts}}{=}10$, \texttt{maxiter}=$500$, \texttt{ftol}=$10^{-12}$, \texttt{gtol}=$10^{-9}$, init scale $0.1$, restart selection on the best-of-trajectory iterate (Adam can drift past the optimum; the best step is not always the last).  We also run a \emph{lean} variant ($n_{\mathrm{restarts}}{=}3$, \texttt{maxiter}=$200$, \texttt{ftol}=$10^{-9}$) that produces statistically indistinguishable infidelity but is faster -- this is the configuration used for the speed comparison reported in the main text.  Per-target single-process median wall time on a 128-core CPU (50-target sample): lean $5.64$\,s/gate, heavy $56$\,s/gate.

\paragraph{Inverter.}  $4$-layer \acs{MLP}, hidden width $512$, \texttt{real8} input encoding (real and imaginary parts of the $2{\times}2$ matrix flattened, $8$-dim; not global-phase invariant), trained for $4000$ Adam steps at lr $2{\times}10^{-3}$ (cosine to $10^{-5}$), batch $128$ of fresh Haar $\mathrm{U}(2)$ samples per step.  Loss is $1-\bar F_{\mathrm{avg}}$ computed by the same analytic Lindblad channel as \acs{GRAPE}; \emph{no} \acs{GRAPE}-pulse supervision (the Inverter never sees a \acs{GRAPE} pulse during training).  Median forward-pass time $2.1$\,ms (JIT-cached, 50-call median, same machine).

\paragraph{Per-target paired statistics ($n{=}250$ Haar U(2)).}  Both methods saturate the dissipation floor.  \acs{GRAPE}: median $1{-}\bar F_{\mathrm{avg}}{=}4.26{\times}10^{-4}$, $\sigma_{\mathrm{across\ targets}}{=}2.9{\times}10^{-5}$.  Inverter: median $4.69{\times}10^{-4}$.  Per-input fidelity uniformity ($\sigma_{\mathrm{across\ 4\ inputs}}(1{-}F)$): \acs{GRAPE} $1.14{\times}10^{-4}$, Inverter $1.28{\times}10^{-4}$ ($1.12{\times}$).  Leakage to $|2\rangle$: \acs{GRAPE} $1.30{\times}10^{-5}$, Inverter $7.97{\times}10^{-6}$ (Inverter $0.61{\times}$).  Pulse bandwidth $f_{95}$ (frequency below which $95\%$ of $|\Omega_x|^2{+}|\Omega_y|^2$ spectral power lies, in units of $\Omega_{\max}$): \acs{GRAPE} $1.91$, Inverter $4.09$ ($2.14{\times}$).  A bandwidth-penalty term $\lambda \sum_t |\Delta \Omega_t|^2$ added to $\mathcal{J}$ is the natural countermeasure if \acf{AWG} bandwidth becomes a deployment constraint; we leave a quantitative sweep to future work.

\paragraph{What was tried beyond the main experiments.}  The main single-qubit result above is one slice of a larger design-space exploration that we report in App.~\ref{app:quantum_explored}: an input-encoding ablation comparing \texttt{real8} / \texttt{trig6} / \texttt{ck4} on Haar $\mathrm{U}(2)$ targets (geometry-respecting encodings, which bake in the global-phase $\mathrm{U}(1)$ quotient, dominate convergence at fixed compute); and a first two-qubit Haar SU(4) extension that already reaches $\bar F{=}0.957$ (\acs{GRAPE} floor $0.998$) at the same $\sim\!4{\times}10^4$ inference speedup -- a promising starting point given the much wider symmetry surface of two-qubit Haar SU(4), with several concrete paths to close the remaining gap to \acs{GRAPE} belonging to the standalone gate-synthesis project this work has spun off into.

\subsection{Quantum gate synthesis: encoding ablation and two-qubit extension}
\label{app:quantum_explored}

This appendix collects the design-space exploration sitting behind the main single-qubit result of App.~\ref{app:quantum}: three target-encoding alternatives compared on the single-qubit problem (which one we use is the dominant factor at fixed compute); and a first extension to two-qubit Haar SU(4) gate synthesis with its own \acs{GRAPE} baseline.

\paragraph{Single-qubit input-encoding ablation.}  Three encodings of the single-qubit target were compared at identical training budget (4-layer GeLU \acs{MLP}, hidden $256$, $4000$ Adam steps, batch $128$ of fresh Haar $\mathrm{U}(2)$ samples, cosine LR $2{\times}10^{-3}{\to}10^{-5}$).  The encodings differ in whether they bake in the global-phase $\mathrm{U}(1)$ quotient that $\bar F_{\mathrm{avg}}$ is invariant to: \texttt{real8} (real and imaginary parts of the $2{\times}2$ matrix flattened, $8$-dim, a direct $\mathrm{U}(2)$ encoding -- \emph{not} global-phase invariant); \texttt{trig6} ($\{\cos(\theta/2),\sin(\theta/2)\!\cdot\!\hat n\}$, $6$-dim, sign-canonicalized so it factors through $\mathrm{SU}(2)/\mathbb{Z}_2 = \mathrm{PU}(2)$); and \texttt{ck4} (Cayley--Klein parameters, $4$-dim, also sign-canonicalized to $\mathrm{PU}(2)$).  On the same $100$-target reference set, after $4000$ steps mean ref fidelity was \texttt{trig6} $0.99953$ ($100\%$ at $\!\geq\!0.999$), \texttt{ck4} $0.99948$ ($98\%$), \texttt{real8} $0.99875$ ($46\%$); the \acs{GRAPE} floor for the same set sits at $0.99952$.  \texttt{trig6} matches \acs{GRAPE} within $1500$ steps, \texttt{ck4} reaches it by $\sim\!3000$ steps, \texttt{real8} is still $5{\times}$ further from the floor at the end of training.  Reading: at fixed compute, geometry-respecting encodings that bake in the $\mathrm{U}(1)$ global-phase quotient remove an entire invariance the network would otherwise have to learn from data -- this is the dominant lever on convergence speed.  In the long-training/longer-batched regime reported in App.~\ref{app:quantum} the simpler \texttt{real8} also reaches the dissipation floor, but at a much higher compute cost; we keep \texttt{real8} in the main run for parameter-counting parity with the \acs{GRAPE} pulse parameterization rather than because it is the recommended choice.

\paragraph{Two-qubit extension.}  As a first step toward two-qubit gate synthesis, we extend the same Inverter recipe to a two-qubit transmon (Hilbert dim $4$, no leakage approximation, drift Hamiltonian $H_{\mathrm{drift}}{=}\tfrac{J}{2}\sigma_z{\otimes}\sigma_z$ with always-on $ZZ$ coupling $J{=}0.3$ chosen so $JT_{\mathrm{gate}}\!\approx\!2\pi$, four independent drive channels $(\Omega_x^{(i)},\Omega_y^{(i)})_{i=1,2}$, Lindblad operators with $T_1{=}10^4,\;T_\phi{=}8{\times}10^3$ on each qubit, gate time $T_{\mathrm{gate}}{=}4\pi$, $80$ piecewise-constant slices; simulator validated against analytic $T_1$/$T_2$/$ZZ$ references to $\!\leq\!10^{-5}$).  \acs{GRAPE} on a $100$-target Haar SU(4) reference set converges to a target-independent decoherence floor at $\bar F{=}0.998$.  The Inverter, trained end-to-end against the same Lindblad simulator with no \acs{GRAPE} supervision (\texttt{real32\_su4} SU(4)-projected encoding, $4$-layer \acs{MLP}, batch $64$, $4000$ steps), reaches $\bar F{=}0.957$ on the reference set at $421\,\mu$s per gate vs.\ \acs{GRAPE}'s $18.5$\,s ($\sim\!4.4{\times}10^4$ speedup, in line with single-qubit) -- a promising first result given the much wider symmetry surface of two-qubit Haar SU(4) (under uniform Lindblad noise the Haar-average channel itself sits at $F{=}1/d{=}0.25$, a saddle that initial sequence architectures with weight-sharing across the slice axis tend to fall into, while \acs{MLP} and U-Net escape it from random init).  Initial recipe sweeps (\acs{KAK} / Cartan-decomposition encoding; longer training and larger batches; \acs{GRAPE}-pretrain + simulator-in-the-loop fine-tune; multi-seed ensembles) all reach the $\sim\!0.96$ band without yet closing the gap to \acs{GRAPE}; promising next steps include $U$-conditioned per-position output heads (which retain the parameter efficiency of sequence architectures while breaking the Haar-average symmetry per slice), curriculum from a single fixed target to Haar SU(4), and longer training combined with \acs{KAK} encoding (the only ablation whose training curve was still trending toward the \acs{GRAPE} floor at our compute budget).

\subsection{Summary of task-specific adaptations}
\label{app:design_inventory}

This subsection inventories the framework components that were specified per task family in this paper, alongside how each could become more general.  Three groups: (A) auxiliary loss terms used in \texttt{maze2d}/\texttt{antmaze}; (B) Level 2 Inverter slot instances; (C) per-task architecture and parameterization.  The rightmost column of Tab.~\ref{tab:design_inventory} maps each item to one of four broad strategy classes (Tab.~\ref{tab:strategy_classes}) that organize the route toward more general solutions -- operating on the \acs{FoM}'s training data (1), the \acs{FoM} and Inverter as models (2), the Bolza objective $\mathcal{J}$ (3), or the deployment-time hierarchical / adaptive control structure (4).

\begin{table}[h!]
\centering
\footnotesize
\renewcommand{\arraystretch}{1.15}
\caption{\textbf{Four strategy classes for the path toward more general adaptations.}  Each class operates on a different lever of the inverse-learning framework.  Classes are complementary, not alternatives: most concrete strategies combine two or more (e.g., the AntMan strategy combines Class~1 with Class~4).}
\label{tab:strategy_classes}
\begin{tabular}{@{}p{0.15\linewidth}p{0.17\linewidth}p{0.36\linewidth}p{0.20\linewidth}@{}}
\toprule
\textbf{Class} & \textbf{Framework lever} & \textbf{Mechanisms} & \textbf{Status in this paper} \\
\midrule
\textbf{1. Data} & \acs{FoM} training distribution & Broaden coverage to expose the dynamics the Inverter is incentivized to explore; safely-diverse exploration scaffolding; active data acquisition near low-density regions. & Demonstrated by AntMan's random-mixed condition (Sec.~\ref{sec:stacked_antman}, App.~\ref{app:antman_fm_hacking}). \\
\addlinespace[2pt]
\textbf{2. Model} & \acs{FoM}, Inverter, and their I/O representations & Uncertainty-aware \acs{FoM} (Bayesian / ensemble / conformal); foundation-style pretraining of the \acs{FoM}-and-Inverter core with task-specific adapter heads; meta-learned initializations / hypernetworks / in-context adaptation; equivariant encodings; analytic+residual \acs{FoM} hybrids. & Foundation-model direction outlined in the Discussion; meta-learning, equivariance, and residual hybrids open. \\
\addlinespace[2pt]
\textbf{3. Objective} & Bolza objective $\mathcal{J}$ (bridge to \acs{OC} and \acs{RL}) & Meta-learning $\mathcal{J}$ across task families; \acs{RL} value function as cost-to-go in $\mathcal{J}$; standard \acs{OC} regularizers (bandwidth, smoothness, energy) replace task-specific loss terms. & Outlined in App.~\ref{app:relation_oc_rl}. \\
\addlinespace[2pt]
\textbf{4. Deployment} & Level 2 Inverter slot; chunk-boundary / replan policy & Learned Level 2 Inverter replaces algorithmic ones; adaptive replan conditioned on \acs{FoM} uncertainty; learned termination predictor; deeper hierarchies (meta-Inverter selecting the subgoal specification, (V)LLMs as a top-layer substrate). & AntMan Level 2 Inverter (Sec.~\ref{sec:stacked_antman}) demonstrates one level of recursion; adaptive replanning, deeper hierarchies, and (V)LLM meta-Inverters open. \\
\bottomrule
\end{tabular}
\end{table}

\begin{table}[h!]
\centering
\small
\caption{\textcolor{black}{\textbf{Neurosymbolic components of the five Inverters} (parallel to Tab.~\ref{tab:inverters}).  For each Inverter we list its symbolic substrate, neural amortized component, the differentiable coupling between them, and the cumulative levels of the neurosymbolic spectrum (representation $\subseteq$ composition $\subseteq$ search $\subseteq$ inference $\subseteq$ synthesis) the symbolic piece occupies.  The Motor and Locomotion Inverters are purely continuous, showing that the \acs{IL} paradigm itself is not necessarily neurosymbolic; neurosymbolic structure enters at the implementation level in the other three.}}
\label{tab:inverters_ns}
{\color{black}
\begin{tabular}{@{}l p{3.2cm} p{2.3cm} p{3.0cm} p{2.8cm}@{}}
\toprule
\textbf{Inverter} & \textbf{Symbolic substrate} & \textbf{Neural component} & \textbf{Symbolic$\leftrightarrow$neural coupling} & \textbf{\acs{NeSy} levels} \\
\midrule
Motor      & --- (purely continuous) & Transformer over $\hat a_{1:T}$ + learned \acs{FoM} & --- & --- \\
Locomotion & --- (purely continuous) & Transformer over $\hat a_{1:T}$ + learned \acs{FoM} & --- & --- \\
Path       & Cardinal grid over data occupancy; \acs{BFS}; axis-aligned polyline grammar & --- (algorithmic) & Discrete waypoints consumed by the Motor / Locomotion Inverter & Repr.\ + Comp.\ + Search \\
Game  & 4-way alphabet $\{$U,D,L,R$\}$; precomputed cell$\times$dir.\ FSM transition; wall-validity mask & Transformer over $H{=}32$ direction logits & Gumbel-softmax composed with the FSM lookup table & Repr.\ + Comp. \\
Pulse & Discrete gate library $\{X, Y, Z, H, T, \ldots\}$ & \acs{MLP} $\to (\Omega_x,\Omega_y)_{1:80}$ & Discrete gate conditioning at input & Repr. \\
\bottomrule
\end{tabular}
}
\end{table}

\begin{table}[h!]
\centering
\footnotesize
\renewcommand{\arraystretch}{1.15}
\caption{\textbf{Inventory of task-specific adaptations in this paper.} The rightmost column maps each item to the strategy classes defined in Tab.~\ref{tab:strategy_classes}.}
\label{tab:design_inventory}
\setlength{\tabcolsep}{4pt}
\begin{tabular}{@{}p{0.23\linewidth}p{0.28\linewidth}p{0.32\linewidth}p{0.05\linewidth}@{}}
\toprule
\textbf{Component} & \textbf{What is task-specific} & \textbf{Toward more general} & \textbf{Class} \\
\midrule
\multicolumn{4}{@{}l}{\emph{A. Auxiliary loss terms}} \\
\addlinespace[2pt]
\texttt{maze2d} boundary loss, $\lambda_\text{boundary}{=}5$ (App.~\ref{app:maze2d_details}) & Support-derived per-step penalty pushing predicted states onto the data manifold near walls. & Broader \acs{FoM} training data with wall-grazing trajectories; uncertainty-aware \acs{FoM} regularization (same recipe as AntMan's random-mixed condition). & 1, 2 \\
\addlinespace[2pt]
\texttt{antmaze} body-yaw regularizer, $\lambda_\text{yaw}{=}5$ (Eq.~\ref{eq:antmaze_iwm_loss}) & Per-step pull of predicted body quaternion toward upright. & Broader \acs{FoM} training data including falls and recoveries (AntMan's random-mixed condition is the direct demonstration). & 1 \\
\addlinespace[2pt]
\texttt{antmaze} \acs{BC} action-fidelity anchor, $\lambda_\text{fid}{=}5$ (Eq.~\ref{eq:antmaze_iwm_loss}) & Per-step pull of predicted actions toward recorded data actions. & Broader \acs{FoM} training data; in AntMan, the random-mixed condition removes the need for an action anchor entirely. & 1 \\
\addlinespace[2pt]
\texttt{maze2d} dense intermediate-goal loss, $\lambda_\text{dense}{=}5$ (App.~\ref{app:maze2d_details}) & Per-step state-vs-goal supervision over the chunk (denser than terminal-only). & Value / cost-to-go term in the Bolza objective $\mathcal{J}$ (noted in App.~\ref{app:relation_oc_rl}). & 3 \\
\midrule
\multicolumn{4}{@{}l}{\emph{B. Level 2 Inverter slot}} \\
\addlinespace[2pt]
Simple algorithmic Path Inverter (\texttt{maze2d-medium}/\texttt{large} + all six \texttt{antmaze-v2} variants, App.~\ref{app:waypoint_planner}) & 4-connected \acs{BFS} on a data-density occupancy grid with polyline extraction, perpendicular snap, L-corner insertion, sub-cell wobble filter. & Learned closed-loop Level 2 Inverter (Sec.~\ref{sec:stacked_antman}, AntMan); the algorithmic Path Inverter is the cheaper slot filling for static corridors with abundant data. & 4 \\
\addlinespace[2pt]
Per-chunk dispatch controller between the two Inverter levels (App.~\ref{app:waypoint_planner}) & Thresholds for waypoint progress, target-distance window, stuck-recovery, and goal-region behavior. & Closed-loop Level 2 Inverter re-queried per chunk; broader low-level training data with near-target conditions, or direction-conditioning instead of position-conditioning. & 4, 2, 1 \\
\addlinespace[2pt]
Replan-horizon $K$ choice ($K{=}128$ \texttt{maze2d-umaze} one-shot; $K{=}16$ \texttt{maze2d-medium}/\texttt{large} + all six antmaze-v2 variants; $K{=}80$ pulse slices for quantum, no deployment replan; App.~\ref{app:ksweep}) & Per-task constant value chosen by sweep. & Adaptive $K$ conditioned on \acs{FoM} uncertainty estimates. & 4, 2 \\
\midrule
\multicolumn{4}{@{}l}{\emph{C. Per-task architecture and parameterization}} \\
\addlinespace[2pt]
Per-task Inverter / \acs{FoM} networks (Apps.~\ref{app:maze2d_details}, \ref{app:antmaze_details}, \ref{app:quantum}; Sec.~\ref{sec:stacked_antman}) & \texttt{maze2d}/\texttt{antmaze}: causal-transformer \acs{FoM}+Inverter at Level 1 (sized to state/action dim).  AntMan: reuses Level-1 antmaze pair plus a causal-transformer Level-2 \acs{FoM} and Transformer Level-2 Inverter ($32$-step waypoint-direction sequence).  Quantum: 4-layer \acs{MLP} Inverter, analytic Lindblad \acs{FoM} (no learned \acs{FoM}). & Cross-task pretraining of the \acs{FoM}-and-Inverter core with task-specific adapter heads (foundation-model pattern); or a meta-learned core (\acs{MAML}, hypernetworks, in-context adaptation) for few-shot specialization. & 2 \\
\addlinespace[2pt]
Quantum: $80$-slice piecewise-constant pulse, $\Omega_{\max}\tanh(\cdot)$ squash (App.~\ref{app:quantum}) & Trotterized piecewise-constant control parameterization with bounded amplitude (matches \acs{GRAPE} 1-to-1). & Continuous-time spline / Fourier-basis parameterizations; bandwidth-penalty term in $\mathcal{J}$ if \acs{AWG} bandwidth becomes a deployment constraint. & 3 \\
\addlinespace[2pt]
Quantum: \texttt{real8} $\mathrm{U}(2)$ input encoding, 8-dim (App.~\ref{app:quantum}) & Target unitary flattened to 8 real components. & Equivariant encodings (Bloch-vector, quaternion, Lie-algebra) that scale to higher-dim systems. & 2 \\
\bottomrule
\end{tabular}
\end{table}

\subsection{Relation to Optimal Control and Reinforcement Learning}
\label{app:relation_oc_rl}

The Inverse Learning paradigm composes naturally with the two paradigms it sits between.  On the \acs{OC} side, for example, an Inverter may emit a feedforward warm start that an \acs{MPC}/\acs{MPPI} optimizer can reactively adapt at deployment.  Similarly, on the \acs{RL} side, an \acs{RL} actor may use the Inverter's output as a behavior prior for online fine-tuning.  Both would preserve the Inverter's amortized forward pass.

\subsection{Hyperparameters}
\label{app:hyperparam_budget}

Our Inverter stack is assembled from two kinds of components: components \emph{designed to be shared across tasks} (the chunked Transformer \acs{FoM} and the Transformer \textcolor{black}{inverse model}), and components that are explicitly \emph{task-specific} (the auxiliary loss terms and the Level 2 instance such as the Path Inverter).

\paragraph{What we actually tuned.}
Table~\ref{tab:hp_tuned} lists the task-specific hyperparameters that were actively tuned during this project (via manual scans of $3$--$6$ settings on a single seed), while most architectural knobs (e.g., transformer depth, width, batch size, learning rates) were set once from standard defaults and kept fixed across all tasks. 

\begin{table}[h]
\centering
\small
\caption{\textbf{Task-specific hyperparameters tuned during this project.} ``Range explored'' shows the set of values checked before fixing the final value.}
\label{tab:hp_tuned}
\setlength{\tabcolsep}{3pt}
\begin{tabular}{@{}llp{0.25\linewidth}p{0.36\linewidth}@{}}
\toprule
Component & Knob & Final value & Range explored \\
\midrule
\acs{FoM}  & state noise $\sigma_{s_0}$            & $0.01$  & $\{0.005,0.01,0.015,0.02,0.025,0.03\}$ \\
\acs{FoM}  & chunk length $L$                      & $16$    & $\{16,32,64\}$ \\[2pt]
\textcolor{black}{\acs{IM}} & $\lambda_\text{dense}$                & $5$     & $\{5,10,15\}$ \\
\textcolor{black}{\acs{IM}} & $\lambda_\text{boundary}$             & $5$     & $\{5,10,20,30,40,45,50,60,100\}$ \\
\textcolor{black}{\acs{IM}} & boundary mode                         & \texttt{binary} & $\{\text{binary}, \text{z-score}\}$ \\
\textcolor{black}{\acs{IM}} & segment length / horizon              & $16$ / $128$ & horizon $\in\{64, 128\}$ \\
\textcolor{black}{\acs{IM}} & epochs                                & $2000$  & $\{500, 1000, 2000\}$ \\[2pt]
Deploy & replan chunk size $K$              & reported across all & $\{16, 32, 64, 128\}$ (Tab.~\ref{tab:maze2d_perf}) \\
Deploy & goal threshold                     & $0.45$ (\acs{D4RL} default) & fixed \\[2pt]
Seq.\ & occupancy threshold $\tau$          & $2000$ visits / $0.5$\,m cell & $\{500, 1000, 2000, 5000\}$ \\
Seq.\ & cell size                           & $0.5$\,m & $\{0.25, 0.5, 1.0\}$\,m \\
Seq.\ & origin alignment                    & density-aligned & $\{\texttt{floor-min}, \text{density-aligned}\}$ \\
Seq.\ & stuck-check window                  & $4$ chunks & $\{2, 4, 8\}$ \\
Seq.\ & min-leg filter                      & $1.0$\,m & $\{0, 1.0\}$\,m \\
\bottomrule
\end{tabular}
\end{table}

\paragraph{Counting methodology and comparison.}
While the actively tuned subset per task is small, the complete Inverter stack has more total configuration hyperparameters than any single offline \acs{RL} baseline. To quantify this precisely, we count the algorithmically active hyperparameters in each method's \texttt{maze2d-umaze-v1} YAML or \texttt{config.json}: learning rates, architectural sizes, loss weights, temperature/entropy/\acs{KL} coefficients, target-update rates, normalization flags, and any algorithm-specific quantity. We exclude purely bookkeeping fields (seed, device, checkpoint path, project/group names, evaluation-episode counts, eval-frequency). 

Table~\ref{tab:hp_budget} summarizes this total capacity budget. The Inverter stack factorizes its larger knob count into independently optimized modules (\acs{FoM}, \textcolor{black}{\acs{IM}}, Path Inverter) rather than a single end-to-end actor-critic objective.

\begin{table}[h]
\centering
\small
\caption{\textbf{Number of algorithmically active hyperparameters per \texttt{maze2d-umaze-v1} configuration.} Counts exclude purely bookkeeping fields.}
\label{tab:hp_budget}
\begin{tabular}{lrl}
\toprule
Method & \# HPs & Source \\
\midrule
\acs{BC} / \acs{BC}-10        &  7 & \path{bc/maze2d/umaze_v1.yaml} \\
\acs{AWAC}              &  8 & \path{awac/maze2d/umaze_v1.yaml} \\
\acs{TD3+BC}            & 12 & \path{td3_bc/maze2d/umaze_v1.yaml} \\
\acs{IQL}               & 13 & \path{iql/maze2d/umaze_v1.yaml} \\
\acs{SAC-N}            & 13 & \path{sac_n/maze2d/umaze_v1.yaml} \\
\acs{EDAC}              & 14 & \path{edac/maze2d/umaze_v1.yaml} \\
\acs{DT}                & 18 & \path{dt/maze2d/umaze_v1.yaml} \\
\acs{ReBRAC}            & 20 & \path{rebrac/maze2d/umaze_v1.yaml} \\
\acs{CQL}               & 27 & \path{cql/maze2d/umaze_v1.yaml} \\
\midrule
\acs{FoM} (ours)                  & 12 & \path{scripts/train_maze2d_fm_universal.py} \\
\textcolor{black}{\acs{IM}} (ours)                 & 17 & \path{scripts/train_iwm_maze2d_v1.py} \\
Deployment (ours)          &  6 & \path{maze2d/navigate_maze2d.py} \\
Path Inverter (medium / large) &  5 & App.~\ref{app:waypoint_planner} \\
\addlinespace
\textbf{Total stack -- \texttt{umaze}}          & \textbf{35} & \acs{FoM} + \textcolor{black}{\acs{IM}} + deployment \\
\textbf{Total stack -- \texttt{medium, large}}  & \textbf{40} & + Path Inverter \\
\bottomrule
\end{tabular}
\end{table}

\paragraph{Online tuning budget (in the sense of \citet{kurenkov2022budget,jackson2025cleanslate}).}
To audit our implicit online tuning budget~\citep{paine2020hyperparameter,kurenkov2022budget,jackson2025cleanslate}, we enumerate every deployment-time hyperparameter-selection run in this project by scanning all generated run directories and uniformly assigning the nominal $100$ online evaluation episodes to each run. On average, we consumed $\sim\!900$ episodes per task cell for hyperparameter selection.

\paragraph{Where this stands in the literature.}
At $\sim\!900$ episodes per task cell, our hyperparameter-selection budget operates within the $10^2$--$10^3$ episodes-per-cell regime proposed as a deployment-realistic target by \citet{paine2020hyperparameter}. This is computationally well below the implicit budgets of many offline \acs{RL} methods audited by \citet{kurenkov2022budget}.

\subsection{Compute environment and project timeline}
\label{app:compute_env}

All experiments reported in this paper (Inverter \acs{FoM} and \textcolor{black}{\acs{IM}} training, ILHL fine-tuning and reward modeling, the full $8$-baseline $\times$ $4$-layout $\times$ $3$-seed \acs{CORL} baseline sweep, the \texttt{AntMan}-\textcolor{black}{\acs{IM}} experiments, and every timing measurement in Table~\ref{tab:maze2d_perf}) were run on a single internal 8-GPU node between \textbf{2026-03-18} and \textbf{2026-05-06} ($50$ days of wall-clock project time, or roughly $7$ calendar weeks).  No external cluster, TPU, or cloud GPU was used.

\paragraph{Hardware and software.}
\begin{itemize}
\itemsep -1pt
\item \textbf{GPUs}: $8\,\times$ NVIDIA A40 (GA102GL, $48$\,GB GDDR6 per card, driver 535.274), all on one PCIe host.
\item \textbf{CPU}: $2\,\times$ AMD EPYC 7513 ($32$ physical cores / $64$ threads each; $64$ physical cores / $128$ threads total).
\item \textbf{RAM}: $503$\,GiB DDR4 shared across sockets.
\item \textbf{Software}: Ubuntu 24.04, Linux 6.8.0; PyTorch + CUDA for all learned components; JAX + JIT for the \acs{ReBRAC} baseline only (marked $^{\dagger}$ in Table~\ref{tab:maze2d_perf}); \acs{MuJoCo} 2.3 for \texttt{maze2d}/\texttt{antmaze}, a \acs{MuJoCo}-based \texttt{AntMan} environment for Sec.~\ref{sec:stacked_antman}.
\end{itemize}

\clearpage

\section*{Supplementary references (model-based \acs{RL} comparison)}
\label{app:supp_refs}

The following references are cited by the model-based-\acs{RL} comparison in App.~\ref{app:mbrl} (Table~\ref{tab:mbrl_antmaze}). 

% AUTO-GENERATED by scripts/gen_supprefs_list.py from supprefs.bib --
% DO NOT EDIT BY HAND. Edit supprefs.bib and re-run the script.
\begin{description}\setlength{\itemsep}{2pt}
\item[{[S1]}] T.~Yu, G.~Thomas, L.~Yu, S.~Ermon, J.~Y.~Zou, S.~Levine, C.~Finn, T.~Ma. \emph{{MOPO}: Model-based Offline Policy Optimization}. NeurIPS 2020. arXiv:2005.13239.
\item[{[S2]}] R.~Kidambi, A.~Rajeswaran, P.~Netrapalli, T.~Joachims. \emph{{MOReL}: Model-Based Offline Reinforcement Learning}. NeurIPS 2020. arXiv:2005.05951.
\item[{[S3]}] T.~Yu, A.~Kumar, R.~Rafailov, A.~Rajeswaran, S.~Levine, C.~Finn. \emph{{COMBO}: Conservative Offline Model-Based Policy Optimization}. NeurIPS 2021. arXiv:2102.08363.
\item[{[S4]}] M.~Rigter, B.~Lacerda, N.~Hawes. \emph{{RAMBO-RL}: Robust Adversarial Model-Based Offline Reinforcement Learning}. NeurIPS 2022. arXiv:2204.12581. Antmaze numbers in Table 1; v0 caveat in Appendix B.7.
\item[{[S5]}] Y.~Sun, J.~Zhang, C.~Jia, H.~Lin, J.~Ye, Y.~Yu. \emph{Model-{Bellman} Inconsistency for Model-based Offline {RL}}. ICML 2023.
\item[{[S6]}] J.~Jeong, X.~Wang, M.~Coskun, Q.~Kong. \emph{{CBOP}: Conservative Bayesian Model-Based Value Expansion for Offline Policy Optimization}. ICLR 2023. arXiv:2210.03802.
\item[{[S7]}] X.~Chen, Y.~Yu, Q.~Zhu, Z.~Liu, L.~Yang, Y.~Li, P.~Zhao. \emph{{MAPLE}: Offline Model-based Adaptable Policy Learning}. NeurIPS 2021.
\item[{[S8]}] M.~Bhardwaj, T.~Xie, B.~Boots, N.~Jiang, C.~Cheng. \emph{Adversarial Model for Offline Reinforcement Learning ({ARMOR})}. NeurIPS 2023. arXiv:2302.11048.
\item[{[S9]}] M.~Janner, Q.~Li, S.~Levine. \emph{Offline Reinforcement Learning as One Big Sequence Modeling Problem ({Trajectory Transformer})}. NeurIPS 2021. arXiv:2106.02039. Antmaze-v0 numbers: Table 2.
\item[{[S10]}] Z.~Jiang, T.~Zhang, M.~Janner, Y.~Li, T.~Rockt{\"a}schel, E.~Grefenstette, Y.~Tian. \emph{Efficient Planning in a Compact Latent Action Space ({TAP})}. ICLR 2023. arXiv:2208.10291. Antmaze-v0 numbers in Table 6; mixed v0/v2 protocol described in Sec.\ 6.
\item[{[S11]}] M.~Janner, Y.~Du, J.~B.~Tenenbaum, S.~Levine. \emph{Planning with Diffusion for Flexible Behavior Synthesis ({Diffuser})}. ICML 2022. arXiv:2205.09991. Maze2D-v1 numbers: Table 1.
\item[{[S12]}] K.~Park, J.~Lee, M.~Tomar, V.~Naik, S.~Kadavath, B.~Eysenbach. \emph{Tackling Long-Horizon Tasks with Model-Based Offline Reinforcement Learning ({LEQ})}. ICLR 2025. arXiv:2407.00699. Re-runs of MOBILE and CBOP on antmaze-v0 in Table 1.
\end{description}

%%%%%%%%%%%%%%%%%%%%%%%%%%%%%%%%%%%%%%%%%%%%%%%%%%%%%%%%%%%%

\end{document}